\documentclass{article}

\PassOptionsToPackage{sort, compress, numbers}{natbib}

\usepackage[preprint]{neurips_2021}

\usepackage[utf8]{inputenc} % allow utf-8 input
\usepackage[T1]{fontenc}    % use 8-bit T1 fonts
\usepackage[bookmarksnumbered=true, unicode]{hyperref}
\usepackage{url}            % simple URL typesetting
\usepackage{booktabs}       % professional-quality tables
\usepackage{amsfonts}       % blackboard math symbols
\usepackage{nicefrac}       % compact symbols for 1/2, etc.
\usepackage{microtype}      % microtypography
\usepackage{xcolor}         % colors
\usepackage{amsmath}
\usepackage{amssymb}
\usepackage[amsmath, thmmarks]{ntheorem}
\usepackage{bm}
\usepackage{textgreek}
\usepackage{graphicx}
\usepackage{subfigure}
\usepackage{tikz}
\usepackage{makecell}
\usepackage{multirow}
\usepackage{enumitem}

\theoremseparator{.}
\newtheorem{theorem}{Theorem}
\newtheorem{proposition}[theorem]{Proposition}
\newtheorem{lemma}[theorem]{Lemma}
\newtheorem{corollary}[theorem]{Corollary}

\qedsymbol{\ensuremath{\Box}}
{
   \theoremstyle{nonumberplain}
   \theoremheaderfont{\bfseries}
   \theorembodyfont{\normalfont}
   \theoremsymbol{\ensuremath{\Box}}
   \newtheorem{proof}{Proof}
}
\newcommand{\drawpath}[1]{
    \draw[-] \foreach \x/\y in {#1} {(v\x)--(v\y)};
}
\newcommand{\nodethree}{
    \node[dot] (v1) at (0, 1) {};
    \node[dot] (v2) at (0, 0) {};
    \node[dot] (v3) at (1, 0) {};
}
\newcommand{\nodefour}{
    \tikzstyle{every node} = [dot]
    \node (v1) at (0, 0) {};
    \node (v2) at (0, 1) {};
    \node (v3) at (1, 1) {};
    \node (v4) at (1, 0) {};
}

\newcommand{\GTHREE}[1]{
    \begin{tikzpicture}[scale=0.5, dot/.style={draw, circle, minimum size =
                2mm, inner sep = 0pt, outer sep=0pt, fill=black}]
    \nodethree
    \ifthenelse{\equal{#1}{1}} {
    \drawpath{1/2, 2/3}
    }{}
    \ifthenelse{\equal{#1}{2}} {
    \drawpath{1/2, 2/3, 1/3}
    }{}
    \end{tikzpicture}
}
\newcommand{\GCLIQUE}{
    \begin{tikzpicture}[scale=0.5, dot/.style={draw, circle, minimum size =
                2mm, inner sep = 0pt, outer sep=0pt, fill=black}]
    \nodefour
    \drawpath{1/2, 2/3, 3/4, 4/1, 2/4, 1/3}
    \end{tikzpicture}
}
\DeclareMathOperator*{\concat}{\scalebox{1.25}[2.0]{$\parallel$}}

\DeclareFontFamily{U}  {MnSymbolC}{}
\DeclareSymbolFont{MnSyC}         {U}  {MnSymbolC}{m}{n}
\DeclareFontShape{U}{MnSymbolC}{m}{n}{
    <-6>  MnSymbolC5
   <6-7>  MnSymbolC6
   <7-8>  MnSymbolC7
   <8-9>  MnSymbolC8
   <9-10> MnSymbolC9
  <10-12> MnSymbolC10
  <12->   MnSymbolC12}{}
\DeclareFontShape{U}{MnSymbolC}{b}{n}{
    <-6>  MnSymbolC-Bold5
   <6-7>  MnSymbolC-Bold6
   <7-8>  MnSymbolC-Bold7
   <8-9>  MnSymbolC-Bold8
   <9-10> MnSymbolC-Bold9
  <10-12> MnSymbolC-Bold10
  <12->   MnSymbolC-Bold12}{}

\DeclareMathSymbol{\upY}{\mathbin}{MnSyC}{41}
\DeclareMathSymbol{\diamondplus}{\mathbin}{MnSyC}{124}
\title{Going Deeper into Permutation-Sensitive \\ Graph Neural Networks}

\author{%
  Zhongyu Huang\textsuperscript{*} \\
  NLPR, Institute of Automation \\
  Chinese Academy of Sciences \\
  \texttt{huangzhongyu2020@ia.ac.cn}
  \And
  Yingheng Wang\textsuperscript{*} \\
  Tsinghua University \\
  Johns Hopkins University \\
  \texttt{ywang584@jhu.edu}
  \AND
  Chaozhuo Li \\
  Microsoft Research Asia \\
  \texttt{cli@microsoft.com} \\
  \makebox[14.5em]{} \\
  \And
  Huiguang He\textsuperscript{\dag} \\
  NLPR, Institute of Automation \\
  Chinese Academy of Sciences \\
  \texttt{huiguang.he@ia.ac.cn}
}

\begin{document}

\maketitle

\begin{abstract}
    The invariance to permutations of the adjacency matrix, i.e., graph isomorphism, is an overarching requirement for Graph Neural Networks (GNNs).
    Conventionally, this prerequisite can be satisfied by the invariant operations over node permutations when aggregating messages.
    However, such an invariant manner may ignore the relationships among neighboring nodes, thereby hindering the expressivity of GNNs.
    In this work, we devise an efficient permutation-sensitive aggregation mechanism via permutation groups, capturing pairwise correlations between neighboring nodes.
    We prove that our approach is strictly more powerful than the 2-dimensional Weisfeiler-Lehman (2-WL) graph isomorphism test and not less powerful than the 3-WL test.
    Moreover, we prove that our approach achieves the linear sampling complexity.
    Comprehensive experiments on multiple synthetic and real-world datasets demonstrate the superiority of our model.
\end{abstract}

%%%%%%%%%%%%%%%%%%%%%%%%%%%%%%%%%%%%%%%%%%%%%%%%%%%%%%%%%%%%

\section{Introduction}
\label{sec:intro}

The invariance to permutations of the adjacency matrix, i.e., graph isomorphism, is a key inductive bias for graph representation learning \cite{murphy2019relational}.
Graph Neural Networks (GNNs) invariant to graph isomorphism are more amenable to generalization as different orderings of the nodes result in the same representations of the underlying graph.
% \cite{gilmer2017neural, chen2020can, vignac2020building}
Therefore, many previous studies \cite{duvenaud2015convolutional, kipf2017semi, gilmer2017neural, hamilton2017inductive, ying2018hierarchical, xu2019powerful, maron2019invariant} devote much effort to designing permutation-invariant aggregators to make the overall GNNs \emph{permutation-invariant} (permutation of the nodes of the input graph does not affect the output) or \emph{permutation-equivariant} (permutation of the input permutes the output) to node orderings.

\begin{figure}
    \centering
    \subfigure[Msg-passing scheme (permutation-invariant)]{
	\includegraphics[width=0.22\linewidth]{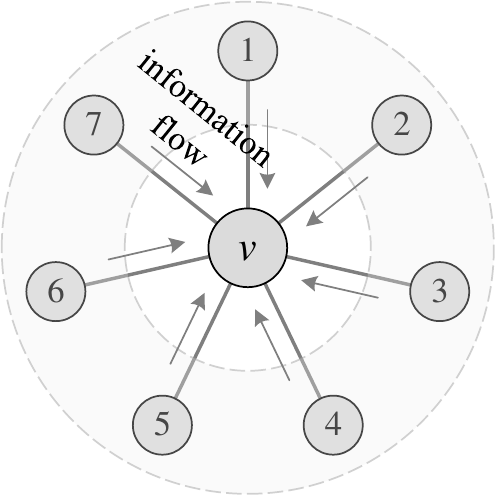}
	\label{fig:comparison:MP}
	}
	\ \ 
    \subfigure[GraphSAGE with an LSTM aggregator]{
	\includegraphics[width=0.22\linewidth]{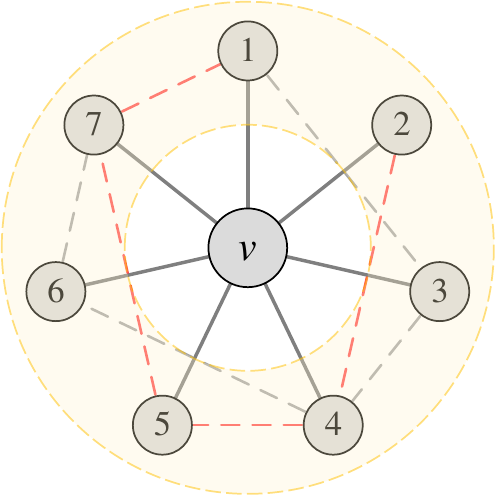}
	\label{fig:comparison:GraphSAGE}
	}
	\ \ 
    \subfigure[Janossy Pooling with $\pi$-SGD optimization]{
	\includegraphics[width=0.22\linewidth]{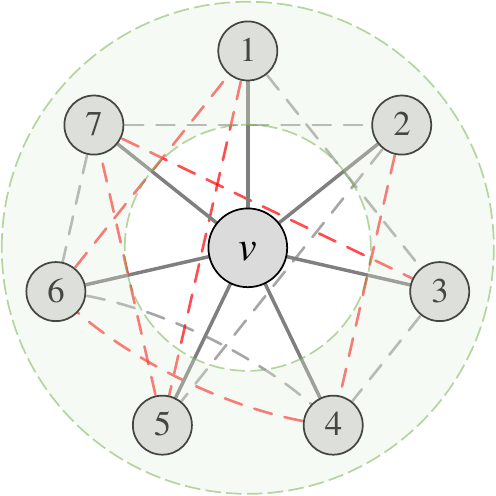}
	\label{fig:comparison:Janossy}
	}
	\ \ 
    \subfigure[Our goal of capturing all pairwise correlations]{
	\includegraphics[width=0.22\linewidth]{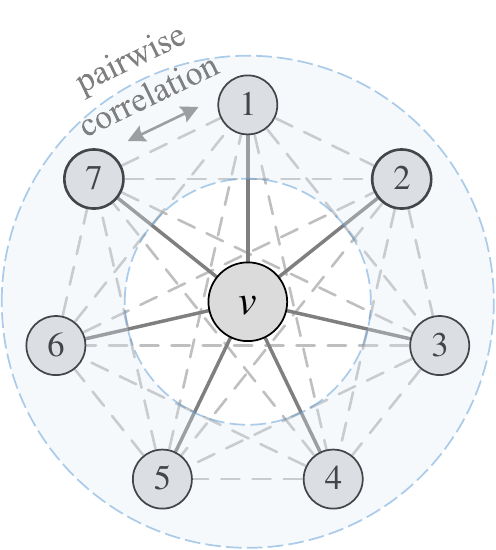}
	\label{fig:comparison:goal}
	}
    \caption{Comparison of the pairwise correlations modeled by various aggregation functions in 1-hop neighborhood.
    Here we illustrate with one central node $v$ and $n = 7$ neighbors.
    The dashed lines represent the pairwise correlations between neighbors modeled by the aggregators, the real topological connections between neighbors are hidden for clarity.
    Subfigure (b) shows 2 sampled batches with the neighborhood sample size $k = 5$.
    Subfigure (c) shows 2 sampled permutations.
    Dashed lines \textcolor{gray}{- -} and \textcolor{red}{- -} in (b)/(c) denote different batches/permutations.}
    \label{fig:comparison}
\end{figure}

Despite their great success, \citet{kondor2018covariant} and \citet{de2020natural} expound that such a permutation-invariant manner may hinder the expressivity of GNNs.
Specifically, the strong symmetry of these permutation-invariant aggregators presumes equal statuses of all neighboring nodes, ignoring the relationships among neighboring nodes.
% Node representations do not encode whether two nodes have the same neighbor or distinct neighbors with the same features, limiting their ability to learn an expressive representation of the entire graph \cite{murphy2019relational}.
Consequently, the central nodes cannot distinguish whether two neighboring nodes are adjacent, failing to recognize and reconstruct the fine-grained substructures within the graph topology.
% \cite{loukas2020graph}
As shown in Figure \ref{fig:comparison:MP}, the general Message Passing Neural Networks (MPNNs) \cite{gilmer2017neural} can only explicitly reconstruct a star graph from the 1-hop neighborhood, but are powerless to model any connections between neighbors \cite{chen2020can}.
To address this problem, some latest advances \cite{chen2020can, thiede2021autobahn, zhao2022stars, bevilacqua2022equivariant} propose to use subgraphs or ego-nets to improve the expressive power while preserving the property of permutation-invariance.
Unfortunately, they usually suffer from high computational complexity when operating on multiple subgraphs \cite{bevilacqua2022equivariant}.

In contrast, the \emph{permutation-sensitive} (as opposed to permutation-invariant) function\footnote{One of the typical permutation-sensitive functions is Recurrent Neural Networks (RNNs), e.g., Simple Recurrent Network (SRN) \cite{elman1990finding}, Gated Recurrent Unit (GRU) \cite{cho2014learning}, and Long Short-Term Memory (LSTM) \cite{hochreiter1997long}.} can \mbox{be regarded} as a ``symmetry-breaking'' mechanism, which breaks the equal statuses of neighboring nodes.
% the isomorphism counting of graph substructures is employed as a symmetry breaking mechanism to disambiguate neighbours \cite{bodnar2021weisfeilers}.
The relationships among neighboring nodes, e.g., the \emph{pairwise correlation} between each pair of neighboring nodes, are explicitly modeled in the permutation-sensitive paradigm.
These pairwise correlations help capture whether two neighboring nodes are connected, thereby exploiting the local graph substructures to improve the expressive power.
We illustrate a concrete example in Appendix \ref{sec:example}.

Different permutation-sensitive aggregation functions behave variously when modeling pairwise correlations.
GraphSAGE with an LSTM aggregator \cite{hamilton2017inductive} in Figure \ref{fig:comparison:GraphSAGE} is capable of modeling some pairwise correlations among the \emph{sampled subset} of neighboring nodes.
Janossy Pooling with the $\pi$-SGD strategy \cite{murphy2019janossy} in Figure \ref{fig:comparison:Janossy} samples random permutations of \emph{all} neighboring nodes, thus modeling pairwise correlations more efficiently.
The number of modeled pairwise correlations is proportional to the number of sampled permutations.
After sampling permutations with a costly nonlinear complexity of $\mathcal{O}(n \ln n)$ (see Appendix \ref{sec:obs} for detailed analysis), all the pairwise correlations between $n$ neighboring nodes can be modeled and all the possible connections are covered.
% In this way, our ultimate goal in Figure \ref{fig:comparison:goal} can be achieved.

In fact, previous works \cite{murphy2019janossy, murphy2019relational} have explored that incorporating permutation-sensitive functions into GNNs is indeed an effective way to improve their expressive power.
Janossy Pooling \cite{murphy2019janossy} and Relational Pooling \cite{murphy2019relational} both design the most powerful GNN models by exploiting \mbox{permutation-sensitive} functions to cover all $n!$ possible permutations.
They explicitly learn all representations of the underlying graph with possible $n!$ node orderings to guarantee the permutation-invariance and generalization capability, overcoming the limited generalization of permutation-sensitive GNNs \cite{vignac2020building}.
However, the complete modeling of all $n!$ permutations also leads to an intractable computational complexity $\mathcal{O}(n!)$.
Thus, we expect to design a powerful yet efficient GNN, which can guarantee the expressive power, and significantly reduce the complexity with a minimal loss of generalization capability.

Different from explicitly modeling all $n!$ permutations, we propose to sample a small number of representative permutations to cover all $n(n-1)/2$ pairwise correlations (as shown in Figure \ref{fig:comparison:goal}) by the permutation-sensitive functions.
% The pairwise correlations encoded in the sampled permutations can be further captured by the permutation-sensitive aggregators.
Accordingly, the permutation-invariance is approximated by the invariance to pairwise correlations.
Moreover, we mathematically analyze the complexity of permutation sampling and reduce it from $\mathcal{O}(n \ln n)$ to $\mathcal{O}(n)$ via a well-designed \emph{Permutation Group} (PG).
Based on the proposed permutation sampling strategy, we then devise an aggregation mechanism and theoretically prove that its expressivity is strictly more powerful than the 2-WL test and not less powerful than the 3-WL test.
Thus, our model is capable of significantly reducing the computational complexity while guaranteeing the expressive power.
To the best of our knowledge, our model achieves the lowest time and space complexity among all the GNNs beyond \mbox{2-WL test so far}.

\section{Related Work}

\paragraph{Permutation-Sensitive Graph Neural Networks.}
\citet{loukas2020graph} first analyzes that it is necessary to sacrifice the permutation-invariance and -equivariance of MPNNs to improve their expressive power when nodes lose discriminative attributes.
However, only a few models (GraphSAGE with LSTM aggregators \cite{hamilton2017inductive}, RP with $\pi$-SGD \cite{murphy2019janossy}, CLIP \cite{dasoulas2020coloring}) are permutation-sensitive GNNs.
These studies provide either theoretical proofs or empirical results that their approaches can capture some substructures, especially triangles, which can be served as special cases of our Theorem \ref{thm:triangle}.
Despite their powerful expressivity, the nonlinear complexity of sampling or coloring limits their \mbox{practical application}.

\paragraph{Expressive Power of Graph Neural Networks.}
\citet{xu2019powerful} and \citet{morris2019weisfeiler} first investigate the GNNs' ability to distinguish non-isomorphic graphs and demonstrate that the traditional message-passing paradigm \cite{gilmer2017neural} is at most as powerful as the 2-WL test \cite{weisfeiler1968reduction}, which cannot distinguish some graph pairs like regular graphs with identical attributes.
In order to theoretically improve the expressive power of the 2-WL test, a direct way is to equip nodes with distinguishable attributes, e.g., identifier \cite{murphy2019relational, loukas2020graph}, port numbering \cite{sato2019approximation}, coloring \cite{sato2019approximation, dasoulas2020coloring}, and random feature \cite{sato2021random, abboud2021surprising}.
Another series of researches \cite{morris2019weisfeiler, maron2019invariant, maron2019universality, maron2019provably, keriven2019universal, azizian2021expressive} consider high-order relations to design more powerful GNNs but suffer from high computational complexity when handling high-order tensors and performing global computations on the graph.
% Theoretically more powerful methods use non-local updates, breaking one of the most important inductive bias in Euclidean learning named locality principle \cite{balcilar2021breaking}.
Some pioneering works \cite{de2020natural, thiede2021autobahn} use the automorphism group of local subgraphs to obtain more expressive representations and overcome the problem of global computations, but their pre-processing stages still require solving the NP-hard subgraph isomorphism problem.
Recent studies \cite{sato2019approximation, garg2020generalization, loukas2020graph, tahmasebi2020counting} also characterize the expressive power of GNNs from the perspectives of what they cannot learn.

\paragraph{Leveraging Substructures for Learning Representations.}
Previous efforts mainly focused on the isomorphism tasks, but did little work on understanding their capacity to capture and exploit the graph substructure.
Recent studies \cite{dasoulas2020coloring, chen2020can, de2020natural, vignac2020building, you2021identity, sato2021random, balcilar2021breaking, barcelo2021graph, bouritsas2022improving} show that the expressive power of GNNs is highly related to the local substructures in graphs.
\citet{chen2020can} demonstrate that the substructure counting ability of GNN architectures not only serves as an intuitive theoretical measure of their expressive power but also is highly relevant to practical tasks.
\citet{barcelo2021graph} and \citet{bouritsas2022improving} propose to incorporate some handcrafted subgraph features to improve the expressive power, while they require expert knowledge to select task-relevant features.
Several latest advances \cite{vignac2020building, balcilar2021breaking, bodnar2021weisfeilers, bodnar2021weisfeilerc} have been made to enhance the standard MPNNs by leveraging high-order structural information while retaining the locality of message-passing.
However, the complexity issue has not been satisfactorily solved because they introduce memory/time-consuming context matrices \cite{vignac2020building}, eigenvalue decomposition \cite{balcilar2021breaking}, and lifting transformation \cite{bodnar2021weisfeilers, bodnar2021weisfeilerc} in pre-processing.

\paragraph{\emph{Relations to Our Work.}}
Some crucial differences between related works \cite{hamilton2017inductive, murphy2019relational, dasoulas2020coloring, chen2020can, vignac2020building, sato2021random, balcilar2021breaking, bodnar2021weisfeilers} and ours can be summarized as follows:
(\textit{i}) we propose to design powerful permutation-sensitive GNNs while approximating the property of permutation-invariance, balancing the expressivity and computational efficiency;
(\textit{ii}) our approach realizes the linear complexity of permutation sampling and reaches the theoretical lower bound;
(\textit{iii}) our approach can directly learn substructures from data instead of pre-computing or strategies based on handcrafted structural features.
We also provide detailed discussions in Appendix \ref{sec:GraphSAGE} for \cite{hamilton2017inductive}, \ref{sec:obs_expt} for \cite{murphy2019janossy, murphy2019relational}, and \ref{sec:SCIN} for \cite{bodnar2021weisfeilers, bodnar2021weisfeilerc}.

\section{Designing Powerful Yet Efficient GNNs via Permutation Groups}

In this section, we begin with the analysis of theoretically most powerful but intractable GNNs.
Then, we propose a tractable strategy to achieve linear permutation sampling and significantly reduce the complexity.
Based on this strategy, we design our permutation-sensitive aggregation mechanism via permutation groups.
Furthermore, we mathematically analyze the expressivity of permutation-sensitive GNNs and prove that our proposed model is more powerful than the 2-WL test and not less powerful than the 3-WL test via incidence substructure counting.

\subsection{Preliminaries}
\label{sec:pre}

Let $G = (\mathcal{V}, \mathcal{E}) \in \mathcal{G}$ be a graph with vertex set $\mathcal{V} = \{v_1, v_2, \ldots, v_N\}$ and edge set $\mathcal{E}$, directed or undirected.
Let $\bm A \in \mathbb{R}^{N \times N}$ be the adjacency matrix of $G$.
For a node $ v \in \mathcal{V}$, $d_v$ denotes its degree, i.e., the number of 1-hop neighbors of node $v$, which is equivalent to $n$ in this section for simplicity.
Suppose these $n$ neighboring nodes of the central node $v$ are randomly numbered as $u_1, \ldots, u_n$ (also abbreviated as $1, \ldots, n$ in the following), the set of neighboring nodes is represented as $\mathcal{N}(v)$ (or $S = [n] = \{1, \ldots, n\}$).
Given a set of graphs $\{ G_1, G_2, \ldots, G_M \} \subseteq \mathcal{G}$, each graph $G$ has a label $y_G$.
Our goal is to learn a representation vector $\bm h_G$ of the entire graph $G$ and classify it into the correct category from $C$ classes.
In this paper, we use the normal $G$ to denote a \emph{graph} and the Gothic $\mathfrak{G}$ to denote a \emph{group}.
The necessary backgrounds of graph theory and group theory are attached in Appendixes \ref{sec:graph} and \ref{sec:group}.
The rigorous definition of the $k$-WL test is provided in Appendix \ref{sec:kWL}.

\subsection{Theoretically Most Powerful GNNs}

Relational Pooling (RP) \cite{murphy2019relational} proposes the theoretically most powerful permutation-invariant model by averaging over all permutations of the nodes, which can be formulated as follows:
\begin{equation}
\label{eq:RP}
    \bm h_G = \frac{1}{|\mathcal{S}_N|} \sum_{\pi \in \mathcal{S}_N} {\vec{f} \left( \bm h_{\pi v_1}, \bm h_{\pi v_2}, \cdots , \bm h_{\pi v_N} \right)}
\end{equation}
where $\pi v_i (i = 1, \ldots, N)$ denotes the result of acting $\pi \in \mathcal{S}_N$ on $v_i \in \mathcal{V}$, $\mathcal{S}_N$ is the symmetric group on the set $[N]$ (or $\mathcal{V}$), $\vec{f}$ is a sufficiently expressive (possibly permutation-sensitive) function, $\bm h_{v_i}$ is the feature vector of node $v_i$.

The permutation-sensitive functions, especially sequential models, are capable of modeling the \emph{$k$-ary dependency} \cite{murphy2019janossy, murphy2019relational} among $k$ input nodes.
Meanwhile, the different input node orderings will lead to a total number of $k!$ different $k$-ary dependencies.
These $k$-ary dependencies indicate the relations and help capture the topological connections among the corresponding $k$ nodes, thereby exploiting the substructures within these $k$ nodes to improve the expressive power of GNN models.
For instance, the expressivity of Eq.~\eqref{eq:RP} is mainly attributed to the modeling of all possible \emph{$N$-ary dependencies} (\emph{full-dependencies}) among all $N$ nodes, which can capture all graphs isomorphic to $G$.
However, it is intractable and practically prohibitive to model all permutations ($N!$ $N$-ary dependencies) due to the extremely high computational cost.
Thus, it is necessary to design a tractable strategy to reduce the computational cost while maximally preserving the expressive power.

\subsection{Permutation Sampling Strategy}
\label{sec:strategy}

Intuitively, the simplest way is to replace $N$-ary dependencies with \emph{2-ary dependencies}, i.e., the \emph{pairwise correlations} in Section \ref{sec:intro}.
Moreover, since the inductive bias of locality results in lower complexity on sparse graphs \cite{battaglia2018relational, balcilar2021breaking}, we restrict the permutation-sensitive functions to aggregate information and model the 2-ary dependencies in the 1-hop neighborhoods.
% The message-passing framework applies a common information propagation procedure and learns relationships between a local neighborhood \cite{battaglia2018relational}.
Thus, we will further discuss how to model all 2-ary dependencies between $n$ neighboring nodes with the lowest sampling complexity $\mathcal{O}(n)$.

Suppose $n$ neighboring nodes are arranged as a ring, we define this ring as an \emph{arrangement}.
An initial arrangement can be simply defined as $1 - 2 - \cdots - n - 1$, including an $n$-ary dependency $\{1 - 2 - \cdots - n - 1\}$ and $n$ 2-ary dependencies $\{1-2, 2-3, \cdots, n-1\}$.
Since a permutation adjusts the node ordering in the arrangement, we can use a permutation to generate a new arrangement, which corresponds to a new $n$-ary dependency covering $n$ 2-ary dependencies.
The following theorem provides a lower bound of the number of arrangements to cover all 2-ary dependencies.

\begin{theorem}
\label{thm:arrange}
	Let $n (n \ge 4)$ denote the number of 1-hop neighboring nodes around the central node $v$.
	There are $\lfloor (n-1)/2 \rfloor$ kinds of arrangements in total, satisfying that their corresponding 2-ary dependencies are disjoint.
	Meanwhile, after at least $\lfloor n/2 \rfloor$ arrangements (including the initial one), all 2-ary dependencies have been covered at least once.
\end{theorem}

We first give a sketch of the proof.
Construct a simple undirected graph $G' = (\mathcal{V}', \mathcal{E}')$, where $\mathcal{V}'$ denotes the $n$ neighboring nodes (abbreviated as nodes in the following), and $\mathcal{E}'$ represents an edge set in which each edge indicates the corresponding 2-ary dependency has been covered in some arrangements.
Each arrangement corresponds to a Hamiltonian cycle in graph $G'$.
In addition, we define the following permutation $\sigma$ to generate new arrangements:
\begin{equation}
\label{eq:sigma}\small
\sigma = \left\{
    \begin{aligned}
    & \left( \begin{matrix}
    1 & 2 & 3 & 4 & 5 & \cdots & n-1 & n  \\
    1 & 4 & 2 & 6 & 3 & \cdots & n & n-2  \\
    \end{matrix} \right)
    = \left( 2\ 4\ 6\ \cdots \ n-1\ n\ n-2\ \cdots \ 7\ 5\ 3 \right), n~\text{is odd}, \\
    & \left( \begin{matrix}
    1 & 2 & 3 & 4 & \cdots & n-1 & n  \\
    3 & 1 & 5 & 2 & \cdots & n & n-2  \\
    \end{matrix} \right)
    = \left( 1\ 3\ 5\ \cdots \ n-1\ n\ n-2\ \cdots \ 6\ 4\ 2 \right), n~\text{is even}.
    \end{aligned}
\right.
\end{equation}
After performing the permutation $\sigma$ once, a new arrangement is generated and a Hamiltonian cycle is constructed.
Since every pair of nodes can form a 2-ary dependency, covering all 2-ary dependencies is equivalent to constructing a complete graph $K_n$.
Besides, as a $K_n$ has $n(n-1)/2$ edges and each Hamiltonian cycle has $n$ edges, a $K_n$ can only be constructed with at least $\lceil n(n-1) / 2n \rceil = \lceil (n-1)/2 \rceil = \lfloor n/2 \rfloor$ Hamiltonian cycles.
It can be proved that after performing the permutation $\sigma$ for $\lfloor n/2 \rfloor - 1 = \mathcal{O}(n)$ times in succession (excluding the initial one), all 2-ary dependencies are covered at least once.
Detailed proof of Theorem \ref{thm:arrange} is provided in Appendix \ref{sec:pf_arrange}.

Note that Theorem \ref{thm:arrange} has the constraint $n \ge 4$ because all 2-ary dependencies have already been covered in the initial arrangement when $1 < n < 4$, and there is only a single node when $n = 1$.
If $n = 2,3,4$, $\sigma = \left( \begin{matrix} 1 & 2 \end{matrix} \right),$ $\left( \begin{matrix} 2 & 3 \end{matrix} \right), \left( \begin{matrix} 1 & 3 & 4 & 2 \end{matrix} \right)$, respectively (the case of $n = 1$ is trivial).
Thus the permutation $\sigma$ defined in Theorem \ref{thm:arrange} is available for an arbitrary $n$, while Eq.~\eqref{eq:sigma} shows the general case with a large $n$.

\begin{figure}
    \centering
    \subfigure[$n = 5$ \qquad \qquad \qquad \quad \ \ \ (b) $n = 6$]{
	\includegraphics[width=0.9\linewidth]{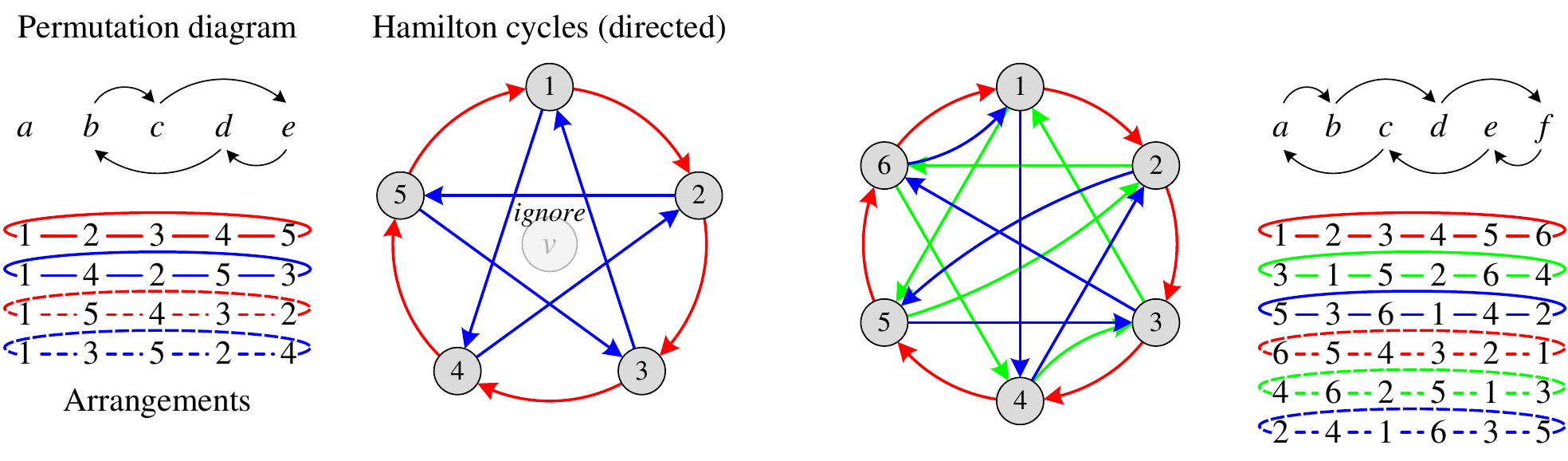}
	}
	\addtocounter{subfigure}{+1}
    \subfigure[$n = 7$ \qquad \qquad \qquad \quad \ \ \ (d) $n = 8$]{
	\includegraphics[width=0.9\linewidth]{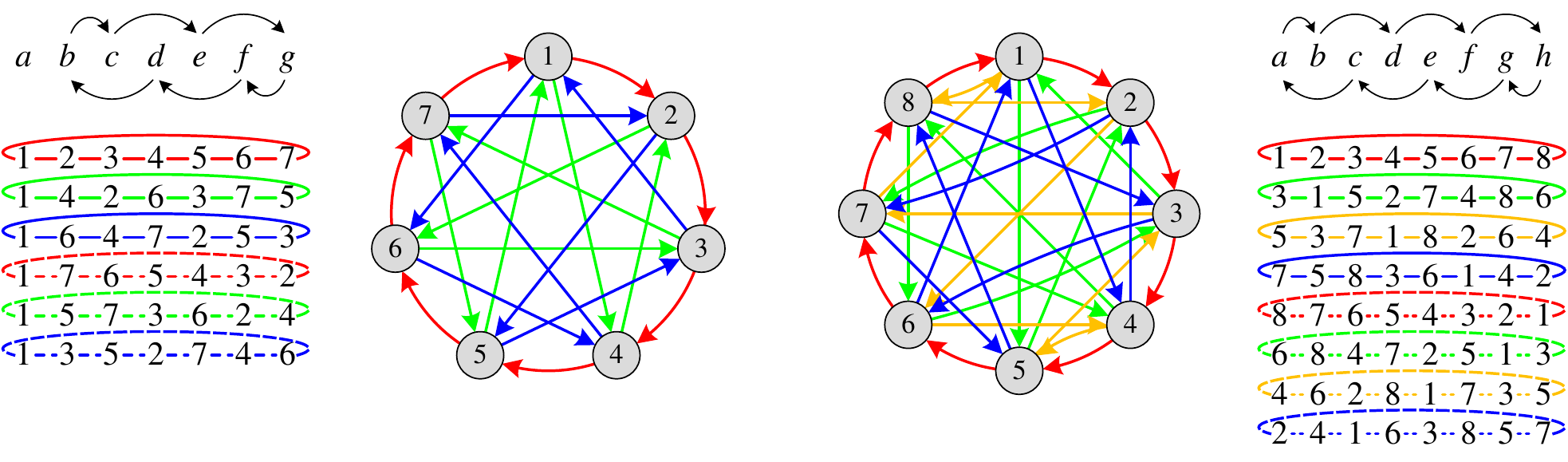}
	}
    \caption{Modeling all pairwise correlations between $n$ neighboring nodes via permutations.
    Subfigures (a) to (d) characterize the cases when $n = 5~\text{to}~8$ (ignoring the central node $v$).
    The monochrome permutation diagram illustrates the mapping process of permutation $\sigma$, where the directed arc $a \to b$ indicates that moving $a$ to the original position of $b$.
    All arrangements generated by $\sigma^i$ are shown in color below the diagram.
    The first and the last $\lfloor n/2 \rfloor$ arrangements are marked with solid and dashed lines, respectively.
    Solid and dashed lines with the same color indicate that they correspond to a pair of bi-directional Hamiltonian cycles.
    Only the Hamiltonian cycles corresponding to the first $\lfloor n/2 \rfloor$ arrangements are displayed for clarity.
    For a further explanation and the relationships among Theorem \ref{thm:arrange}, Lemma \ref{thm:cyclic}, Corollary \ref{thm:action}, Figure \ref{fig:complete}, and Eq.~\eqref{eq:agg}, please refer to Appendix \ref{sec:illustration} and Figure \ref{fig:illustration}.}
    \label{fig:complete}
\end{figure}

According to the ordering of $n$ neighboring nodes in the arrangement, we can apply a permutation-sensitive function to model an $n$-ary dependency among these $n$ nodes while covering $n$ 2-ary dependencies.
Since the input orderings $a \to b$ and $b \to a$ lead to different results in the permutation-sensitive function, these dependencies and the corresponding Hamiltonian cycles (the solid arrows in Figure \ref{fig:complete}) are modeled in a directed manner.
We continue performing the permutation $\sigma$ for $\lfloor n/2 \rfloor$ times successively to get additional $\lfloor n/2 \rfloor$ arrangements (the dashed lines in Figure \ref{fig:complete}) and reversely directed Hamiltonian cycles (not shown in Figure \ref{fig:complete}).
After the bi-directional modeling, edges in Hamiltonian cycles are transformed into undirected edges.
% For the difference between bi-directional and bi-directed, please refer to \url{https://www.quora.com/What-is-the-difference-between-undirected-graph-and-bi-directed-graph-Doesn\%E2\%80\%99t-an-undirected-edge-between-two-nodes-symbolize-bi-direction}.
Figure \ref{fig:complete} briefly illustrates the above process when $n = 5~\text{to}~8$.
In conclusion, all 2-ary dependencies can be modeled in an undirected manner by the tailored permutations.
The number of permutations is $n$ if $n$ is even and $(n-1)$ if $n$ is odd, ensuring the linear sampling complexity $\mathcal{O}(n)$.

In fact, all permutations above form a permutation group.
In order to incorporate the strategy proposed by Theorem \ref{thm:arrange} into the aggregation process of GNN, we propose to use the permutation group and group action, defined as follows.

\begin{lemma}
\label{thm:cyclic}
	For the permutation $\sigma$ of $n$ indices, $\mathfrak{G} = \{ e, \sigma, \sigma^2, \ldots, \sigma^{n-2} \}$ is a permutation group isomorphic to the cyclic group $\mathbb{Z}_{n-1}$ if $n$ is odd.
	And $\mathfrak{G} = \{ e, \sigma, \sigma^2, \ldots, \sigma^{n-1} \}$ is a permutation group isomorphic to the cyclic group $\mathbb{Z}_{n}$ if $n$ is even.
\end{lemma}
\begin{corollary}
\label{thm:action}
	The map $\alpha: \mathfrak{G} \times S \to S$ denoted by $(g,s) \mapsto gs$ is a group action of $\mathfrak{G}$ on $S$.
\end{corollary}

To better illustrate the results of Lemma \ref{thm:cyclic} and Corollary \ref{thm:action}, the detailed discussion and diagram are attached in Appendix \ref{sec:diagram}.
Next, we apply the permutation group and group action to design our permutation-sensitive aggregation mechanism.

\subsection{Network Architecture}
\label{sec:arch}

Without loss of generality, we apply the widely-used Recurrent Neural Networks (RNNs) as the permutation-sensitive function to model the dependencies among neighboring nodes.
Let the group elements (i.e., permutations) in $\mathfrak{G}$ act on $S$, our proposed strategy in Section \ref{sec:strategy} is formulated as:
\begin{equation}
\label{eq:agg}
    \bm h_{v}^{(k)} = \sum_{g \in \mathfrak{G}} {\text{RNN} \left( \bm h_{g u_1}^{(k-1)}, \bm h_{g u_2}^{(k-1)}, \cdots, \bm h_{g u_n}^{(k-1)}, \bm h_{g u_1}^{(k-1)} \right)} + \bm W_{\text{self}}^{(k-1)} \bm h_{v}^{(k-1)}, u_{1:n} \in \mathcal{N}(v)
\end{equation}
where $gu_i (i = 1, \ldots, n)$ denotes the result of acting $g \in \mathfrak{G}$ on $u_i \in S$, and $\bm h_{v}^{(k)} \in \mathbb{R}^{d_k}$ is the feature vector of central node $v$ at the $k$-th layer.
We provide more discussion on the groups and model variants in Appendixes \ref{sec:2-ary} and \ref{sec:variant}.
Eq.~\eqref{eq:agg} takes advantage of the locality and permutation group $\mathfrak{G}$ to simplify the group actions in Eq.~\eqref{eq:RP}, which acts the symmetric group $\mathcal{S}_N$ on vertex set $\mathcal{V}$, thereby avoiding the complete modeling of $N!$ permutations.
Meanwhile, Eq.~\eqref{eq:agg} models all 2-ary dependencies and achieves the invariance to 2-ary dependencies.
Thus, we can conclude that Eq.~\eqref{eq:agg} realizes the efficient approximation of permutation-invariance with low complexity.
In practice, we merge the central node $v$ into RNN for simplicity:
\begin{equation}
    \bm h_{v}^{(k)} = \sum_{g \in \mathfrak{G}} {\text{RNN} \left( \bm h_{v}^{(k-1)}, \bm h_{g u_1}^{(k-1)}, \bm h_{g u_2}^{(k-1)}, \cdots, \bm h_{g u_n}^{(k-1)}, \bm h_{v}^{(k-1)} \right)}, u_{1:n} \in \mathcal{N}(v)
\end{equation}
Then, we apply a READOUT function (e.g., SUM$(\cdot)$) to obtain the graph representation $\bm h_{G}^{(k)}$ at the $k$-th layer and combine representations learned by different layers to get the score $\bm s$ for classification:
\begin{equation}
\label{eq:final}
    \bm h_{G}^{(k)} = \sum_{v \in \mathcal{V}} {\bm h_{v}^{(k)}}, \quad
    \bm s = \sum_{k} {\bm W^{(k)} \bm h_{G}^{(k)}}
\end{equation}
here $\bm W^{(k)} \in \mathbb{R}^{C \times d_k}$ represents a learnable scoring matrix for the $k$-th layer.
Finally, we input score $\bm s$ to the softmax function and obtain the predicted class of graph $G$.

\paragraph{Complexity.}
We briefly analyze the computational complexity of Eq.~\eqref{eq:agg}.
Suppose the input and output dimensions are both $c$ for each layer, let $\Delta$ denote the maximum degree of graph $G$.
In the worst-case scenario, Eq.~\eqref{eq:agg} requires summing over $\Delta$ terms processed in a serial manner.
Since there is no interdependence between these $\Delta$ terms, they can also be computed in a parallel manner with the time complexity of $\Theta(\Delta c^2)$ (caused by RNN computation), while sacrificing the memory to save time.
Let $M$ denote the number of edges.
Table \ref{tab:complexity} compares our approach with other powerful GNNs on the per-layer space and time complexity.
The results of baselines are taken from \citet{vignac2020building}.
Since the complexity analysis of GraphSAGE \cite{hamilton2017inductive}, MPSN \cite{bodnar2021weisfeilers}, and CWN \cite{bodnar2021weisfeilerc} involves many other notations, we analyze GraphSAGE in Appendix \ref{sec:GraphSAGE}, and MPSN and CWN in Appendix \ref{sec:SCIN}.
In a nutshell, our approach theoretically outperforms other powerful GNNs in terms of time and space complexity, even being on par with MPNN.

\begin{table}\small
\begin{minipage}[t]{0.52\linewidth}
    \centering
    \caption{Memory and time complexity per layer.}
    \label{tab:complexity}
    \begin{tabular}{lcc}
    \toprule
    Model       & Memory                & Time complexity \\
    \midrule
    GIN \cite{xu2019powerful}
                & $\Theta(Nc)$          & $\Theta(Mc + Nc^2)$       \\
    MPNN \cite{gilmer2017neural}
                & $\Theta(Nc)$          & $\Theta(Mc^2)$            \\
    Fast SMP \cite{vignac2020building}
                & $\Theta(N^2c)$        & $\Theta(MNc + N^2c^2)$    \\
    SMP \cite{vignac2020building}
                & $\Theta(N^2c)$        & $\Theta(MNc^2)$           \\
    PPGN \cite{maron2019provably}
                & $\Theta(N^2c)$        & $\Theta(N^3c + N^2c^2)$   \\
    3-WL \cite{morris2019weisfeiler}
                & $\Theta(N^3c)$        & $\Theta(N^4c + N^3c^2)$   \\
    \midrule
    Ours (serial)
                & $\Theta(Nc)$          & $\Theta(N\Delta^2c^2)$    \\
    Ours (parallel)
                & $\Theta(N\Delta c)$   & $\Theta(N\Delta c^2)$     \\
    \bottomrule
    \end{tabular}
\end{minipage}
\begin{minipage}[t]{0.46\linewidth}
    \centering
    \caption{Results (measured by MAE) on incidence triangle counting.}
    \label{tab:substruc}
    \begin{tabular}{lcc}
    \toprule
    Model   & \makecell[c]{Erd\H{o}s-R{\'e}nyi\\random graph}
            & \makecell[c]{Random\\regular graph}       \\
    \midrule
    GCN \cite{kipf2017semi}
            & 0.599 $\pm$ 0.006     & 0.500 $\pm$ 0.012 \\
    SAGE \cite{hamilton2017inductive}
            & 0.118 $\pm$ 0.005     & 0.127 $\pm$ 0.011 \\
    GIN \cite{xu2019powerful}
            & 0.219 $\pm$ 0.016     & 0.342 $\pm$ 0.005 \\
    rGIN \cite{sato2021random}
            & 0.194 $\pm$ 0.009     & 0.325 $\pm$ 0.006 \\
    RP \cite{murphy2019relational}
            & 0.058 $\pm$ 0.006     & 0.161 $\pm$ 0.003 \\
    LRP \cite{chen2020can}
            & 0.023 $\pm$ 0.011     & 0.037 $\pm$ 0.019 \\
    \midrule
    PG-GNN
            & \textbf{0.019 $\pm$ 0.002}    & \textbf{0.027 $\pm$ 0.001}    \\
    \bottomrule
    \end{tabular}
\end{minipage}
\end{table}

\subsection{Expressivity Analysis}
\label{sec:express}

In this subsection, we theoretically analyze the expressive power of a typical category of permutation-sensitive GNNs, i.e., GNNs with RNN aggregators (Theorem \ref{thm:triangle}), and that of our proposed PG-GNN (Proposition \ref{thm:express}).
We begin with GIN \cite{xu2019powerful}, which possesses the equivalent expressive power as the 2-WL test \cite{xu2019powerful, azizian2021expressive}.
In fact, the variants of GIN can be recovered by GNNs with RNN aggregators (see Appendix \ref{sec:pf_express} for details), which implies that this category of permutation-sensitive GNNs can be at least as powerful as the 2-WL test.
Next, we explicate why they go beyond the 2-WL test from the perspective of substructure counting.

Triangular substructures are rich in various networks, and counting triangles is an important task in network analysis \cite{al2018triangle}.
For example, in social networks, the formation of a triangle indicates that two people with a common friend will also become friends \cite{mitzenmacher2017probability}.
A triangle $\triangle u_ivu_j$ is \emph{incident} to the node $v$ if $u_i$ and $u_j$ are adjacent and node $v$ is their common neighbor.
We define the triangle $\triangle u_ivu_j$ as an \emph{incidence triangle} over node $v$ (also $u_i$ and $u_j$), and denote the number of incidence triangles over node $v$ as $\tau_v$.
Formally, the number of incidence triangles over each node in an \textbf{undirected} graph can be calculated as follows (proof and discussion for the directed graph are provided in \mbox{Appendix \ref{sec:triangle}}):
\begin{equation}
\label{eq:triangle}
    \bm \tau = \frac{1}{2} \bm A^2 \odot \bm A \cdot \bm 1_N
\end{equation}
where $\bm \tau \in \mathbb{R}^N$ and its $i$-th element $\tau_i$ represents the number of incidence triangles over node $i$, $\odot$ denotes element-wise product (i.e., Hadamard product), $\bm 1_N = (1, 1, \cdots, 1)^\top \in \mathbb{R}^N$ is a sum vector.

Besides the WL-test, the capability of counting graph substructures also characterizes the expressive power of GNNs \cite{chen2020can}.
Thus, we verify the expressivity of permutation-sensitive GNNs by evaluating their abilities to count triangles.

\begin{theorem}
\label{thm:triangle}
	Let $x_v, \forall v \in \mathcal{V}$ denote the feature inputs on graph $G = (\mathcal{V}, \mathcal{E})$, and $\mathbb{M}$ be a general GNN model with RNN aggregators.
	Suppose that $x_v$ is initialized as the degree $d_v$ of node $v$, and each node is distinguishable.
	For any $0 < \epsilon \le 1/8$ and $0 < \delta < 1$, there exists a parameter setting $\Theta$ for $\mathbb{M}$ so that after $\mathcal{O} \left( \frac{d_v(2d_v+\tau_v)\mathfrak{t}} {d_v+\tau_v} \right)$ samples,
	\begin{equation*}
	\Pr \left( \left| \frac{z_v}{\tau_v} - 1 \right| \le \epsilon \right) \ge 1-\delta, \forall v \in \mathcal{V},
	\end{equation*}
	where $z_v \in \mathbb{R}$ is the final output value generated by $\mathbb{M}$ and $\tau_v$ is the number of incidence triangles.
\end{theorem}

Detailed proof can be found in Appendix \ref{sec:incidence}.
Theorem \ref{thm:triangle} concludes that, if the input node features are node degrees and nodes are distinguishable, there exists a parameter setting for a \emph{general} GNN with RNN aggregators such that it can approximate the number of incidence triangles to arbitrary precision for every node.
Since 2-WL and MPNNs cannot count triangles \cite{chen2020can}, we conclude that this category of permutation-sensitive GNNs is more powerful.
% This theorem partially answers why \emph{general} permutation-sensitive GNNs are more powerful, and we hope to draw more attention to the permutation-sensitive GNNs in the community.
However, the required samples are related to $\tau_v$ and proportional to the mixing time $\mathfrak{t}$ (see Appendix \ref{sec:incidence}), leading to a practically prohibitive aggregation complexity.
Many existing permutation-sensitive GNNs like GraphSAGE with LSTM and RP with $\pi$-SGD suffer from this issue (see Appendixes \ref{sec:GraphSAGE} and \ref{sec:obs_expt} for more discussion).

On the contrary, our approach can estimate the number of incidence triangles in linear sampling complexity $\mathcal{O}(n) = \mathcal{O}(d_v)$.
According to the definition of incidence triangles and the fact that they always appear within $v$'s 1-hop neighborhood, we know that \emph{the number of connections between the central node $v$'s neighboring nodes is equivalent to the number of incidence triangles over $v$}.
% If a pattern is present in the graph, it can always be found in a sufficiently large local neighborhood, or egonet, of some node in the graph \cite{chen2020can}.
Meanwhile, Theorem \ref{thm:arrange} and Eq.~\eqref{eq:agg} ensure that all 2-ary dependencies between $n$ neighboring nodes are modeled with $\mathcal{O}(n)$ sampling complexity.
These dependencies capture the information of whether two neighboring nodes are connected, thereby estimating the number of connections and counting incidence triangles in linear sampling complexity.

Recently, \citet{balcilar2021breaking} claimed that the trace (tr) and Hadamard product ($\odot$) operations are crucial requirements to go further than 2-WL to reach 3-WL from the perspective of Matrix Language \cite{brijder2019expressive, geerts2021expressive}.
In fact, for any two neighbors $u_i$ and $u_j$ of the central node $v$, the locality and 2-ary dependency of Eq.~\eqref{eq:agg} introduce the information of $\bm A^2$ (i.e., $u_i-v-u_j$) and $\odot \bm A$ (i.e., $u_i$ $-\!\!\!\!?$ $u_j$), respectively.
Thus Eq.~\eqref{eq:agg} can mimic Eq.~\eqref{eq:triangle} to count incidence triangles.
Moreover, we also prove that $\bm 1_N^\top \bm \tau = \frac{1}{2} \text{tr}(\bm A^3)$ (see Appendix \ref{sec:triangle} for details), which indicates that PG-GNN can realize the trace (tr) operation when we use SUM$(\cdot)$ or MEAN$(\cdot)$ (i.e., $\bm 1_N$) as the graph-level READOUT function.
Note that even though MPNNs and 2-WL test are equipped with distinguishable attributes, they still have difficulty performing triangle counting since they cannot implement the trace or Hadamard product operations \cite{balcilar2021breaking}.

Beyond the incidence triangle, we can also leverage 2-ary dependencies of $u_i$ $-\!\!\!\!?$ $u_j$, $u_i$ $-\!\!\!\!?$ $u_k$, and \mbox{$u_j$ $-\!\!\!\!?$ $u_k$} to discover the \emph{incidence 4-clique} $\diamondplus vu_iu_ju_k$, which is completely composed of triangles and only appears within $v$'s 1-hop neighborhood.
In this way, the expressive power of PG-GNN can be further improved by its capability of counting incidence 4-cliques.
As illustrated in Figure \ref{fig:SRG}, these incidence 4-cliques help distinguish some pairs of non-isomorphic strongly regular graphs while the 3-WL test fails.
Consequently, the expressivity of our model is guaranteed to be not less powerful than 3-WL\footnote{``A is no/not less powerful than B'' means that there exists a pair of non-isomorphic graphs such that A can distinguish but B cannot.
The terminology ``no/not less powerful'' used here follows the standard definition in the literature \cite{chen2020can, bodnar2021weisfeilers, bodnar2021weisfeilerc, zhao2022stars}.}.
% For confused readers, please refer to \url{https://openreview.net/forum?id=uVPZCMVtsSG&noteId=JmqD7VZ2Qxt} for more details.

From the analysis above, we confirm the expressivity of PG-GNN as follows.
The strict proof and more detailed discussion on PG-GNN and 3-WL are provided in Appendix \ref{sec:pf_express}.

\begin{proposition}
\label{thm:express}
	PG-GNN is strictly more powerful than the 2-WL test and not less powerful than the 3-WL test.
\end{proposition}

\section{Experiments}
\label{sec:expt}

In this section, we evaluate PG-GNN on multiple synthetic and real-world datasets from a wide range of domains.
Dataset statistics and details are presented in Appendix \ref{sec:dataset}.
The hyper-parameter search space and final hyper-parameter configurations are provided in Appendix \ref{sec:hyper}.
Computing infrastructures can be found in Appendix \ref{sec:infra}.
The code is publicly available at \url{https://github.com/zhongyu1998/PG-GNN}.

\subsection{Counting Substructures in Random Graphs}

We conduct synthetic experiments of counting incidence substructures (triangles and 4-cliques) on two types of random graphs: Erd\H{o}s-R{\'e}nyi random graphs and random regular graphs \cite{chen2020can}.
The incidence substructure counting task is designed on the node level, which is more rigorous than traditional graph-level counting tasks.
Table \ref{tab:substruc} summarizes the results measured by Mean Absolute Error (MAE, lower is better) for incidence triangle counting.
We report the average and standard deviation of testing MAEs over 5 runs with 5 different seeds.
In addition, the testing MAEs of PG-GNN on ER and random regular graphs are 0.029 $\pm$ 0.002 and 0.023 $\pm$ 0.001 for incidence 4-clique counting, respectively.
Overall, the negligible MAEs of our model support our claim that PG-GNN is powerful enough for counting incidence triangles and 4-cliques.

Another phenomenon is that permutation-sensitive GNNs consistently outperform permutation-invariant GNNs on substructure counting tasks.
This indicates that permutation-sensitive GNNs are capable of learning these substructures directly from data, without explicitly assigning them as node features, but the permutation-invariant counterparts like GCN and GIN fail.
Therefore, permutation-sensitive GNNs can implicitly leverage the information of characteristic substructures in representation learning and thus benefit real-world tasks in practical scenarios.

\begin{table}\small
    \centering
    \caption{Results (measured by accuracy: \%) on TUDataset.}
    \label{tab:tud}
    \begin{tabular}{lccccc}
    \toprule
    Model
    & PROTEINS          & NCI1
    & IMDB-B            & IMDB-M            & COLLAB            \\
    \midrule
    WL \cite{shervashidze2011weisfeiler}
    & 75.0 $\pm$ 3.1    & \textbf{86.0 $\pm$ 1.8}
    & 73.8 $\pm$ 3.9    & 50.9 $\pm$ 3.8    & 78.9 $\pm$ 1.9    \\
    DGCNN \cite{zhang2018end}
    & 75.5 $\pm$ 0.9    & 74.4 $\pm$ 0.5
    & 70.0 $\pm$ 0.9    & 47.8 $\pm$ 0.9    & 73.8 $\pm$ 0.5    \\
    IGN \cite{maron2019invariant}
    & 76.6 $\pm$ 5.5    & 74.3 $\pm$ 2.7
    & 72.0 $\pm$ 5.5    & 48.7 $\pm$ 3.4    & 78.4 $\pm$ 2.5    \\
    GIN \cite{xu2019powerful}
    & 76.2 $\pm$ 2.8    & 82.7 $\pm$ 1.7
    & 75.1 $\pm$ 5.1    & 52.3 $\pm$ 2.8    & 80.2 $\pm$ 1.9    \\
    PPGN \cite{maron2019provably}
    & \textbf{77.2 $\pm$ 4.7}               & 83.2 $\pm$ 1.1
    & 73.0 $\pm$ 5.8    & 50.5 $\pm$ 3.6    & 80.7 $\pm$ 1.7    \\
    CLIP \cite{dasoulas2020coloring}
    & 77.1 $\pm$ 4.4    & N/A
    & 76.0 $\pm$ 2.7    & 52.5 $\pm$ 3.0    & N/A               \\
    NGN \cite{de2020natural}
    & 71.7 $\pm$ 1.0    & 82.7 $\pm$ 1.4
    & 74.8 $\pm$ 2.0    & 51.3 $\pm$ 1.5    & N/A               \\
    WEGL \cite{kolouri2021wasserstein}
    & 76.5 $\pm$ 4.2    & N/A
    & 75.4 $\pm$ 5.0    & 52.3 $\pm$ 2.9    & 80.6 $\pm$ 2.0    \\
    SIN \cite{bodnar2021weisfeilers}
    & 76.5 $\pm$ 3.4    & 82.8 $\pm$ 2.2
    & 75.6 $\pm$ 3.2    & 52.5 $\pm$ 3.0    & N/A               \\
    CIN \cite{bodnar2021weisfeilerc}
    & 77.0 $\pm$ 4.3    & 83.6 $\pm$ 1.4
    & 75.6 $\pm$ 3.7    & 52.7 $\pm$ 3.1    & N/A               \\
    \midrule
    PG-GNN (Ours)
    & 76.8 $\pm$ 3.8    & 82.8 $\pm$ 1.3
    & \textbf{76.8 $\pm$ 2.6}  & \textbf{53.2 $\pm$ 3.6}  & \textbf{80.9 $\pm$ 0.8}\\
    \bottomrule
    \end{tabular}
\end{table}

\subsection{Real-World Benchmarks}

\paragraph{Datasets.}
We evaluate our model on 7 real-world datasets from various domains.
PROTEINS and NCI1 are bioinformatics datasets; IMDB-BINARY, IMDB-MULTI, and COLLAB are social network datasets.
They are all popular graph classification tasks from the classical TUDataset \cite{morris2020tudataset}.
We follow \citet{xu2019powerful} to create the input features for each node.
More specifically, the input node features of bioinformatics graphs are categorical node labels, and the input node features of social networks are node degrees.
All the input features are encoded in a one-hot manner.
In addition, MNIST is a computer vision dataset for the graph classification task, and ZINC is a chemistry dataset for the graph regression task.
They are both modern benchmark datasets, and we obtain the features from the original paper \cite{dwivedi2020benchmarking}, but \emph{do not} take edge features into account.
We summarize the statistics of all 7 real-world datasets in Table \ref{tab:realstats}, and more details about these datasets can be found in Appendix \ref{sec:dataset}.

\paragraph{Evaluations.}
For TUDataset, we follow the same data split and evaluation protocol as \citet{xu2019powerful}.
We perform 10-fold cross-validation with random splitting and report our results (the average and standard deviation of testing accuracies) at the epoch with the best average accuracy across the 10 folds.
For MNIST and ZINC, we follow the same data splits and evaluation metrics as \citet{dwivedi2020benchmarking}, please refer to Appendix \ref{sec:dataset} for more details.
The experiments are performed over 4 runs with 4 different seeds, and we report the average and standard deviation of testing results.

\paragraph{Baselines.}
We compare our PG-GNN with multiple state-of-the-art baselines:
Weisfeiler-Lehman Graph Kernels (WL) \cite{shervashidze2011weisfeiler},
% Graph Convolutional Network (GCN) \cite{kipf2017semi},
Graph SAmple and aggreGatE (GraphSAGE) \cite{hamilton2017inductive},
Gated Graph ConvNet (GatedGCN) \cite{bresson2017residual},
Deep Graph Convolutional Neural Network (DGCNN) \cite{zhang2018end},
3-WL-GNN \cite{morris2019weisfeiler},
Invariant Graph Network (IGN) \cite{maron2019invariant},
Graph Isomorphism Network (GIN) \cite{xu2019powerful},
% Relational Pooling (RP) \cite{murphy2019relational},
Provably Powerful Graph Network (PPGN) \cite{maron2019provably},
Ring-GNN \cite{chen2019equivalence},
Colored Local Iterative Procedure (CLIP) \cite{dasoulas2020coloring},
Natural Graph Network (NGN) \cite{de2020natural},
(Deep-)Local Relation Pooling (LRP) \cite{chen2020can},
Principal Neighbourhood Aggregation (PNA) \cite{corso2020principal},
% Structural Message-Passing (SMP) \cite{vignac2020building},
% GIN with random features (rGIN) \cite{sato2021random},
Wasserstein Embedding for Graph Learning (WEGL) \cite{kolouri2021wasserstein},
Simplicial Isomorphism Network (SIN) \cite{bodnar2021weisfeilers},
and Cell Isomorphism Network (CIN) \cite{bodnar2021weisfeilerc}.

\paragraph{Results and Analysis.}
Tables \ref{tab:tud} and \ref{tab:bench} present a summary of the results.
The results of baselines in Table \ref{tab:tud} are taken from their original papers, except WL taken from \citet{xu2019powerful}, and IGN from \citet{maron2019provably} for preserving the same evaluation protocol.
The results of baselines in Table \ref{tab:bench} are taken from \citet{dwivedi2020benchmarking}, except PPGN and Deep-LRP are taken from \citet{chen2020can}, and PNA from \citet{corso2020principal}.
% If a model has multiple variants, we report the best result among all variants.
Obviously, our model achieves outstanding performance on most datasets, even outperforming competitive baselines by a considerable margin.

From Tables \ref{tab:tud} and \ref{tab:bench}, we notice that our model significantly outperforms other approaches on all social network datasets, but slightly underperforms main baselines on molecular datasets such as NCI1 and ZINC.
Recall that in Section \ref{sec:express}, we demonstrate that our model is capable of estimating the number of incidence triangles.
The capability of counting incidence triangles benefits our model on graphs with many triangular substructures, e.g., social networks.
However, triangles rarely exist in chemical compounds (verified in Table \ref{tab:realstats}) due to their instability in the molecular structures.
Thus our model achieves sub-optimal performance on molecular datasets.
Suppose we extend the 1-hop neighborhoods to 2-hop (even $k$-hop) in Eq.~\eqref{eq:agg}.
In that case, our model will exploit more sophisticated substructures such as pentagon (cyclopentadienyl) and hexagon (benzene ring), which will benefit tasks on molecular graphs but increase the complexity.
Thus, we leave it to future work.

\begin{table}\small
    \centering
    \caption{Results and running times on MNIST and ZINC.}
    \label{tab:bench}
    \begin{tabular}{lcccc}
    \toprule
    \multirow{2}{*}{\normalsize{Model}}
    & \multicolumn{2}{c}{MNIST}
    & \multicolumn{2}{c}{ZINC}  \\
    \cmidrule(lr){2-3} \cmidrule(lr){4-5}
    & Accuracy $\uparrow$ & Time / Epoch & MAE $\downarrow$ & Time / Epoch \\
    \midrule
    GraphSAGE \cite{hamilton2017inductive}
    & 97.31 $\pm$ 0.10  & 113.12s   & 0.468 $\pm$ 0.003     & 3.74s     \\
    GatedGCN \cite{bresson2017residual}
    & 97.34 $\pm$ 0.14  & 128.79s   & 0.435 $\pm$ 0.011     & 5.76s     \\
    GIN \cite{xu2019powerful}
    & 96.49 $\pm$ 0.25  & 39.22s    & 0.387 $\pm$ 0.015     & 2.29s     \\
    3-WL-GNN \cite{morris2019weisfeiler}
    & 95.08 $\pm$ 0.96  & 1523.20s  & 0.407 $\pm$ 0.028     & 286.23s   \\
    Ring-GNN \cite{chen2019equivalence}
    & 91.86 $\pm$ 0.45  & 2575.99s  & 0.512 $\pm$ 0.023     & 327.65s   \\
    PPGN \cite{maron2019provably}
    & N/A               & N/A       & 0.256 $\pm$ 0.054     & 334.69s   \\
    Deep-LRP \cite{chen2020can}
    & N/A               & N/A & \textbf{0.223 $\pm$ 0.008}  & 72s       \\
    PNA \cite{corso2020principal}
    & 97.41 $\pm$ 0.16  & N/A       & 0.320 $\pm$ 0.032     & N/A       \\
    \midrule
    PG-GNN (Ours)
    & \textbf{97.51 $\pm$ 0.07} & 82.60s & 0.282 $\pm$ 0.011 & 6.92s    \\
    \bottomrule
    \end{tabular}
\end{table}

\subsection{Running Time Analysis}

As discussed above, compared to other powerful GNNs, one of the most important advantages of PG-GNN is efficiency.
To evaluate, we compare the average running times between PG-GNN and baselines on two large-scale benchmarks, MNIST and ZINC.
Table \ref{tab:bench} also presents the average running times per epoch for various models.
% The results of baselines are taken from \citet{dwivedi2020benchmarking}, except PPGN and Deep-LRP are taken from \citet{chen2020can}.
As shown in Table \ref{tab:bench}, PG-GNN is significantly faster than other powerful baselines, even on par with several variants of MPNNs.
Thus, we can conclude that our approach outperforms other powerful GNNs in terms of time complexity.
We also provide memory cost analysis in Tables \ref{tab:RAM} and \ref{tab:memory}, please refer to Appendix \ref{sec:memory} for more details.

\section{Conclusion and Future Work}
\label{sec:conclusion}

In this work, we devise an efficient permutation-sensitive aggregation mechanism via permutation groups, capturing pairwise correlations between neighboring nodes while ensuring linear sampling complexity.
We throw light on the reasons why permutation-sensitive functions can improve GNNs' expressivity.
% We point out the limitations of existing permutation-invariant GNNs and provide a deeper insight into understanding their behavior.
Moreover, we propose to approximate the property of permutation-invariance to significantly reduce the complexity with a minimal loss of generalization capability.
In conclusion, we take an important step forward to better understand the permutation-sensitive GNNs.

However, Eq.~\eqref{eq:agg} only models a small portion of $n$-ary dependencies while covering all 2-ary dependencies.
Although these 2-ary dependencies are invariant to an arbitrary permutation, the invariance to higher-order dependencies may not be guaranteed.
It would be interesting to extend the 1-hop neighborhoods to 2-hop (even $k$-hop) in Eq.~\eqref{eq:agg}, thereby completely modeling higher-order dependencies and exploiting more sophisticated substructures, which is left for future work.

\section*{Acknowledgements}

We would like to express our sincere gratitude to the anonymous reviewers for their insightful comments and constructive feedback, which helped us polish this paper better.
We thank Suixiang Gao, Xing Xie, Shiming Xiang, Zhiyong Liu, and Hao Zhang for their helpful discussions and valuable suggestions.
This work was supported in part by the National Natural Science Foundation of China (61976209, 62020106015), CAS International Collaboration Key Project (173211KYSB20190024), and Strategic Priority Research Program of CAS (XDB32040000).

\bibliographystyle{unsrtnat}
\bibliography{neurips_2021}

\begin{thebibliography}{71}
\providecommand{\natexlab}[1]{#1}
\providecommand{\url}[1]{\texttt{#1}}
\expandafter\ifx\csname urlstyle\endcsname\relax
  \providecommand{\doi}[1]{doi: #1}\else
  \providecommand{\doi}{doi: \begingroup \urlstyle{rm}\Url}\fi

\bibitem[Murphy et~al.(2019{\natexlab{a}})Murphy, Srinivasan, Rao, and
  Ribeiro]{murphy2019relational}
Ryan Murphy, Balasubramaniam Srinivasan, Vinayak Rao, and Bruno Ribeiro.
\newblock Relational pooling for graph representations.
\newblock In \emph{Proceedings of the 36th International Conference on Machine
  Learning}, pages 4663--4673, 2019{\natexlab{a}}.

\bibitem[Duvenaud et~al.(2015)Duvenaud, Maclaurin, Aguilera-Iparraguirre,
  G{\'o}mez-Bombarelli, Hirzel, Aspuru-Guzik, and
  Adams]{duvenaud2015convolutional}
David Duvenaud, Dougal Maclaurin, Jorge Aguilera-Iparraguirre, Rafael
  G{\'o}mez-Bombarelli, Timothy Hirzel, Al{\'a}n Aspuru-Guzik, and Ryan~P
  Adams.
\newblock Convolutional networks on graphs for learning molecular fingerprints.
\newblock In \emph{Advances in Neural Information Processing Systems}, pages
  2224--2232, 2015.

\bibitem[Kipf and Welling(2017)]{kipf2017semi}
Thomas~N Kipf and Max Welling.
\newblock Semi-supervised classification with graph convolutional networks.
\newblock In \emph{Proceedings of the 5th International Conference on Learning
  Representations}, 2017.

\bibitem[Gilmer et~al.(2017)Gilmer, Schoenholz, Riley, Vinyals, and
  Dahl]{gilmer2017neural}
Justin Gilmer, Samuel~S Schoenholz, Patrick~F Riley, Oriol Vinyals, and
  George~E Dahl.
\newblock Neural message passing for quantum chemistry.
\newblock In \emph{Proceedings of the 34th International Conference on Machine
  Learning}, pages 1263--1272, 2017.

\bibitem[Hamilton et~al.(2017)Hamilton, Ying, and
  Leskovec]{hamilton2017inductive}
William~L Hamilton, Rex Ying, and Jure Leskovec.
\newblock Inductive representation learning on large graphs.
\newblock In \emph{Advances in Neural Information Processing Systems}, pages
  1024--1034, 2017.

\bibitem[Ying et~al.(2018)Ying, You, Morris, Ren, Hamilton, and
  Leskovec]{ying2018hierarchical}
Rex Ying, Jiaxuan You, Christopher Morris, Xiang Ren, William~L Hamilton, and
  Jure Leskovec.
\newblock Hierarchical graph representation learning with differentiable
  pooling.
\newblock In \emph{Advances in Neural Information Processing Systems}, pages
  4800--4810, 2018.

\bibitem[Xu et~al.(2019)Xu, Hu, Leskovec, and Jegelka]{xu2019powerful}
Keyulu Xu, Weihua Hu, Jure Leskovec, and Stefanie Jegelka.
\newblock How powerful are graph neural networks?
\newblock In \emph{Proceedings of the 7th International Conference on Learning
  Representations}, 2019.

\bibitem[Maron et~al.(2019{\natexlab{a}})Maron, Ben-Hamu, Shamir, and
  Lipman]{maron2019invariant}
Haggai Maron, Heli Ben-Hamu, Nadav Shamir, and Yaron Lipman.
\newblock Invariant and equivariant graph networks.
\newblock In \emph{Proceedings of the 7th International Conference on Learning
  Representations}, 2019{\natexlab{a}}.

\bibitem[Kondor et~al.(2018)Kondor, Son, Pan, Anderson, and
  Trivedi]{kondor2018covariant}
Risi Kondor, Hy~Truong Son, Horace Pan, Brandon Anderson, and Shubhendu
  Trivedi.
\newblock Covariant compositional networks for learning graphs.
\newblock \emph{arXiv preprint arXiv:1801.02144}, 2018.

\bibitem[de~Haan et~al.(2020)de~Haan, Cohen, and Welling]{de2020natural}
Pim de~Haan, Taco Cohen, and Max Welling.
\newblock Natural graph networks.
\newblock In \emph{Advances in Neural Information Processing Systems}, pages
  3636--3646, 2020.

\bibitem[Chen et~al.(2020)Chen, Chen, Villar, and Bruna]{chen2020can}
Zhengdao Chen, Lei Chen, Soledad Villar, and Joan Bruna.
\newblock Can graph neural networks count substructures?
\newblock In \emph{Advances in Neural Information Processing Systems}, pages
  10383--10395, 2020.

\bibitem[Thiede et~al.(2021)Thiede, Zhou, and Kondor]{thiede2021autobahn}
Erik~Henning Thiede, Wenda Zhou, and Risi Kondor.
\newblock Autobahn: Automorphism-based graph neural nets.
\newblock In \emph{Advances in Neural Information Processing Systems}, pages
  29922--29934, 2021.

\bibitem[Zhao et~al.(2022)Zhao, Jin, Akoglu, and Shah]{zhao2022stars}
Lingxiao Zhao, Wei Jin, Leman Akoglu, and Neil Shah.
\newblock From stars to subgraphs: Uplifting any {GNN} with local structure
  awareness.
\newblock In \emph{Proceedings of the 10th International Conference on Learning
  Representations}, 2022.

\bibitem[Bevilacqua et~al.(2022)Bevilacqua, Frasca, Lim, Srinivasan, Cai,
  Balamurugan, Bronstein, and Maron]{bevilacqua2022equivariant}
Beatrice Bevilacqua, Fabrizio Frasca, Derek Lim, Balasubramaniam Srinivasan,
  Chen Cai, Gopinath Balamurugan, Michael~M Bronstein, and Haggai Maron.
\newblock Equivariant subgraph aggregation networks.
\newblock In \emph{Proceedings of the 10th International Conference on Learning
  Representations}, 2022.

\bibitem[Elman(1990)]{elman1990finding}
Jeffrey~L Elman.
\newblock Finding structure in time.
\newblock \emph{Cognitive Science}, 14\penalty0 (2):\penalty0 179--211, 1990.

\bibitem[Cho et~al.(2014)Cho, van Merri{\"e}nboer, Gulcehre, Bahdanau,
  Bougares, Schwenk, and Bengio]{cho2014learning}
Kyunghyun Cho, Bart van Merri{\"e}nboer, Caglar Gulcehre, Dzmitry Bahdanau,
  Fethi Bougares, Holger Schwenk, and Yoshua Bengio.
\newblock Learning phrase representations using {RNN} encoder-decoder for
  statistical machine translation.
\newblock In \emph{Proceedings of the 2014 Conference on Empirical Methods in
  Natural Language Processing}, pages 1724--1734, 2014.

\bibitem[Hochreiter and Schmidhuber(1997)]{hochreiter1997long}
Sepp Hochreiter and J{\"u}rgen Schmidhuber.
\newblock Long short-term memory.
\newblock \emph{Neural computation}, 9\penalty0 (8):\penalty0 1735--1780, 1997.

\bibitem[Murphy et~al.(2019{\natexlab{b}})Murphy, Srinivasan, Rao, and
  Ribeiro]{murphy2019janossy}
Ryan~L Murphy, Balasubramaniam Srinivasan, Vinayak Rao, and Bruno Ribeiro.
\newblock Janossy pooling: {Learning} deep permutation-invariant functions for
  variable-size inputs.
\newblock In \emph{Proceedings of the 7th International Conference on Learning
  Representations}, 2019{\natexlab{b}}.

\bibitem[Vignac et~al.(2020)Vignac, Loukas, and Frossard]{vignac2020building}
Cl\'{e}ment Vignac, Andreas Loukas, and Pascal Frossard.
\newblock Building powerful and equivariant graph neural networks with
  structural message-passing.
\newblock In \emph{Advances in Neural Information Processing Systems}, pages
  14143--14155, 2020.

\bibitem[Loukas(2020)]{loukas2020graph}
Andreas Loukas.
\newblock What graph neural networks cannot learn: {Depth} vs width.
\newblock In \emph{Proceedings of the 8th International Conference on Learning
  Representations}, 2020.

\bibitem[Dasoulas et~al.(2020)Dasoulas, Santos, Scaman, and
  Virmaux]{dasoulas2020coloring}
George Dasoulas, Ludovic~Dos Santos, Kevin Scaman, and Aladin Virmaux.
\newblock Coloring graph neural networks for node disambiguation.
\newblock In \emph{Proceedings of the 29th International Joint Conference on
  Artificial Intelligence}, pages 2126--2132, 2020.

\bibitem[Morris et~al.(2019)Morris, Ritzert, Fey, Hamilton, Lenssen, Rattan,
  and Grohe]{morris2019weisfeiler}
Christopher Morris, Martin Ritzert, Matthias Fey, William~L Hamilton, Jan~Eric
  Lenssen, Gaurav Rattan, and Martin Grohe.
\newblock Weisfeiler and {Leman} go neural: {Higher-order} graph neural
  networks.
\newblock In \emph{Proceedings of the 33rd AAAI Conference on Artificial
  Intelligence}, pages 4602--4609, 2019.

\bibitem[Weisfeiler and Leman(1968)]{weisfeiler1968reduction}
Boris Weisfeiler and Andrei~A Leman.
\newblock A reduction of a graph to a canonical form and an algebra arising
  during this reduction.
\newblock \emph{Nauchno-Technicheskaya Informatsiya}, 2\penalty0 (9):\penalty0
  12--16, 1968.

\bibitem[Sato et~al.(2019)Sato, Yamada, and Kashima]{sato2019approximation}
Ryoma Sato, Makoto Yamada, and Hisashi Kashima.
\newblock Approximation ratios of graph neural networks for combinatorial
  problems.
\newblock In \emph{Advances in Neural Information Processing Systems}, pages
  4081--4090, 2019.

\bibitem[Sato et~al.(2021)Sato, Yamada, and Kashima]{sato2021random}
Ryoma Sato, Makoto Yamada, and Hisashi Kashima.
\newblock Random features strengthen graph neural networks.
\newblock In \emph{Proceedings of the 2021 SIAM International Conference on
  Data Mining}, pages 333--341, 2021.

\bibitem[Abboud et~al.(2021)Abboud, Ceylan, Grohe, and
  Lukasiewicz]{abboud2021surprising}
Ralph Abboud, {\.I}smail~{\.I}lkan Ceylan, Martin Grohe, and Thomas
  Lukasiewicz.
\newblock The surprising power of graph neural networks with random node
  initialization.
\newblock In \emph{Proceedings of the 30th International Joint Conference on
  Artificial Intelligence}, pages 2112--2118, 2021.

\bibitem[Maron et~al.(2019{\natexlab{b}})Maron, Fetaya, Segol, and
  Lipman]{maron2019universality}
Haggai Maron, Ethan Fetaya, Nimrod Segol, and Yaron Lipman.
\newblock On the universality of invariant networks.
\newblock In \emph{Proceedings of the 36th International Conference on Machine
  Learning}, pages 4363--4371, 2019{\natexlab{b}}.

\bibitem[Maron et~al.(2019{\natexlab{c}})Maron, Ben-Hamu, Serviansky, and
  Lipman]{maron2019provably}
Haggai Maron, Heli Ben-Hamu, Hadar Serviansky, and Yaron Lipman.
\newblock Provably powerful graph networks.
\newblock In \emph{Advances in Neural Information Processing Systems}, pages
  2156--2167, 2019{\natexlab{c}}.

\bibitem[Keriven and Peyr{\'e}(2019)]{keriven2019universal}
Nicolas Keriven and Gabriel Peyr{\'e}.
\newblock Universal invariant and equivariant graph neural networks.
\newblock In \emph{Advances in Neural Information Processing Systems}, pages
  7092--7101, 2019.

\bibitem[Azizian and Lelarge(2021)]{azizian2021expressive}
Waïss Azizian and Marc Lelarge.
\newblock Expressive power of invariant and equivariant graph neural networks.
\newblock In \emph{Proceedings of the 9th International Conference on Learning
  Representations}, 2021.

\bibitem[Garg et~al.(2020)Garg, Jegelka, and Jaakkola]{garg2020generalization}
Vikas Garg, Stefanie Jegelka, and Tommi Jaakkola.
\newblock Generalization and representational limits of graph neural networks.
\newblock In \emph{Proceedings of the 37th International Conference on Machine
  Learning}, pages 3419--3430, 2020.

\bibitem[Tahmasebi et~al.(2020)Tahmasebi, Lim, and
  Jegelka]{tahmasebi2020counting}
Behrooz Tahmasebi, Derek Lim, and Stefanie Jegelka.
\newblock Counting substructures with higher-order graph neural networks:
  Possibility and impossibility results.
\newblock \emph{arXiv preprint arXiv:2012.03174}, 2020.

\bibitem[You et~al.(2021)You, Gomes-Selman, Ying, and
  Leskovec]{you2021identity}
Jiaxuan You, Jonathan Gomes-Selman, Rex Ying, and Jure Leskovec.
\newblock Identity-aware graph neural networks.
\newblock In \emph{Proceedings of the 35th AAAI Conference on Artificial
  Intelligence}, pages 10737--10745, 2021.

\bibitem[Balcilar et~al.(2021)Balcilar, H{\'e}roux, Ga{\"u}z{\`e}re, Vasseur,
  Adam, and Honeine]{balcilar2021breaking}
Muhammet Balcilar, Pierre H{\'e}roux, Benoit Ga{\"u}z{\`e}re, Pascal Vasseur,
  S{\'e}bastien Adam, and Paul Honeine.
\newblock Breaking the limits of message passing graph neural networks.
\newblock In \emph{Proceedings of the 38th International Conference on Machine
  Learning}, pages 599--608, 2021.

\bibitem[Barcel{\'o} et~al.(2021)Barcel{\'o}, Geerts, Reutter, and
  Ryschkov]{barcelo2021graph}
Pablo Barcel{\'o}, Floris Geerts, Juan Reutter, and Maksimilian Ryschkov.
\newblock Graph neural networks with local graph parameters.
\newblock In \emph{Advances in Neural Information Processing Systems}, pages
  25280--25293, 2021.

\bibitem[Bouritsas et~al.(2022)Bouritsas, Frasca, Zafeiriou, and
  Bronstein]{bouritsas2022improving}
Giorgos Bouritsas, Fabrizio Frasca, Stefanos~P Zafeiriou, and Michael~M
  Bronstein.
\newblock Improving graph neural network expressivity via subgraph isomorphism
  counting.
\newblock \emph{IEEE Transactions on Pattern Analysis and Machine
  Intelligence}, 2022.

\bibitem[Bodnar et~al.(2021{\natexlab{a}})Bodnar, Frasca, Wang, Otter,
  Mont{\'u}far, Li{\`o}, and Bronstein]{bodnar2021weisfeilers}
Cristian Bodnar, Fabrizio Frasca, Yuguang Wang, Nina Otter, Guido~F
  Mont{\'u}far, Pietro Li{\`o}, and Michael Bronstein.
\newblock Weisfeiler and {Lehman} go topological: {Message} passing simplicial
  networks.
\newblock In \emph{Proceedings of the 38th International Conference on Machine
  Learning}, pages 1026--1037, 2021{\natexlab{a}}.

\bibitem[Bodnar et~al.(2021{\natexlab{b}})Bodnar, Frasca, Otter, Wang, Li{\`o},
  Mont{\'u}far, and Bronstein]{bodnar2021weisfeilerc}
Cristian Bodnar, Fabrizio Frasca, Nina Otter, Yu~Guang Wang, Pietro Li{\`o},
  Guido~F Mont{\'u}far, and Michael Bronstein.
\newblock Weisfeiler and {Lehman} go cellular: {CW} networks.
\newblock In \emph{Advances in Neural Information Processing Systems}, pages
  2625--2640, 2021{\natexlab{b}}.

\bibitem[Battaglia et~al.(2018)Battaglia, Hamrick, Bapst, Sanchez-Gonzalez,
  Zambaldi, Malinowski, Tacchetti, Raposo, Santoro, Faulkner,
  et~al.]{battaglia2018relational}
Peter~W Battaglia, Jessica~B Hamrick, Victor Bapst, Alvaro Sanchez-Gonzalez,
  Vin{\'\i}cius~Flores Zambaldi, Mateusz Malinowski, Andrea Tacchetti, David
  Raposo, Adam Santoro, Ryan Faulkner, et~al.
\newblock Relational inductive biases, deep learning, and graph networks.
\newblock \emph{arXiv preprint arXiv:1806.01261}, 2018.

\bibitem[Al~Hasan and Dave(2018)]{al2018triangle}
Mohammad Al~Hasan and Vachik~S Dave.
\newblock Triangle counting in large networks: {A} review.
\newblock \emph{Wiley Interdisciplinary Reviews: Data Mining and Knowledge
  Discovery}, 8\penalty0 (2):\penalty0 e1226, 2018.

\bibitem[Mitzenmacher and Upfal(2017)]{mitzenmacher2017probability}
Michael Mitzenmacher and Eli Upfal.
\newblock \emph{Probability and computing: {Randomization} and probabilistic
  techniques in algorithms and data analysis}.
\newblock Cambridge University Press, 2017.

\bibitem[Brijder et~al.(2019)Brijder, Geerts, Bussche, and
  Weerwag]{brijder2019expressive}
Robert Brijder, Floris Geerts, Jan Van~Den Bussche, and Timmy Weerwag.
\newblock On the expressive power of query languages for matrices.
\newblock \emph{ACM Transactions on Database Systems}, 44\penalty0
  (4):\penalty0 1--31, 2019.

\bibitem[Geerts(2021)]{geerts2021expressive}
Floris Geerts.
\newblock On the expressive power of linear algebra on graphs.
\newblock \emph{Theory of Computing Systems}, 65\penalty0 (1):\penalty0
  179--239, 2021.

\bibitem[Shervashidze et~al.(2011)Shervashidze, Schweitzer, Van~Leeuwen,
  Mehlhorn, and Borgwardt]{shervashidze2011weisfeiler}
Nino Shervashidze, Pascal Schweitzer, Erik~Jan Van~Leeuwen, Kurt Mehlhorn, and
  Karsten~M Borgwardt.
\newblock {Weisfeiler-Lehman} graph kernels.
\newblock \emph{Journal of Machine Learning Research}, 12\penalty0
  (77):\penalty0 2539--2561, 2011.

\bibitem[Zhang et~al.(2018)Zhang, Cui, Neumann, and Chen]{zhang2018end}
Muhan Zhang, Zhicheng Cui, Marion Neumann, and Yixin Chen.
\newblock An end-to-end deep learning architecture for graph classification.
\newblock In \emph{Proceedings of the 32nd AAAI Conference on Artificial
  Intelligence}, pages 4438--4445, 2018.

\bibitem[Kolouri et~al.(2021)Kolouri, Naderializadeh, Rohde, and
  Hoffmann]{kolouri2021wasserstein}
Soheil Kolouri, Navid Naderializadeh, Gustavo~K Rohde, and Heiko Hoffmann.
\newblock Wasserstein embedding for graph learning.
\newblock In \emph{Proceedings of the 9th International Conference on Learning
  Representations}, 2021.

\bibitem[Morris et~al.(2020)Morris, Kriege, Bause, Kersting, Mutzel, and
  Neumann]{morris2020tudataset}
Christopher Morris, Nils~M Kriege, Franka Bause, Kristian Kersting, Petra
  Mutzel, and Marion Neumann.
\newblock {TUDataset: A} collection of benchmark datasets for learning with
  graphs.
\newblock \emph{ICML 2020 Workshop on Graph Representation Learning and Beyond
  (GRL+)}, 2020.
\newblock URL \url{www.graphlearning.io}.

\bibitem[Dwivedi et~al.(2020)Dwivedi, Joshi, Laurent, Bengio, and
  Bresson]{dwivedi2020benchmarking}
Vijay~Prakash Dwivedi, Chaitanya~K Joshi, Thomas Laurent, Yoshua Bengio, and
  Xavier Bresson.
\newblock Benchmarking graph neural networks.
\newblock \emph{arXiv preprint arXiv:2003.00982}, 2020.

\bibitem[Bresson and Laurent(2017)]{bresson2017residual}
Xavier Bresson and Thomas Laurent.
\newblock Residual gated graph convnets.
\newblock \emph{arXiv preprint arXiv:1711.07553}, 2017.

\bibitem[Chen et~al.(2019)Chen, Villar, Chen, and Bruna]{chen2019equivalence}
Zhengdao Chen, Soledad Villar, Lei Chen, and Joan Bruna.
\newblock On the equivalence between graph isomorphism testing and function
  approximation with {GNNs}.
\newblock In \emph{Advances in Neural Information Processing Systems}, pages
  15894--15902, 2019.

\bibitem[Corso et~al.(2020)Corso, Cavalleri, Beaini, Li{\`o}, and
  Veli{\v{c}}kovi{\'c}]{corso2020principal}
Gabriele Corso, Luca Cavalleri, Dominique Beaini, Pietro Li{\`o}, and Petar
  Veli{\v{c}}kovi{\'c}.
\newblock Principal neighbourhood aggregation for graph nets.
\newblock In \emph{Advances in Neural Information Processing Systems}, pages
  13260--13271, 2020.

\bibitem[West(2001)]{west2001introduction}
Douglas~Brent West.
\newblock \emph{Introduction to graph theory}.
\newblock Prentice Hall, 2001.

\bibitem[Artin(2011)]{artin2011algebra}
Michael Artin.
\newblock \emph{Algebra}.
\newblock Pearson Prentice Hall, 2011.

\bibitem[Birkhoff and Mac~Lane(2017)]{birkhoff2017survey}
Garrett Birkhoff and Saunders Mac~Lane.
\newblock \emph{A survey of modern algebra}.
\newblock CRC Press, 2017.

\bibitem[Grohe(2017)]{grohe2017descriptive}
Martin Grohe.
\newblock \emph{Descriptive complexity, canonisation, and definable graph
  structure theory}, volume~47.
\newblock Cambridge University Press, 2017.

\bibitem[Cai et~al.(1992)Cai, F{\"u}rer, and Immerman]{cai1992optimal}
Jin-Yi Cai, Martin F{\"u}rer, and Neil Immerman.
\newblock An optimal lower bound on the number of variables for graph
  identification.
\newblock \emph{Combinatorica}, 12\penalty0 (4):\penalty0 389--410, 1992.

\bibitem[Deo(2017)]{deo2017graph}
Narsingh Deo.
\newblock \emph{Graph theory with applications to engineering and computer
  science}.
\newblock Courier Dover Publications, 2017.

\bibitem[Harary and Manvel(1971)]{harary1971number}
Frank Harary and Bennet Manvel.
\newblock On the number of cycles in a graph.
\newblock \emph{Matematick{\`y} {\v{c}}asopis}, 21\penalty0 (1):\penalty0
  55--63, 1971.

\bibitem[Chung et~al.(2012)Chung, Lam, Liu, and
  Mitzenmacher]{chung2012chernoff}
Kai-Min Chung, Henry Lam, Zhenming Liu, and Michael Mitzenmacher.
\newblock {Chernoff-Hoeffding} bounds for {Markov} chains: {Generalized} and
  simplified.
\newblock In \emph{Proceedings of the 29th International Symposium on
  Theoretical Aspects of Computer Science}, pages 124--135, 2012.

\bibitem[Haykin(2010)]{haykin2010neural}
Simon Haykin.
\newblock \emph{Neural networks and learning machines, 3/E}.
\newblock Pearson Education India, 2010.

\bibitem[Chen et~al.(2016)Chen, Li, Wang, and Lui]{chen2016general}
Xiaowei Chen, Yongkun Li, Pinghui Wang, and John~CS Lui.
\newblock A general framework for estimating graphlet statistics via random
  walk.
\newblock \emph{Proceedings of the VLDB Endowment}, 10\penalty0 (3):\penalty0
  253--264, 2016.

\bibitem[Hardiman and Katzir(2013)]{hardiman2013estimating}
Stephen~J Hardiman and Liran Katzir.
\newblock Estimating clustering coefficients and size of social networks via
  random walk.
\newblock In \emph{Proceedings of the 22nd International Conference on World
  Wide Web}, pages 539--550, 2013.

\bibitem[LeCun et~al.(1998)LeCun, Bottou, Bengio, and
  Haffner]{lecun1998gradient}
Yann LeCun, L{\'e}on Bottou, Yoshua Bengio, and Patrick Haffner.
\newblock Gradient-based learning applied to document recognition.
\newblock \emph{Proceedings of the IEEE}, 86\penalty0 (11):\penalty0
  2278--2324, 1998.

\bibitem[Achanta et~al.(2012)Achanta, Shaji, Smith, Lucchi, Fua, and
  S{\"u}sstrunk]{achanta2012slic}
Radhakrishna Achanta, Appu Shaji, Kevin Smith, Aurelien Lucchi, Pascal Fua, and
  Sabine S{\"u}sstrunk.
\newblock {SLIC} superpixels compared to state-of-the-art superpixel methods.
\newblock \emph{IEEE Transactions on Pattern Analysis and Machine
  Intelligence}, 34\penalty0 (11):\penalty0 2274--2282, 2012.

\bibitem[Irwin et~al.(2012)Irwin, Sterling, Mysinger, Bolstad, and
  Coleman]{irwin2012zinc}
John~J Irwin, Teague Sterling, Michael~M Mysinger, Erin~S Bolstad, and Ryan~G
  Coleman.
\newblock {ZINC: A} free tool to discover chemistry for biology.
\newblock \emph{Journal of Chemical Information and Modeling}, 52\penalty0
  (7):\penalty0 1757--1768, 2012.

\bibitem[Ioffe and Szegedy(2015)]{ioffe2015batch}
Sergey Ioffe and Christian Szegedy.
\newblock Batch normalization: {Accelerating} deep network training by reducing
  internal covariate shift.
\newblock In \emph{Proceedings of the 32nd International Conference on Machine
  Learning}, pages 448--456, 2015.

\bibitem[Glorot and Bengio(2010)]{glorot2010understanding}
Xavier Glorot and Yoshua Bengio.
\newblock Understanding the difficulty of training deep feedforward neural
  networks.
\newblock In \emph{Proceedings of the 13th International Conference on
  Artificial Intelligence and Statistics}, pages 249--256, 2010.

\bibitem[Kingma and Ba(2015)]{kingma2015adam}
Diederik~P Kingma and Jimmy~Lei Ba.
\newblock Adam: {A} method for stochastic optimization.
\newblock In \emph{Proceedings of the 3rd International Conference on Learning
  Representations}, 2015.

\bibitem[Hagberg et~al.(2008)Hagberg, Swart, and S~Chult]{hagberg2008exploring}
Aric Hagberg, Pieter Swart, and Daniel S~Chult.
\newblock Exploring network structure, dynamics, and function using {NetworkX}.
\newblock In \emph{Proceedings of the 7th Python in Science Conference}, pages
  11--15, 2008.

\bibitem[Paszke et~al.(2019)Paszke, Gross, Massa, Lerer, Bradbury, Chanan,
  Killeen, Lin, Gimelshein, Antiga, et~al.]{paszke2019pytorch}
Adam Paszke, Sam Gross, Francisco Massa, Adam Lerer, James Bradbury, Gregory
  Chanan, Trevor Killeen, Zeming Lin, Natalia Gimelshein, Luca Antiga, et~al.
\newblock {PyTorch: An} imperative style, high-performance deep learning
  library.
\newblock In \emph{Advances in Neural Information Processing Systems}, pages
  8026--8037, 2019.

\bibitem[Wang et~al.(2019)Wang, Yu, Zheng, Gan, Gai, Ye, Li, Zhou, Huang, Ma,
  et~al.]{wang2019deep}
Minjie Wang, Lingfan Yu, Da~Zheng, Quan Gan, Yu~Gai, Zihao Ye, Mufei Li,
  Jinjing Zhou, Qi~Huang, Chao Ma, et~al.
\newblock Deep graph library: {Towards} efficient and scalable deep learning on
  graphs.
\newblock \emph{ICLR Workshop on Representation Learning on Graphs and
  Manifolds}, 2019.

\end{thebibliography}

%%%%%%%%%%%%%%%%%%%%%%%%%%%%%%%%%%%%%%%%%%%%%%%%%%%%%%%%%%%%

\clearpage

\appendix

\newtheorem{apptheorem}{Theorem}[section]
\newtheorem{appproblem}[apptheorem]{Problem}
\newtheorem{appconjecture}[apptheorem]{Conjecture}
\newtheorem{appobservation}[apptheorem]{Observation}
\newtheorem{appproposition}[apptheorem]{Proposition}
\newtheorem{applemma}[apptheorem]{Lemma}
\newtheorem{appcorollary}[apptheorem]{Corollary}

\section{Background on Graph Theory}
\label{sec:graph}

Given a graph $G = (\mathcal{V}, \mathcal{E})$, a \emph{walk} in $G$ is a finite sequence of alternating vertices and edges such as $v_0, e_1, v_1, e_2, \ldots, e_m, v_m$, where each edge $e_i = (v_{i-1}, v_i)$.
A walk may have repeated edges.
A \emph{trail} is a walk in which all the edges are distinct.
A \emph{path} is a trail in which all vertices (hence all edges) are distinct (except, possibly, $v_0 = v_m$).
A trail or path is \emph{closed} if $v_0 = v_m$, and a closed path containing at least one edge is a \emph{cycle} \cite{west2001introduction}.

A \emph{Hamiltonian path} is a path in a graph that passes through each vertex exactly once.
A \emph{Hamiltonian cycle} is a cycle in a graph that passes through each vertex exactly once.
A \emph{Hamiltonian graph} is a graph that contains a Hamiltonian cycle.

Let $G = (\mathcal{V}, \mathcal{E})$ and $G' = (\mathcal{V}', \mathcal{E}')$ be graphs.
If $G' \subseteq G$ and $G'$ contains all the edges $(v_i, v_j) \in \mathcal{E}$ with $v_i, v_j \in \mathcal{V}'$, then $G'$ is an \emph{induced subgraph} of $G$, and we say that $\mathcal{V}'$ \emph{induces} $G'$ in $G$.

An \emph{empty graph} is a graph whose edge-set is empty.
A \emph{regular graph} is a graph in which each vertex has the same degree.
If each vertex has degree $r$, the graph is \emph{$r$-regular}.
A \emph{strongly regular graph} in the family SRG($v, r, \lambda, \mu$) is an $r$-regular graph with $v$ vertices, where every two adjacent vertices have $\lambda$ common neighbors, and every two non-adjacent vertices have $\mu$ common neighbors.

A \emph{complete graph} is a simple undirected graph in which every pair of distinct vertices is adjacent. We denote the complete graph on $n$ vertices by $K_n$.
A \emph{tournament} is a directed graph in which each edge of a complete graph is given an orientation. We denote the tournament on $n$ vertices by $\vec{K}_n$.
A \emph{clique} of a graph $G$ is a complete induced subgraph of $G$. A clique of size $k$ is called a $k$-clique.

The \emph{local clustering coefficient} of a vertex quantifies how close its neighbors are to being a clique (complete graph).
The local clustering coefficient $c_v$ of a vertex $v$ is given by the proportion of links between the $n$ vertices within its neighborhood $\mathcal{N}(v)$ divided by the number of links that could possibly exist between them, defined as $c_v = \dfrac{2 \left| \left\{ e_{ij}: i,j \in \mathcal{N}(v), e_{ij} \in \mathcal{E} \right\} \right|} {n(n-1)}$.
This measure is 1 if every neighbor connected to $v$ is also connected to every other vertex within the neighborhood.

Let $G = (\mathcal{V}, \mathcal{E})$ and $G' = (\mathcal{V}', \mathcal{E}')$ be graphs.
An \emph{isomorphism} $\vartheta: \mathcal{V} \to \mathcal{V}'$ between $G$ and $G'$ is a bijective map that maps pairs of connected vertices to pairs of connected vertices, and likewise for pairs of non-connected vertices, i.e., $(\vartheta(u), \vartheta(v)) \in \mathcal{E}'$ iff $(u, v) \in \mathcal{E}$ for all $u$ and $v$ in $\mathcal{V}$.

\section{Background on Group Theory}
\label{sec:group}

Since we deal with finite sets in this paper, all the following definitions are about finite groups.

For an arbitrary element $x$ in a group $\mathfrak{G}$, the \emph{order} of $x$ is the smallest positive integer $n$ such that $x^n = e$, where $e$ is the \emph{identity element}.
$H = \{ e, x, x^2, \ldots, x^{n-1} \}$ is the \emph{cyclic subgroup} generated by $x$ and is often denoted by $H = \left \langle x \right \rangle$.
A \emph{cyclic group} is a group that is equal to one of its cyclic subgroups: $\mathfrak{G} = \left \langle g \right \rangle$ for some element $g$, and the element $g$ is called a \emph{generator}.
The cyclic group with $n$ elements is denoted by $\mathbb{Z}_n$ \cite{artin2011algebra, birkhoff2017survey}.

A \emph{permutation} of a finite set $S$ is a bijective map from $S$ to itself.
In Cauchy’s two-line notation, it denotes such a permutation by listing the ``natural'' order for all the $n$ elements of $S$ in the first row, and for each one, its image below it in the second row:
$\sigma = \left( \begin{matrix}
   1 & 2 & \cdots  & n  \\
   \sigma(1) & \sigma(2) & \cdots  & \sigma(n)  \\
\end{matrix} \right)$.
A \emph{cycle} of length $r$ (or \emph{$r$-cycle}) is a permutation $\sigma$ for which there exists an element $i_1$ in $\{ 1, 2, \ldots, n \}$ such that $\sigma(i_1) = i_2, \sigma(i_2) = i_3, \cdots, \sigma(i_{r-1}) = i_r, \sigma(i_r) = i_1$ are the only elements moved by $\sigma$.
In cycle notation, it denotes such a cycle (or $r$-cycle) by $(i_1\ i_2\ \cdots \ i_r)$.

A \emph{permutation group} is a group whose elements are permutations of a given set $S$, with the group operation ``$\circ$'' being the composition of permutations.
The permutation group on the set $S$ is denoted by $Perm(S)$.
A \emph{symmetric group} is a group whose elements are \textbf{all} permutations of a given set $S$.
The symmetric group on the set $S = [n] = \left\{ 1, 2, \ldots, n \right\}$ is denoted by $\mathcal{S}_n$ \cite{artin2011algebra}.
Every permutation group is a subgroup of a symmetric group.

A \emph{group action} $\alpha$ of a group $\mathfrak{G}$ on a set $S$ is a map $\alpha: \mathfrak{G} \times S \to S$, denoted by $(g, s) \mapsto gs$ (with $\alpha(g, s)$ often shortened to $gs$ or $g \cdot s$) that satisfies the following two axioms:
\begin{itemize}[itemsep= 0pt, topsep = 0pt]
\item[a)] identity: $e \cdot s = s$, for all $s \in S$, where $e$ is the identity element of $\mathfrak{G}$.
\item[b)] associative law: $(g_1 \circ g_2) \cdot s = g_1 \cdot (g_2 \cdot s)$, for all $g_1, g_2 \in \mathfrak{G}$ and $s \in S$, where $\circ$ denotes the operation or composition in $\mathfrak{G}$.
\end{itemize}

Let $\mathfrak{G}$ and $\mathfrak{G}'$ be groups.
A \emph{homomorphism} $\varphi: \mathfrak{G} \to \mathfrak{G}'$ is a map from $\mathfrak{G}$ to $\mathfrak{G}'$ such that $\varphi(ab) = \varphi(a)\varphi(b)$ for all $a$ and $b$ in $\mathfrak{G}$.
An \emph{isomorphism} $\varphi: \mathfrak{G} \to \mathfrak{G}'$ from $\mathfrak{G}$ to $\mathfrak{G}'$ is a bijective group homomorphism - a bijective map such that $\varphi(ab) = \varphi(a)\varphi(b)$ for all $a$ and $b$ in $\mathfrak{G}$ \cite{artin2011algebra}.
We use the symbol $\cong$ to denote two groups $\mathfrak{G}$ and $\mathfrak{G}'$ are isomorphic, i.e., $\mathfrak{G} \cong \mathfrak{G}'$.

\section{Definition of \emph{k}-WL Test}
\label{sec:kWL}

There are different definitions of the $k$-dimensional Weisfeiler-Lehman ($k$-WL) test for $k \ge 2$, while in this work, we follow the definition in \citet{chen2020can}.
Note that the $k$-WL test here is equivalent to the $k$-WL tests in \cite{grohe2017descriptive, morris2019weisfeiler, maron2019provably, azizian2021expressive}, and the $(k - 1)$-WL test in \cite{cai1992optimal} (\citet{grohe2017descriptive} calls this version as $k$-WL$^\prime$).
$(k + 1)$-WL test has been proven to be strictly more powerful than $k$-WL test \cite{cai1992optimal}.

The $k$-WL algorithm is a generalization of the 1-WL, it colors tuples from $\mathcal{V}^k$ instead of nodes.
For any $k$-tuple $s = (i_1, \ldots, i_k) \in \mathcal{V}^k$ and each $j \in [k] = \{1, \ldots, k\}$, define the $j$-th neighborhood
\begin{equation*}
    \mathcal{N}_j(s) = \{ (i_1, \ldots, i_{j-1}, u, i_{j+1}, \ldots, i_k) \ | \ u \in \mathcal{V} \}
\end{equation*}
That is, the $j$-th neighborhood $\mathcal{N}_j(s)$ of the $k$-tuple $s$ is obtained by replacing the $j$-th component of $s$ with every node from $\mathcal{V}$.

Given a pair of graphs $G$ and $G'$, we use the $k$-WL algorithm to test them for isomorphism.
Suppose that the two graphs have the same number of vertices since otherwise, they can be told apart easily.
Without loss of generality, we assume that they share the same set of vertex indices, $\mathcal{V}$ (but may differ in $\mathcal{E}$).
The $k$-WL test follows the following coloring procedure.

\begin{itemize}[itemsep= 0pt, topsep = 0pt]

\item[1)]
For each of the graphs, at iteration 0, the test assigns an initial color in the color space $\Gamma$ to each $k$-tuple according to its \emph{atomic type}, i.e., two $k$-tuples $s$ and $s'$ in $\mathcal{V}^k$ get the same color if the subgraphs induced from nodes of $s$ and $s'$ are isomorphic.

\item[2)]
In each iteration $t > 0$, the test computes a $k$-tuple coloring $c_k^{(t)} :\mathcal{V}^k \to \Gamma$.
More specifically, let $c_k^{(t)}(s)$ denote the color of $s$ in $G$ assigned at the $t$-th iteration, and let ${c'}_k^{(t)}(s')$ denote the color assigned for $s'$ in $G'$. Define
\begin{align*}
    C_j^{(t)}(s) &= \text{HASH}_1^{(t)} \left( \left\{ c_k^{(t-1)}(w) \ \Big| \ w \in \mathcal{N}_j(s) \right\} \right) \\
    {C'}_j^{(t)}(s') &= \text{HASH}_1^{(t)} \left( \left\{ {c'}_k^{(t-1)}(w') \ \Big| \ w' \in \mathcal{N}_j(s') \right\} \right)
\end{align*}
where $\text{HASH}_1^{(t)}$ is a hash function that maps injectively from the space of multisets of colors to some intermediate space.
Then let
\begin{align*}
    c_k^{(t)}(s) &= \text{HASH}_2^{(t)} \left( \left( c_k^{(t-1)}(s), \left( C_1^{(t)}(s), \ldots , C_k^{(t)}(s) \right) \right) \right) \\
    {c'}_k^{(t)}(s') &= \text{HASH}_2^{(t)} \left( \left( {c'}_k^{(t-1)}(s'), \left( {C'}_1^{(t)}(s'), \ldots , {C'}_k^{(t)}(s') \right) \right) \right)
\end{align*}
where $\text{HASH}_2^{(t)}$ maps injectively from its input space to the color space $\Gamma$, $c_k^{(t)}(s)$ and ${c'}_k^{(t)}(s)$ are updated iteratively in this way.

\item[3)]
The test will terminate and return the result that the two graphs are not isomorphic if the following two multisets differ at some iteration $t$:
\begin{equation*}
    \left\{ c_k^{(t)}(s) \ \Big| \ s \in \mathcal{V}^k \right\} \neq
    \left\{ {c'}_k^{(t)}(s') \ \Big| \ s' \in \mathcal{V}^k \right\}
\end{equation*}

\end{itemize}

For the detailed difference between $k$-WL test here and $(k - 1)$-WL test in \citet{cai1992optimal} ($k$-WL$^\prime$ in \citet{grohe2017descriptive}), see Remark 3.5.9 in \citet{grohe2017descriptive}.

\section{Distinguishing Non-Isomorphic Graph Pairs: Permutation-Sensitive vs. Permutation-Invariant Aggregation Functions}
\label{sec:example}

\begin{figure}
    \centering
    \includegraphics[width=0.7\linewidth]{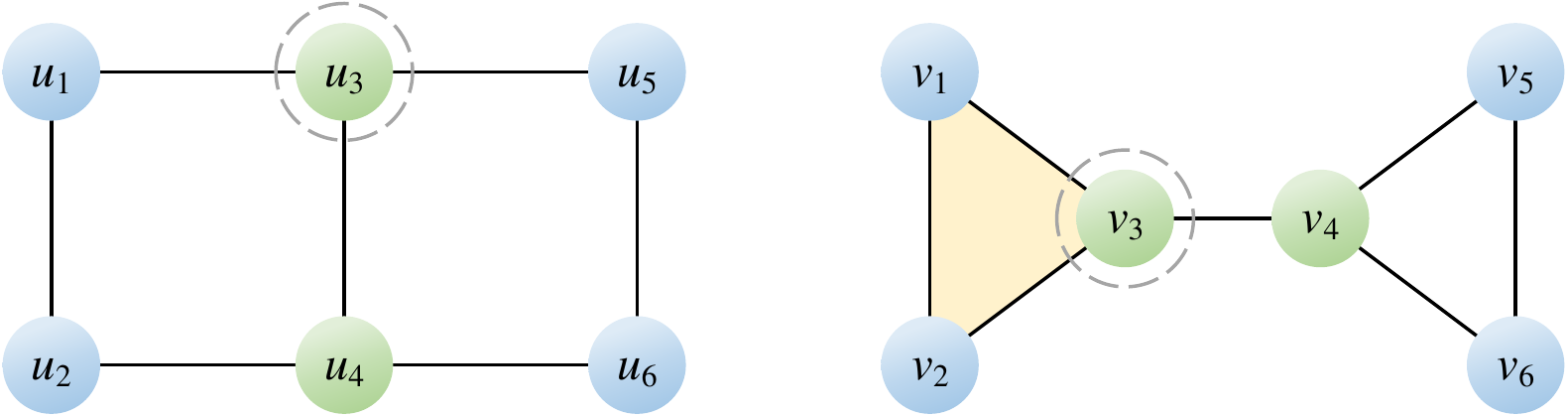}
    \caption{A pair of non-isomorphic graphs that cannot be distinguished by permutation-invariant aggregation functions, but can be easily distinguished by permutation-sensitive aggregation functions.}
    \label{fig:example}
\end{figure}

Let $f$ be an arbitrary aggregation function.
For a node $v$, let $x_v$ ($b$ for \textcolor{blue}{blue}, $g$ for \textcolor{green}{green}) denote the initial node feature, $h_v$ denote the feature transformed by $f$.
In the initial stage, we have:
\begin{gather*}
    x_{u_1} = x_{u_2} = x_{u_5} = x_{u_6} = b,~x_{u_3} = x_{u_4} = g \\
    x_{v_1} = x_{v_2} = x_{v_5} = x_{v_6} = b,~x_{v_3} = x_{v_4} = g
\end{gather*}
Figure \ref{fig:example} illustrates a pair of non-isomorphic graphs that 2-WL test and most permutation-invariant aggregation functions fail to distinguish.
Suppose $f$ is permutation-invariant, we take the sum aggregator SUM$(\cdot)$ as an example to illustrate this process.
After the first round of iteration, the transformed feature of each node is:
\begin{gather*}
    h_{u_1} = h_{u_2} = h_{u_5} = h_{u_6} = b + g,~
    h_{u_3} = h_{u_4} = 2b + g \\
    h_{v_1} = h_{v_2} = h_{v_5} = h_{v_6} = b + g,~
    h_{v_3} = h_{v_4} = 2b + g
\end{gather*}
We can find that the distributions of node features of these two graphs are the same.
Similarly, after each round of iteration, these two graphs always produce the same distributions of node features.
% We can find that the distributions of initial and transformed node features are the same. Similarly, the distributions of node features before and after each round of iteration are also the same.
Hence we can conclude that the 2-WL test and the permutation-invariant function SUM$(\cdot)$ fail to distinguish these two graphs.

In contrast, suppose $f$ is permutation-sensitive, we take a generic permutation-sensitive aggregator $h^{(t)} = k \cdot h^{(t-1)} + x^{(t)}$ as an example to illustrate its process.
Here $x^{(t)}$ is the $t$-th input node feature, $h^{(t)}$ is the corresponding transformed feature with $h^{(0)} = 0$, and the learnable parameter $k > 1$ measures the pairwise correlation between $x^{(t-1)}$ and $x^{(t)}$.
For the left graph $G_1$, we focus on node $u_3$.
Let the input ordering of neighboring nodes be $u_1, u_4, u_5$, i.e., $x_{u_3}^{(1)} \to x_{u_3}^{(2)} \to x_{u_3}^{(3)} = b \to g \to b$, then $f$ only encodes the pairwise correlation between $b$ and $g$.
Thus, we have
\begin{align*}
    h_{u_3}^{(1)} &= k \cdot 0 + b = b \\
    h_{u_3}^{(2)} &= k \cdot b + g = kb + g \\
    h_{u_3}^{(3)} &= k \cdot (kb + g) + b = (k^2+1)b + kg
\end{align*}
For the right graph $G_2$, we focus on node $v_3$.
Let the input ordering of neighboring nodes be $v_1, v_2, v_4$, i.e., $x_{v_3}^{(1)} \to x_{v_3}^{(2)} \to x_{v_3}^{(3)} = b \to b \to g$, then $f$ also encodes the pairwise correlation between $b$ and $b$.
Thus, we have
\begin{align*}
    h_{v_3}^{(1)} &= k \cdot 0 + b = b \\
    h_{v_3}^{(2)} &= k \cdot b + b = kb + b \\
    h_{v_3}^{(3)} &= k \cdot (kb + b) + g = (k^2+k)b + g
\end{align*}
After the first round of iteration, the node feature $h_{u_3}^{(3)}$ of $u_3$ differs from the $h_{v_3}^{(3)}$ of $v_3$.
Hence we can conclude that the permutation-sensitive aggregation function $f$ can distinguish these two graphs.
Moreover, the weight ratio of $b$ and $g$ in $h_{u_3}^{(3)}$ is $(k^2+1):k$, which is smaller than that in $h_{v_3}^{(3)}$, i.e., $(k^2+k):1$.
This fact indicates that, in $G_1$, $f$ focuses more on encoding the pairwise correlation between $b$ and $g$.
In contrast, in $G_2$, $f$ focuses more on encoding the pairwise correlation between $b$ and $b$, thereby exploiting the triangular substructure such as $\triangle v_1v_3v_2$.
It is worth noting that when $k = 1$, the function $f$ is $h^{(t)} = h^{(t-1)} + x^{(t)}$ and degenerates to the permutation-invariant function SUM$(\cdot)$, resulting in $h_{u_3}^{(3)} = h_{v_3}^{(3)} = 2b + g$.

\section{Proof of Theorem \ref{thm:arrange}}
\label{sec:pf_arrange}

\def\theoremautorefname{Theorem}
\textbf{\autoref{thm:arrange}.}
\emph{Let $n (n \ge 4)$ denote the number of 1-hop neighboring nodes around the central node $v$.
There are $\lfloor (n-1)/2 \rfloor$ kinds of arrangements in total, satisfying that their corresponding 2-ary dependencies are disjoint.
Meanwhile, after at least $\lfloor n/2 \rfloor$ arrangements (including the initial one), all 2-ary dependencies have been covered at least once.}

\begin{proof}
Construct a simple undirected graph $G' = (\mathcal{V}', \mathcal{E}')$, where $\mathcal{V}'$ denotes the $n$ neighboring nodes (abbreviated as nodes in the following) around the central node $v$, and $\mathcal{E}'$ represents an edge set in which each edge indicates the corresponding 2-ary dependency has been covered in some arrangements.
Thus, each arrangement corresponds to a Hamiltonian cycle in graph $G'$.
For any two arrangements, detecting whether their corresponding 2-ary dependencies are disjoint can be analogous to finding two edge-disjoint Hamiltonian cycles.
Since every pair of nodes can form a 2-ary dependency, the first problem can be translated into finding the maximum number of edge-disjoint Hamiltonian cycles in a complete graph $K_n$, and the second problem can be translated into finding the minimum number of Hamiltonian cycles to cover a complete graph $K_n$.

\begin{figure}[b]
    \centering
    \subfigure[$n$ is odd]{
	\includegraphics[width=0.315\linewidth]{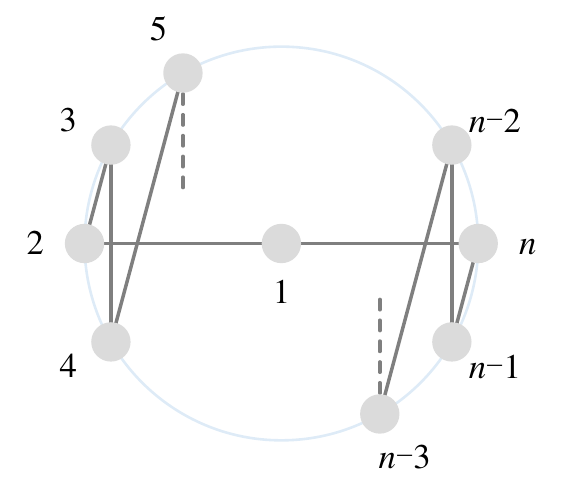}
	\label{fig:arrange:odd}
	}
    \subfigure[$n$ is even]{
	\includegraphics[width=0.315\linewidth]{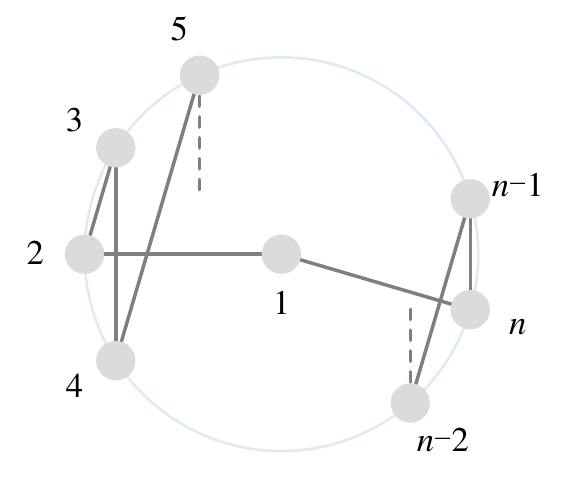}
	\label{fig:arrange:even}
	}
    \subfigure[our revision when $n$ is even]{
	\includegraphics[width=0.315\linewidth]{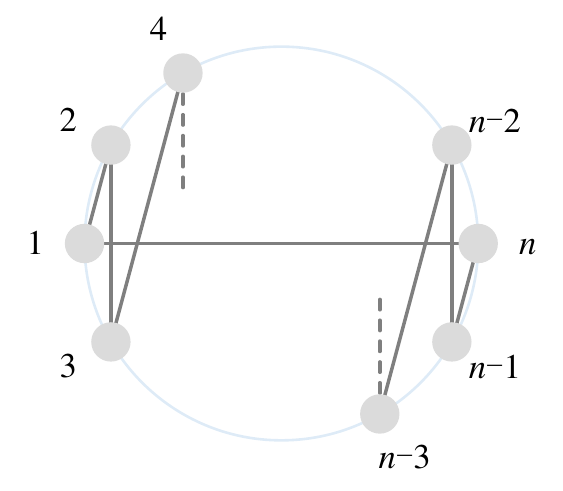}
	\label{fig:arrange:change}
	}
    \caption{The initial arrangements (following the gray solid lines).}
    \label{fig:arrange}
\end{figure}

Since a $K_n$ has $\frac{n(n-1)}{2}$ edges and each Hamiltonian cycle has $n$ edges, there are at most $\lfloor \frac{n(n-1)}{2} / n \rfloor = \lfloor \frac{n-1}{2} \rfloor$ edge-disjoint Hamiltonian cycles in a $K_n$.
In addition, we can specifically construct $\lfloor \frac{n-1}{2} \rfloor$ edge-disjoint Hamiltonian cycles as follows.
If $n$ is odd, keep the nodes fixed on a circle with node 1 at the center, rotate the node numbers on the circle clockwise by $\frac{360^\circ}{n-1}, 2 \times \frac{360^\circ}{n-1}, \ldots, \frac{n-3}{2} \times \frac{360^\circ}{n-1}$, while the graph structure always remains unchanged as the initial arrangement shown in Figure \ref{fig:arrange:odd}.
Each rotation can be formulated as the following permutation $\sigma'$:
\begin{equation*}
\sigma' = \left\{
    \begin{aligned}
    & \small{ \left( \begin{matrix}
    1 & 2 & 3 & 4 & 5 & \cdots & n-1 & n  \\
    1 & 4 & 2 & 6 & 3 & \cdots & n & n-2  \\
    \end{matrix} \right) }
    = \left( 2\ 4\ 6\ \cdots \ n-1\ n\ n-2\ \cdots \ 7\ 5\ 3 \right), \text{if}~n~\text{is odd}, \\
    & \small{ \left( \begin{matrix}
    1 & 2 & 3 & 4 & 5 & \cdots & n-1 & n  \\
    1 & 4 & 2 & 6 & 3 & \cdots & n-3 & n-1  \\
    \end{matrix} \right) }
    = \left( 2\ 4\ 6\ \cdots \ n-2\ n\ n-1\ \cdots \ 7\ 5\ 3 \right), \text{if}~n~\text{is even}.
    \end{aligned}
\right.
\end{equation*}

Observe that each rotation generates a new Hamiltonian cycle containing completely different edges from before.
Thus we have $\frac{n-3}{2} = \lfloor \frac{n-1}{2} \rfloor - 1$ new Hamiltonian cycles with all edges disjoint from the ones in Figure \ref{fig:arrange:odd} and among themselves \cite{deo2017graph}.
If $n$ is even, the node arrangement can be initialized as shown in Figure \ref{fig:arrange:even}, and $\frac{n-4}{2} = \lfloor \frac{n-1}{2} \rfloor - 1$ new Hamiltonian cycles can be constructed successively in a similar way.
We thus conclude that there are $\lfloor \frac{n-1}{2} \rfloor$ kinds of arrangements in total, satisfying that their corresponding 2-ary dependencies are disjoint.

Furthermore, if $n$ is odd, $K_n$ has $\frac{n(n-1)}{2}$ edges \emph{divisible} by the length $n$ of each Hamiltonian cycle.
Therefore, we can exactly cover all edges by the above $\lfloor \frac{n-1}{2} \rfloor = \frac{n-1}{2} = \lfloor \frac{n}{2} \rfloor$ kinds of arrangements.
On the contrary, if $n$ is even, $K_n$ has $\frac{n(n-1)}{2}$ edges \emph{indivisible} by the length $n$ of each Hamiltonian cycle, remaining $\frac{n}{2}$ edges uncovered by the above $\lfloor \frac{n-1}{2} \rfloor = \frac{n-2}{2}$ kinds of arrangements.
Thus we continue to perform the permutation $\sigma'$ once, i.e., $\lfloor \frac{n-1}{2} \rfloor + 1 = \frac{n}{2} = \lfloor \frac{n}{2} \rfloor$ kinds of arrangements in total, to cover all edges but result in $\frac{n}{2}$ edges duplicated twice.

As discussed in the main body, these $\lfloor \frac{n}{2} \rfloor$ arrangements and the corresponding $\lfloor \frac{n}{2} \rfloor$ Hamiltonian cycles are modeled by the permutation-sensitive function in a directed manner.
In addition, we also expect to reverse these $\lfloor \frac{n}{2} \rfloor$ directed Hamiltonian cycles by performing the permutation $\sigma'$ successively, thereby transforming them into an undirected manner.
However, $\sigma'$ cannot satisfy this requirement if $n$ is even.
Thus, we propose to revise the permutation $\sigma'$ into the following one:
\begin{equation*}
\sigma = \left\{
    \begin{aligned}
    & \small{ \left( \begin{matrix}
    1 & 2 & 3 & 4 & 5 & \cdots & n-1 & n  \\
    1 & 4 & 2 & 6 & 3 & \cdots & n & n-2  \\
    \end{matrix} \right) }
    = \left( 2\ 4\ 6\ \cdots \ n-1\ n\ n-2\ \cdots \ 7\ 5\ 3 \right), \text{if}~n~\text{is odd}, \\
    & \small{ \left( \begin{matrix}
    1 & 2 & 3 & 4 & \cdots & n-1 & n  \\
    3 & 1 & 5 & 2 & \cdots & n & n-2  \\
    \end{matrix} \right) }
    = \left( 1\ 3\ 5\ \cdots \ n-1\ n\ n-2\ \cdots \ 6\ 4\ 2 \right), \text{if}~n~\text{is even}.
    \end{aligned}
\right.
\end{equation*}
where $\sigma$ is the same as $\sigma'$ when $n$ is odd, but a little different when $n$ is even.
If $n$ is even, $\sigma$ is an $n$-cycle, but $\sigma'$ is an $(n-1)$-cycle.
The corresponding initial node arrangement after revision is shown in Figure \ref{fig:arrange:change}.
After adding a virtual node 0 at the center in Figure \ref{fig:arrange:change}, $\sigma$ becomes the same as $\sigma'$ with $n+1$ in Figure \ref{fig:arrange:odd}, which can cover all edges with $\lfloor \frac{(n+1)-1}{2} \rfloor = \lfloor \frac{n}{2} \rfloor$ kinds of arrangements.
Moreover, after performing $\sigma$ for $n$ times in succession, it can cover a complete graph bi-directionally but $\sigma'$ fails.

In conclusion, after performing $\sigma$ or $\sigma'$ for $\lfloor \frac{n}{2} \rfloor - 1$ times in succession (excluding the initial one), all 2-ary dependencies have been covered at least once.
\end{proof}

\section{Proof of Lemma \ref{thm:cyclic}}
\label{sec:pf_cyclic}

\begin{apptheorem}
\label{thm:order_permutation}
	The order of any permutation is the least common multiple of the lengths of its disjoint cycles \cite{birkhoff2017survey}.
\end{apptheorem}

\begin{appproposition}
\label{thm:order_cyclic}
	The order of a cyclic group is equal to the order of its generator \cite{artin2011algebra}.
\end{appproposition}

Using Theorem \ref{thm:order_permutation} and Proposition \ref{thm:order_cyclic}, we prove Lemma \ref{thm:cyclic} as follows.

\def\theoremautorefname{Lemma}
\textbf{\autoref{thm:cyclic}.}
\emph{For the permutation $\sigma$ of $n$ indices, $\mathfrak{G} = \{ e, \sigma, \sigma^2, \ldots, \sigma^{n-2} \}$ is a permutation group isomorphic to the cyclic group $\mathbb{Z}_{n-1}$ if $n$ is odd.
And $\mathfrak{G} = \{ e, \sigma, \sigma^2, \ldots, \sigma^{n-1} \}$ is a permutation group isomorphic to the cyclic group $\mathbb{Z}_{n}$ if $n$ is even.}

\begin{proof}
If $n$ is odd, we find the order of permutation $\sigma$ first. Since
\begin{equation*}
    \sigma = \left( \begin{matrix}
    1 & 2 & 3 & 4 & 5 & \cdots  & n-1 & n  \\
    1 & 4 & 2 & 6 & 3 & \cdots  & n & n-2  \\
    \end{matrix} \right) = \left( 1 \right) \left( 2\ 4\ 6\ \cdots \ n-1\ n\ n-2\ \cdots \ 7\ 5\ 3 \right)
\end{equation*}
Let $\pi_1 = \left( 1 \right)$, $\pi_2 = \left( 2\ 4\ 6\ \cdots \ n-1\ n\ n-2\ \cdots \ 7\ 5\ 3 \right)$, then the permutation $\sigma$ can be represented as the product of these two disjoint cycles, i.e., $\sigma = \pi_1 \pi_2$.
Here $\pi_1$ is a $1$-cycle of length $1$, $\pi_2$ is an $(n-1)$-cycle of length $n-1$.
Using Theorem \ref{thm:order_permutation}, the order of permutation $\sigma$ is the least common multiple of $1$ and $n-1$: $\text{lcm}(1, n-1) = n-1$, which indicates that $\sigma^{n-1} = e$.
Therefore, $\mathfrak{G} = \{ e, \sigma, \sigma^2, \ldots, \sigma^{n-2} \}$ is a permutation group generated by $\sigma$, i.e., $\mathfrak{G} = \left \langle \sigma \right \rangle$.
According to the definition of the cyclic group (see Appendix \ref{sec:group}), $\mathfrak{G}$ is isomorphic to a cyclic group.
By Proposition \ref{thm:order_cyclic}, the order of group $\mathfrak{G} = \left \langle \sigma \right \rangle$ is equal to the order of its generator $\sigma$, i.e., $n-1$.
Thus, $\mathfrak{G} = \{ e, \sigma, \sigma^2, \ldots, \sigma^{n-2} \}$ is a permutation group isomorphic to the cyclic group $\mathbb{Z}_{n-1}$.

Similarly, we can prove that $\mathfrak{G} = \{ e, \sigma, \sigma^2, \ldots, \sigma^{n-1} \}$ is a permutation group isomorphic to the cyclic group $\mathbb{Z}_{n}$ if $n$ is even.
\end{proof}

\begin{apptheorem}[Cayley’s Theorem]
\label{thm:Cayley}
	Every finite group is isomorphic to a permutation group \cite{artin2011algebra}.
\end{apptheorem}

The conclusion of Lemma \ref{thm:cyclic} also obeys the most fundamental Cayley’s Theorem in group theory.

\section{Proof of Corollary \ref{thm:action} and the Diagram of Group Action}
\label{sec:diagram}

\def\theoremautorefname{Corollary}
\textbf{\autoref{thm:action}.}
\emph{The map $\alpha: \mathfrak{G} \times S \to S$ denoted by $(g,s) \mapsto gs$ is a group action of $\mathfrak{G}$ on $S$.}

\begin{proof}
Let $e$ be the identity element of $\mathfrak{G}$ and $id_\sigma$ be the identity permutation.
And let $\circ$ denote the composition in $\mathfrak{G}$.
For all $\sigma^i, \sigma^j \in \mathfrak{G}$ and $s \in S$, we have
\begin{gather*}
    \alpha(e, s) = e \cdot s = id_\sigma \cdot s = s \\
    \alpha(\sigma^i \sigma^j, s) = (\sigma^i \circ \sigma^j) \cdot s = \sigma^i \cdot (\sigma^j \cdot s) = \alpha(\sigma^i, \alpha(\sigma^j, s))
\end{gather*}
Thus, the map $\alpha$ defines a group action of the permutation group $\mathfrak{G}$ on the set $S$.
\end{proof}

\begin{figure}
    \centering
    \subfigure[$\mathfrak{G}_1 = \{ e, \sigma, \sigma^2, \ldots, \sigma^5 \} \cong \mathbb{Z}_6$]{
	\includegraphics[width=0.66\linewidth]{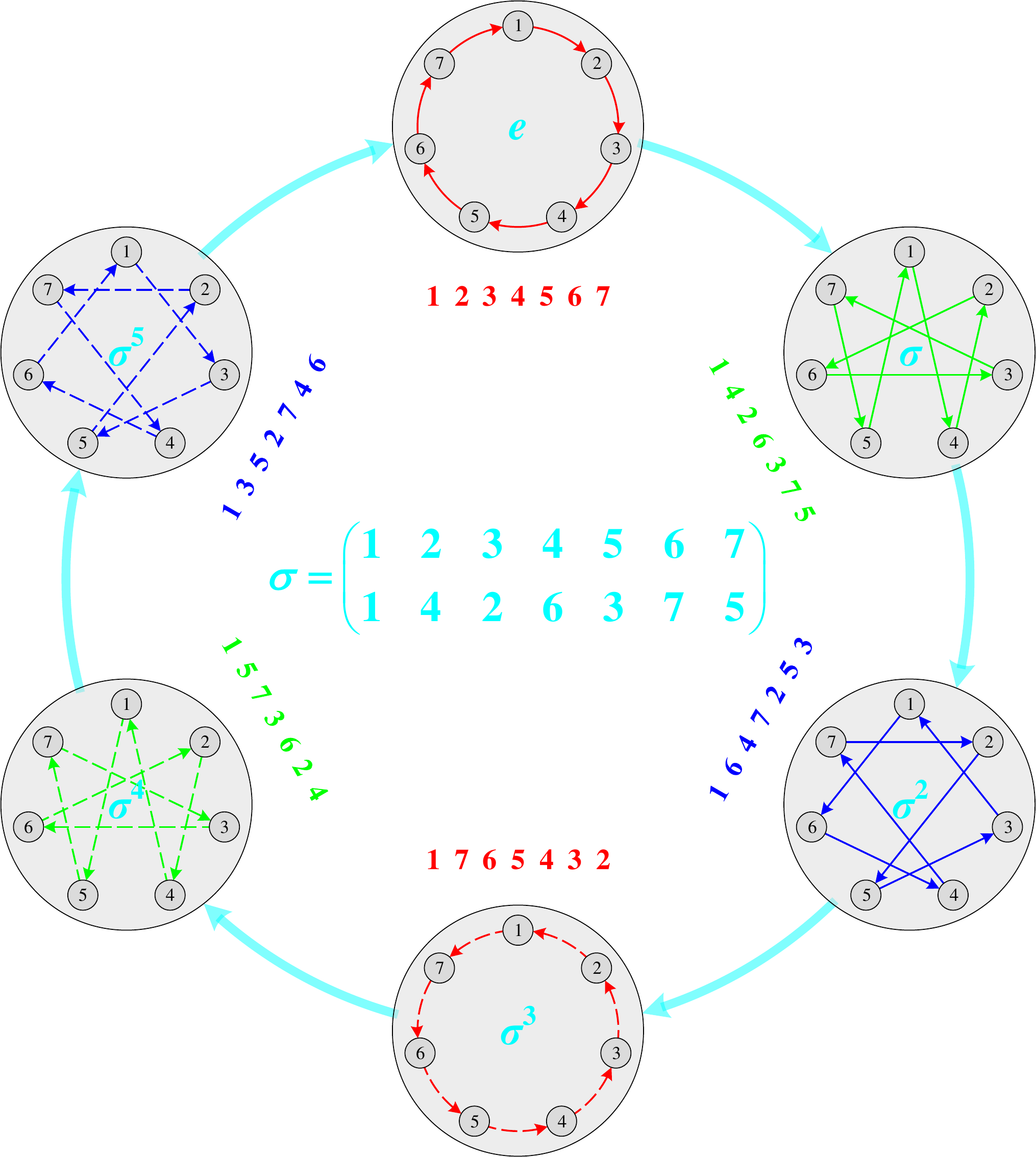}
	\label{fig:action:Z6}
	}
    \subfigure[$\mathfrak{G}_2 = \{ e, \sigma, \sigma^2, \ldots, \sigma^7 \} \cong \mathbb{Z}_8$]{
	\includegraphics[width=0.74\linewidth]{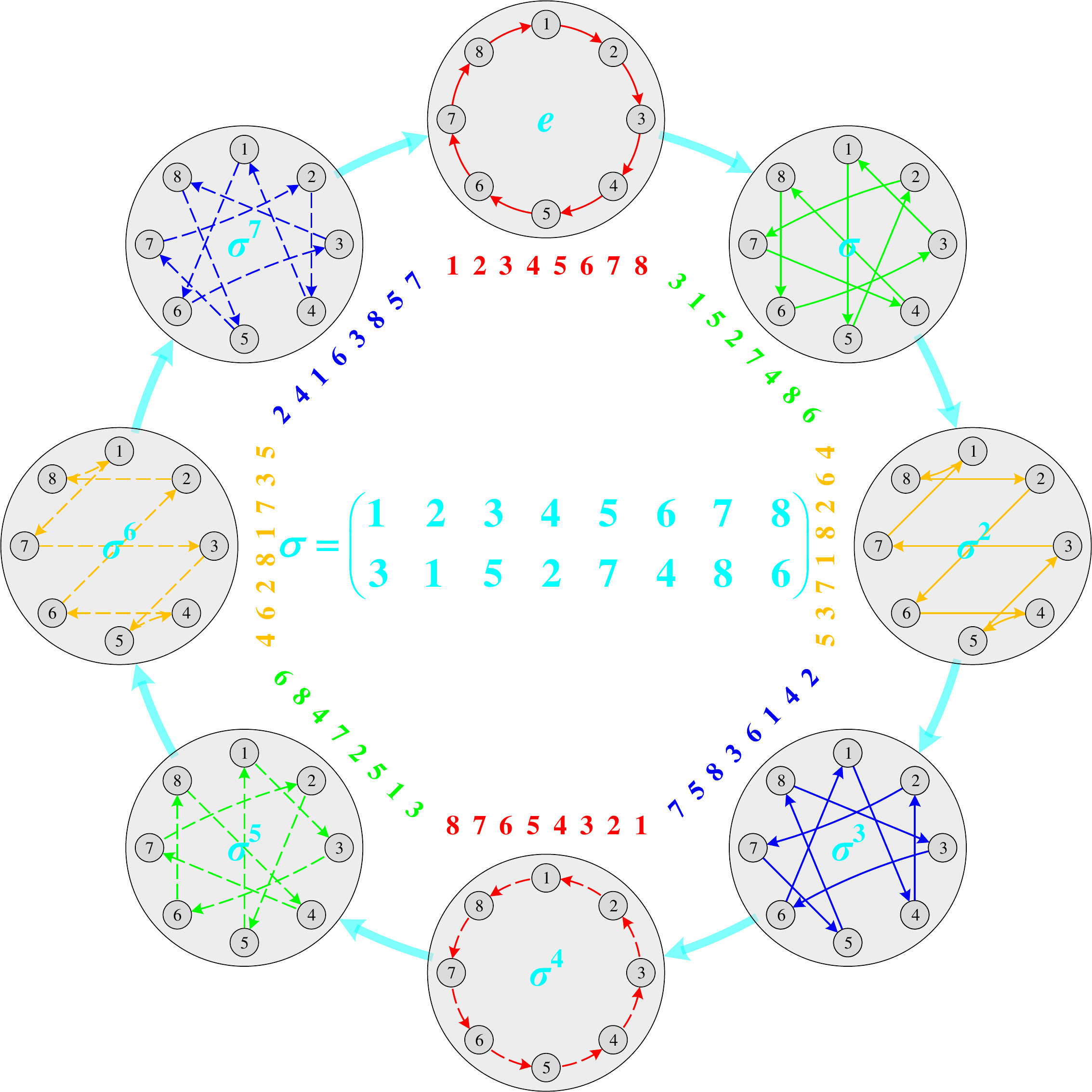}
	\label{fig:action:Z8}
	}
    \caption{The group structure of the permutation group $\mathfrak{G}$ and the results of its actions on the set $S$.}
    \label{fig:action}
\end{figure}

To better understand Lemma \ref{thm:cyclic} and Corollary \ref{thm:action}, we provide diagrams to illustrate the group actions of the permutation groups $\mathfrak{G}_1 = \{ e, \sigma, \sigma^2, \ldots, \sigma^5 \}$ and $\mathfrak{G}_2 = \{ e, \sigma, \sigma^2, \ldots, \sigma^7 \}$ when $n = 7$ and $n = 8$, respectively.
As shown in Figures \ref{fig:action:Z6} and \ref{fig:action:Z8}, the overall frameworks with big light-gray circles and cyan arrows represent the Cayley diagrams of the permutation groups $\mathfrak{G}_1 \cong \mathbb{Z}_6$ and $\mathfrak{G}_2 \cong \mathbb{Z}_8$ constructed by Lemma \ref{thm:cyclic}, respectively.
The center of each subfigure presents the corresponding generator $\sigma$.
Each big light-gray circle represents an element $g$ (i.e., a permutation) of group $\mathfrak{G}$, marked at the center of the circle.
And each cyan arrow $g_i \to g_j$ indicates the relationship $g_j = g_i \circ \sigma$ exists between two group elements $g_i, g_j \in \mathfrak{G}$.
After $g$ acts on the elements $1, \ldots, n$ of the set $S$, the corresponding images are presented as the colored numbers next to the big light-gray circle.
Finally, the 2-ary dependencies (colored arrows) between neighboring nodes (small dark-gray circles) are modeled according to the action results of $g$, shown in each big light-gray circle.

\section{Proofs About Incidence Triangles}
\label{sec:pf_triangle}

\subsection{Proof of Eq.~(\ref{eq:triangle})}
\label{sec:triangle}

\begin{equation*}
    \bm \tau = \frac{1}{2} \bm A^2 \odot \bm A \cdot \bm 1_N, \quad 
    \bm 1_N^\top \bm \tau = \frac{1}{2} \text{tr}(\bm A^3)
\end{equation*}

\begin{proof}
Let $\bm A = (a_{ij})_{N \times N}$, $\bm B = \bm A^2 = (b_{ij})_{N \times N}$, where $a_{ij}$ and $b_{ij}$ denote the $(i, j)$ element of $\bm A$ and $\bm B$, respectively.
Since $a_{ij}$ equals 1 iff nodes $v_i$ and $v_j$ are adjacent in $G$, $b_{ij}$ equals the number of walks of length 2 from nodes $v_i$ to $v_j$ in $G$.
In addition, a walk of length 2 from $v_i$ to $v_j$ and an edge from $v_j$ to $v_i$ form a triangle containing both $v_i$ and $v_j$.
Therefore, the $(i, j)$ element of $\bm A^2 \odot \bm A$ equals $b_{ij}a_{ij} = b_{ij}a_{ji}$, which indicates how many triangles contain both $v_i$ and $v_j$.
We can use a sum vector $\bm 1_N = (1, 1, \cdots, 1)^\top \in \mathbb{R}^N$ to sum up each row of $\bm A^2 \odot \bm A$ and get a result vector, whose $i$-th element gives \textbf{twice} the number of incidence triangles of node $v_i$.
Here the ``twice'' comes from the fact that each incidence triangle $\triangle v_jv_iv_k$ over node $v_i$ has two walks of length 2 starting from node $v_i$, that is, $v_i \to v_j \to v_k$ and $v_i \to v_k \to v_j$.
Hence after dividing each element of the result vector by 2, we finally obtain $ \bm \tau = \frac{1}{2} \bm A^2 \odot \bm A \cdot \bm 1_N$.

For the second equation, we have
\begin{equation*}
    \bm 1_N^\top \bm \tau = \frac{1}{2} \bm 1_N^\top \cdot (\bm A^2 \odot \bm A) \cdot \bm 1_N = \frac{1}{2} \sum_{i=1}^N \sum_{j=1}^N {b_{ij}a_{ij}} = \frac{1}{2} \sum_{i=1}^N (\bm A^2 \cdot \bm A^\top)_{ii} = \frac{1}{2} \text{tr}(\bm A^2 \cdot \bm A^\top) = \frac{1}{2} \text{tr}(\bm A^3)
\end{equation*}
\end{proof}

\textbf{Remark.}\enspace The $i$-th diagonal entry of $\bm A^3$ is equal to twice the number of triangles in which the $i$-th node is contained \cite{mitzenmacher2017probability}.
In addition, each triangle has three vertices.
Hence we can divide the sum of the diagonal entries by 6 to obtain the total number of triangles in graph $G$, i.e., $\frac{1}{6} \text{tr}(\bm A^3)$ \cite{harary1971number}.

For directed graphs, we also have similar results:
\begin{equation*}
    \bm{\vec \tau} = \bm A^2 \odot \bm A^\top \cdot \bm 1_N, \quad 
    \bm 1_N^\top \bm{\vec \tau} = \text{tr}(\bm A^3)
\end{equation*}
where $\bm{\vec \tau} \in \mathbb{R}^N$ and its $i$-th element $\vec \tau_i$ denotes the number of \emph{directed} incidence triangles over node $i$.

\subsection{Proof of Theorem \ref{thm:triangle}}
\label{sec:incidence}

\begin{apptheorem}[Chernoff-Hoeffding Bound for Discrete Time Markov Chain \cite{chung2012chernoff}]
\label{thm:mixdeviation}
	Let $\mathcal{M}$ be an ergodic Markov chain with state space $[n] = \{1, 2, \ldots, n\}$ and stationary distribution $\bm \pi$.
	Let $\mathfrak{t} = \mathfrak{t}(\epsilon)$ be its $\epsilon$-mixing time for $\epsilon \leq 1/8$. Let $(X_1, X_2, \dots, X_r)$ denote an $r$-step random walk on $\mathcal{M}$ starting from an initial distribution $\varphi$ on $[n]$, i.e., $X_1 \leftarrow \varphi$.
	Define $\left\| \varphi \right\|_{\bm\pi} = \sum_{i=1}^{n} {\frac{\varphi_i^2}{\pi_i}}$.
	For every step $k \in [r]$, let $f^{(k)}: [n] \rightarrow [0,1]$ be a weight function such that the expectation $\mathbb{E}_{X_k \leftarrow \bm\pi}[f^{(k)}(X_k)] = \mu$ for all $k$.
	Define the total weight of the walk $(X_1, X_2, \ldots, X_r)$ by $Z \triangleq \sum_{k=1}^r f(X_k)$.
	There exists some constant $c$ (which is independent of $\mu$, $\delta$ and $\epsilon$) such that for $0 < \delta < 1$
    \begin{equation*}
    \Pr\left( \left| Z - \mu r \right| > \epsilon \mu r \right) \leq c \left\| \varphi \right\|_{\bm\pi} \exp{\left( - \frac{\epsilon^2 \mu r}{72\mathfrak{t}} \right)}
    \end{equation*}
    or equivalently
    \begin{equation*}
    \Pr\left( \left| \frac{Z}{r} - \mu \right| > \epsilon \mu \right) \leq c \left\| \varphi \right\|_{\bm\pi} \exp{\left( - \frac{\epsilon^2 \mu r}{72\mathfrak{t}} \right)} .
    \end{equation*}
\end{apptheorem}

\begin{apptheorem}
\label{thm:RNN}
    Any nonlinear dynamic system may be approximated by a recurrent neural network to any desired degree of accuracy and with no restrictions imposed on the compactness of the state space, provided that the network is equipped with an adequate number of hidden neurons \cite{haykin2010neural}.
\end{apptheorem}

Using Theorem \ref{thm:mixdeviation} and Theorem \ref{thm:RNN}, we prove Theorem \ref{thm:triangle} as follows.

\def\theoremautorefname{Theorem}
\textbf{\autoref{thm:triangle}.}
\emph{Let $x_v, \forall v \in \mathcal{V}$ denote the feature inputs on graph $G = (\mathcal{V}, \mathcal{E})$, and $\mathbb{M}$ be a general GNN model with RNN aggregators.
Suppose that $x_v$ is initialized as the degree $d_v$ of node $v$, and each node is distinguishable.
For any $0 < \epsilon \le 1/8$ and $0 < \delta < 1$, there exists a parameter setting $\Theta$ for $\mathbb{M}$ so that after $\mathcal{O} \left( \frac{d_v(2d_v+\tau_v)\mathfrak{t}} {d_v+\tau_v} \right)$ samples,
\begin{equation*}
    \Pr \left( \left| \frac{z_v}{\tau_v} - 1 \right| \le \epsilon \right) \ge 1-\delta, \forall v \in \mathcal{V},
\end{equation*}
where $z_v \in \mathbb{R}$ is the final output value generated by $\mathbb{M}$ and $\tau_v$ is the number of incidence triangles.}

\begin{proof}
Without loss of generality, we discuss how to estimate the number of incidence triangles $\tau_0$ for an arbitrary node $v_0$ based on its $n$ neighbors $v_1^{\prime}, v_2^{\prime}, \ldots, v_n^{\prime}$.
Let $v_0^{\prime} = v_0$, and let $G' = (\mathcal{V}', \mathcal{E}')$ denote the subgraph induced by $\mathcal{V}' = \{v_0^{\prime}, v_1^{\prime}, v_2^{\prime}, \ldots, v_n^{\prime}\}$, with an adjacency matrix $\bm A' \in \mathbb{R}^{(n+1) \times (n+1)}$.
We add a symbol ``$\prime$'' to all notations of the induced subgraph $G'$ to distinguish them from those of graph $G$.
For each node $v_i^{\prime} \in \mathcal{V}'$, $d_i^{\prime}$ denotes the degree of $v_i^{\prime}$ in graph $G'$, $\tau_i^{\prime}$ denotes the number of incidence triangles of $v_i^{\prime}$ in graph $G'$.
In particular, $d_0^{\prime} = d_0 = n$, $\tau_0^{\prime} = \tau_0$.
Our goal is to estimate $\tau_0$ for an arbitrary node $v_0$ in graph $G$, which is equal to $\tau_0^{\prime}$ in graph $G'$.

A \emph{simple random walk} (SRW) with $r$ steps on graph $G'$, denoted by $R = (X_1, X_2, \ldots, X_r)$, is defined as follows:
start from an initial node in $G'$, then move to one of its neighboring nodes chosen uniformly at random, and repeat this process $(r-1)$ times.
This random walk on graph $G'$ can be viewed as a finite Markov chain $\mathcal{M}$ with the state space $\mathcal{V}'$, and the transition probability matrix $\bm P$ of this Markov chain is defined as
\begin{equation*}
\bm P(i,j) = 
    \begin{cases}
    \dfrac{1}{d_i^{\prime}}, & \text{if} \ (v_i^{\prime}, v_j^{\prime}) \in \mathcal{E}', \\
    0, & \text{otherwise.}
    \end{cases}
\end{equation*}
Let $D' = \sum_{i=0}^{n} d_i^{\prime} = 2 \left| \mathcal{E}' \right|$ denote the sum of degrees in graph $G'$.
After many random walk steps, the probability $\Pr(X_r = v_i^{\prime})$ converges to $p_i \triangleq d_i^{\prime} / D'$, and the vector $\bm\pi = (p_0, p_1, \ldots, p_n)$ is called the \emph{stationary distribution} of this random walk.

The \emph{mixing time} of a Markov chain is the number of steps it takes for a random walk to approach its stationary distribution.
We adopt the definition in \cite{chung2012chernoff, chen2016general, mitzenmacher2017probability} and define the mixing time $\mathfrak{t}(\epsilon)$ as follows:
\begin{equation*}
    \mathfrak{t}(\epsilon) = \max_{X_i \in \mathcal{V}'} \min \left\{ t: \left| \bm\pi - \bm\pi^{(i)} \bm P^t \right| < \epsilon \right\}
\end{equation*}
where $\bm\pi$ is the stationary distribution of the Markov chain defined above, $\bm\pi^{(i)}$ is the initial distribution when starting from state $X_i \in \mathcal{V}'$, $\bm P^t$ is the transition matrix after $t$ steps, and $|\cdot|$ is the variation distance between two distributions.

Later on, we will exploit node samples taken from a random walk to construct an estimator $z_0$, then use the mixing time based Chernoff-Hoeffding bound \cite{chung2012chernoff} to compute the number of steps/samples needed, thereby guaranteeing that our estimator $z_0$ is within $(1 \pm \epsilon)$ of the true value $\tau_0$ with the probability of at least $1 - \delta$.

Given a random walk $(X_1, X_2, \ldots, X_r)$ on graph $G'$, we define a new variable $a_k = \bm A_{X_{k-1}, X_{k+1}}^{\prime}$ for every $2 \le k \le r-1$, then we have
\begin{align}
\label{eq:a_kweighted}
    \mathbb{E}\left[a_k d_{X_k}^{\prime}\right]
    &= \sum_{i=0}^{n} {p_i \mathbb{E}\left[a_k d_{X_k}^{\prime} \ \big| \ X_k = v_i^{\prime} \right]} \nonumber \\
    &= \sum_{i=0}^{n} {\frac{d_i^{\prime}}{D'} \frac{2\tau_i^{\prime}}{{d_i^{\prime}}^2} d_i^{\prime}} \nonumber \\
    &= \frac{2}{D'} \sum_{i=0}^{n} {\tau_i^{\prime}}
\end{align}
The second equality holds because there are ${d_i^{\prime}}^2$ equal probability combinations of $(X_{k-1}, v_i^{\prime}, X_{k+1})$, out of which only $2\tau_i^{\prime}$ combinations form a triangle $(u', v_i^{\prime}, w')$ or its reverse $(w', v_i^{\prime}, u')$, where $u'$ is connected to $w'$, i.e., $a_k = \bm A_{X_{k-1}, X_{k+1}}^{\prime} = \bm A_{u', w'}^{\prime} = 1$.

To estimate $\tau_0$, we introduce two variables $Y_1$ and $Y_2$, defined as follows:
\begin{equation*}
    Y_1 \triangleq \frac{1}{r-2}\sum_{k=2}^{r-1}{a_k d_{X_k}^{\prime}}, \quad
    Y_2 \triangleq \frac{1}{r}\sum_{k=1}^{r}{\frac{1}{d_{X_k}^{\prime}}}
\end{equation*}

Using the linearity of expectation and Eq.~\eqref{eq:a_kweighted}, we obtain
\begin{equation}
\label{eq:EY1_temp}
    \mathbb{E}[Y_1] = \frac{1}{r-2}\sum_{k=2}^{r-1} {\mathbb{E}\left[ a_k d_{X_k}^{\prime} \right]}
    = \frac{2}{D'} \sum_{i=0}^{n} {\tau_i^{\prime}}
\end{equation}

Similarly, we have
\begin{equation}
\label{eq:EY2_temp}
    \mathbb{E}[Y_2] = \frac{1}{r}\sum_{k=1}^{r} {\mathbb{E}\left[ \frac{1}{d_{X_k}^{\prime}} \right]}
    = \frac{1}{r}\sum_{k=1}^{r} {\left( \sum_{i=0}^{n} {\frac{d_i^{\prime}}{D'} \frac{1}{d_i^{\prime}} } \right)}
    = \frac{n+1}{D'}
\end{equation}

Recall that $G'$ is a subgraph induced by $\mathcal{V}' = \{v_0^{\prime}, v_1^{\prime}, v_2^{\prime}, \ldots, v_n^{\prime}\}$, where $v_1^{\prime}, v_2^{\prime}, \ldots, v_n^{\prime}$ are $n$ neighbors of an arbitrary node $v_0^{\prime} = v_0$.
Therefore, the maximum degree of graph $G'$ is $\Delta' = n$, which is equal to $d_0^{\prime} = d_0$.
In addition, we have $\sum_{i=0}^{n} {\tau_i^{\prime}} = 3\tau_0^{\prime} = 3\tau_0$, and $D' = 2 \left| \mathcal{E}' \right| = 2(d_0^{\prime} + \tau_0^{\prime}) = 2(d_0 + \tau_0)$.
Substituting them in Eq.~\eqref{eq:EY1_temp} and Eq.~\eqref{eq:EY2_temp}, we get
\begin{equation}
\label{eq:EY1}
    \mathbb{E}[Y_1] = \frac{3\tau_0}{d_0 + \tau_0}
\end{equation}
and
\begin{equation}
\label{eq:EY2}
    \mathbb{E}[Y_2] = \frac{d_0+1}{2(d_0 + \tau_0)}
\end{equation}

From Eq.~\eqref{eq:EY1} and Eq.~\eqref{eq:EY2} we can isolate $\tau_0$ and get
\begin{equation}
\label{eq:tau0}
    \tau_0 = \frac{d_0+1}{6} \cdot \frac{\mathbb{E}[Y_1]} {\mathbb{E}[Y_2]}
\end{equation}
Since $d_0$ is the feature input, the coefficient $\dfrac{d_0+1}{6}$ can be considered as a constant factor here.
Intuitively, both $Y_1$ and $Y_2$ converge to their expected values, and thus the estimator $z_0 \triangleq \dfrac{d_0+1}{6} \cdot \dfrac{Y_1}{Y_2}$ converges to $\tau_0$ as well.
Next, we will find the number of steps/samples $r$ for convergence.

Since $a_k d_{X_k}^{\prime} = \bm A_{X_{k-1}, X_{k+1}}^{\prime} d_{X_k}^{\prime}$ in $Y_1$ only depends on a 3-nodes history, we observe a related Markov chain $\tilde{\mathcal{M}}$ that remembers the three latest visited nodes.
Accordingly, $\tilde{\mathcal{M}}$ has $(n+1) \times (n+1) \times (n+1)$ states, and $(X_{k-1}, X_{k}, X_{k+1}) \to (X_{k}, X_{k+1}, X_{k+2})$ has the same transition probability as $X_{k+1} \to X_{k+2}$ in $\mathcal{M}$.
Define each state $\tilde{X}_k = (X_{k-1}, X_{k}, X_{k+1})$ for $2 \le k \le r-1$.
Let $\tilde{f}_1^{(k)}(\tilde{X}_k) = f_1^{(k)}(X_k) = \dfrac{a_k d_{X_k}^{\prime}}{\Delta'} = \dfrac{a_k d_{X_k}^{\prime}}{d_0}$ such that all values of $\tilde{f}_1^{(k)}(\tilde{X}_k)$ are in $[0, 1]$.
By Eq.~\eqref{eq:a_kweighted}, Eq.~\eqref{eq:EY1_temp}, and Eq.~\eqref{eq:EY1}, we have $\mu_1 = \mathbb{E}_{\tilde{X}_k \leftarrow \bm\pi}(\tilde{f}_1^{(k)}(\tilde{X}_k)) = \dfrac{3\tau_0}{d_0(d_0+\tau_0)}$.
Define $Z_1 \triangleq \sum\limits_{k=2}^{r-1} \tilde{f}_1^{(k)}(\tilde{X}_k) = \dfrac{r-2}{d_0}Y_1$, assume that $\varphi \approx \bm\pi$ thus $\left\| \varphi \right\|_{\bm\pi} = 1$.
By Theorem \ref{thm:mixdeviation} and Eq.~\eqref{eq:EY1}, we have
\begin{equation}
\label{eq:PY1}
    \Pr\left( \left| Y_1 - \mathbb{E}[Y_1] \right| > \frac{\epsilon}{3} \mathbb{E}[Y_1] \right) \leq c_1 \exp{\left( - \frac{3 \cdot \epsilon^2 \tau_0 (r-2)}{9 \cdot 72 \cdot \tilde{\mathfrak{t}} d_0(d_0+\tau_0)} \right)}
\end{equation}
Extracting $r_{Y_1}$ from $\dfrac{\delta}{2} = c_1 \exp{\left( - \dfrac{\epsilon^2 \tau_0 (r-2)}{216 \cdot \tilde{\mathfrak{t}} d_0(d_0+\tau_0)} \right)}$, we obtain $r_{Y_1} = 2 - 216 \dfrac{\ln(\delta/2c_1)}{\epsilon^2} \cdot \dfrac{d_0(d_0+\tau_0)\tilde{\mathfrak{t}}} {\tau_0} = \mathcal{O} \left( \dfrac{d_0(d_0+\tau_0)\tilde{\mathfrak{t}}} {\tau_0} \right)$, where $c_1$, $\epsilon$ and $\delta$ are all constants.

Let $f_2^{(k)}(X_k) = \dfrac{1}{d_{X_k}^{\prime}}$, by Eq.~\eqref{eq:EY2_temp} and Eq.~\eqref{eq:EY2} we have $\mu_2 = \mathbb{E}_{X_k \leftarrow \bm\pi}(f_2^{(k)}(X_k)) = \dfrac{d_0+1}{2(d_0 + \tau_0)}$.
Define $Z_2 \triangleq \sum\limits_{k=1}^{r} f_2^{(k)}(X_k) = rY_2$, assume that $\varphi \approx \bm\pi$ thus $\left\| \varphi \right\|_{\bm\pi} = 1$.
By Theorem \ref{thm:mixdeviation} and Eq.~\eqref{eq:EY2}, we have
\begin{equation}
\label{eq:PY2}
    \Pr\left( \left| Y_2 - \mathbb{E}[Y_2] \right| > \frac{\epsilon}{3} \mathbb{E}[Y_2] \right) \leq c_2 \exp{\left( - \frac{\epsilon^2 (d_0+1) r}{2 \cdot 9 \cdot 72 \cdot \mathfrak{t} (d_0 + \tau_0)} \right)}
\end{equation}
Extracting $r_{Y_2}$ from $\dfrac{\delta}{2} = c_2 \exp{\left( - \dfrac{\epsilon^2 (d_0+1) r}{1296 \cdot \mathfrak{t} (d_0 + \tau_0)} \right)}$, we obtain $r_{Y_2} = -1296 \dfrac{\ln(\delta/2c_2)}{\epsilon^2} \cdot \dfrac{(d_0+\tau_0)\mathfrak{t}} {d_0+1} = \mathcal{O} \left( \dfrac{(d_0+\tau_0)\mathfrak{t}} {d_0+1} \right)$, where $c_2$, $\epsilon$ and $\delta$ are all constants.

Since $\mathfrak{t} \ge \tilde{\mathfrak{t}}$ (see Appendix A in \cite{hardiman2013estimating} for details), choose $r \ge \mathcal{O} \left( \dfrac{d_0(d_0+\tau_0)\mathfrak{t}} {\tau_0} \right) \ge \max \{ r_{Y_1}, r_{Y_2} \}$.
Eq.~\eqref{eq:PY1} and Eq.~\eqref{eq:PY2} find the number of steps/samples $r$, which guarantees both $Y_1$ and $Y_2$ are within $(1 \pm \epsilon/3)$ of their expected values with the probability of at least $1 - \delta/2$.
Since the probability of $Y_1$ or $Y_2$ deviating from their expected value is at most $\delta/2$, the probability of either $Y_1$ or $Y_2$ deviating is at most $\delta$:
\begin{align*}
    & \Pr\left( \left| Y - \mathbb{E}[Y] \right| > \frac{\epsilon}{3} \mathbb{E}[Y] \right) \leq \frac{\delta}{2}, \quad Y = Y_1, Y_2 \\
    \Rightarrow & \Pr\left(
    \left( 1 - \frac{\epsilon}{3} \right) \mathbb{E}[Y] \leq Y \leq \left( 1 + \frac{\epsilon}{3} \right) \mathbb{E}[Y]
    \right) \geq 1 - \frac{\delta}{2}, \quad Y = Y_1, Y_2 \\
    \Rightarrow & \Pr\left( 
    \underbrace{ (1 - \epsilon) \tau_0 \leq
    \dfrac{d_0 + 1}{6} \frac{1 - \frac{\epsilon}{3}} {1 + \frac{\epsilon}{3}} \frac{\mathbb{E}[Y_1]} {\mathbb{E}[Y_2]} }_\star \leq
    \underbrace{ \dfrac{d_0 + 1}{6} \frac{Y_1}{Y_2} }_{\text{estimator}~z_0} \leq
    \underbrace{ \dfrac{d_0 + 1}{6} \frac{1 + \frac{\epsilon}{3}} {1 - \frac{\epsilon}{3}} \frac{\mathbb{E}[Y_1]} {\mathbb{E}[Y_2]} \leq
    (1 + \epsilon) \tau_0 }_\star
    \right) \geq 1 - \delta
\end{align*}
The first line is a summary of Eq.~\eqref{eq:PY1} and Eq.~\eqref{eq:PY2}.
The inequalities ``$\star$'' hold due to Eq.~\eqref{eq:tau0}, and the fact of both $1 - \epsilon \leq \dfrac{1 - \frac{\epsilon}{3}} {1 + \frac{\epsilon}{3}}$ and $1 + \epsilon \geq \dfrac{1 + \frac{\epsilon}{3}} {1 - \frac{\epsilon}{3}}$ when $0 < \epsilon \leq 1/8$.
We thus conclude that after $\mathcal{O} \left( \dfrac{d_0(d_0+\tau_0)\mathfrak{t}} {\tau_0} \right)$ samples, $\Pr \left( \left| \dfrac{z_0}{\tau_0} - 1 \right| \le \epsilon \right) \ge 1-\delta$.

\begin{figure}
    \centering
    \subfigure[Star graph, without any triangles]{
	\includegraphics[width=0.32\linewidth]{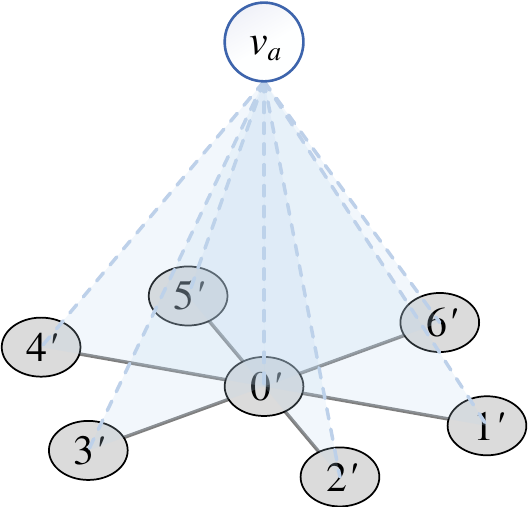}
	}
	\qquad
    \subfigure[General case, with some triangles]{
	\includegraphics[width=0.32\linewidth]{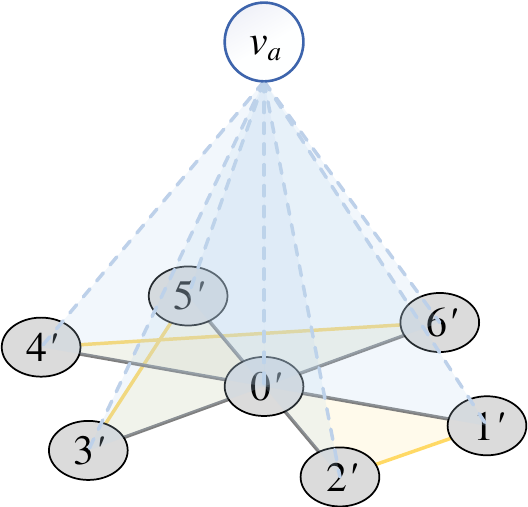}
	}
    \caption{Add an artificial node $v_a$ and connect it to all nodes in $G'$.}
    \label{fig:artificial}
\end{figure}

However, if $\tau_0 = 0$ and $G'$ is a star graph, the number of samples $r \ge \mathcal{O} \left( \dfrac{d_0(d_0+\tau_0)\mathfrak{t}} {\tau_0} \right) \to \infty$.
To avoid that, we add an artificial node $v_a$ and connect it to all nodes in $G'$, as illustrated in Figure \ref{fig:artificial}.
Since $d_0^{(a)} = d_0 + 1$, $\tau_0^{(a)} = d_0 + \tau_0$, we only need to minus a $d_0$ for the estimated result $\tau_0^{(a)}$, and the number of samples can then be reduced to $\mathcal{O} \left( \dfrac{d_0^{(a)}(d_0^{(a)}+\tau_0^{(a)})\mathfrak{t}^{(a)}} {\tau_0^{(a)}} \right) \approx \mathcal{O} \left( \dfrac{d_0(2d_0+\tau_0)\mathfrak{t}} {d_0+\tau_0} \right)$.

We have proved that we can estimate the number of incidence triangles $\tau_0$ for an arbitrary node $v_0$ based on its $n$ neighbors by a random walk.
Consider the random walk as a nonlinear dynamic system, according to the RNNs' universal approximation ability (Theorem \ref{thm:RNN}), this random walk can be approximated by an RNN to any desired degree of accuracy.
Therefore, let the input sequence of RNN follow the random walk above, then the RNN aggregator can mimic this random walk on the subgraph induced by $v_0$ and its 1-hop neighbors when aggregating, finally outputs $z_0 \approx \tau_0$.
This completes the proof.
\end{proof}

Note: This proof is inspired by \citet{hardiman2013estimating} and \citet{chen2016general}.

\subsection{Analysis of GraphSAGE}
\label{sec:GraphSAGE}

\begin{apptheorem}
\label{thm:GraphSAGE}
	Let $x_v \in U, \forall v \in \mathcal{V}$ denote the input features for Algorithm 1 (proposed in GraphSAGE) on graph $G = (\mathcal{V}, \mathcal{E})$, where $U$ is any compact subset of $\mathbb{R}^d$.
	Suppose that there exists a fixed positive constant $C \in \mathbb{R}^+$ such that $\| x_v - x_{v'} \|_2 > C$ for all pairs of nodes.
	Then we have that $\forall \epsilon > 0$ there exists a parameter setting $\Theta^*$ for Algorithm 1 such that after $K = 4$ iterations
    \begin{equation*}
    \left| z_v - c_v \right| < \epsilon, \forall v \in \mathcal{V},
    \end{equation*}
    where $z_v \in \mathbb{R}$ are final output values generated by Algorithm 1 and $c_v$ are node clustering coefficients \cite{hamilton2017inductive}.
\end{apptheorem}

According to Theorem \ref{thm:GraphSAGE}, GraphSAGE can approximate the clustering coefficients in a graph to arbitrary precision.
In addition, since GraphSAGE with LSTM aggregators is a special case of our proposed Theorem \ref{thm:triangle}, it can also approximate the number of incidence triangles to arbitrary precision.
In fact, the number of incidence triangles $\tau_v$ is related to the local clustering coefficient $c_v$.
More specifically, $\tau_v = c_v \cdot d_v (d_v-1) / 2$.
Therefore, the conclusion of Theorem \ref{thm:triangle} is consistent with that of Theorem \ref{thm:GraphSAGE}.
However, Theorem \ref{thm:triangle} reveals that the required samples $\mathcal{O} \left( \frac{d_v(2d_v+\tau_v)\mathfrak{t}} {d_v+\tau_v} \right)$ are related to $\tau_v$ and proportional to the mixing time $\mathfrak{t}$, leading to a practically prohibitive aggregation complexity.

To overcome this problem and improve the efficiency, GraphSAGE performs neighborhood sampling and suggests sampling 2-hop neighborhoods for each node.
Suppose the neighborhood sample sizes of 1-hop and 2-hop are $S_1$ and $S_2$, then the sampling complexity is $\Theta(NS_1S_2)$.
Accordingly, the memory and time complexity of GraphSAGE with LSTM are $\Theta(Nc + NS_1S_2)$ and $\Theta(NS_1S_2c^2 + NS_1S_2)$.

\section{Proof of Proposition \ref{thm:express}}
\label{sec:pf_express}

\begin{apptheorem}
\label{thm:2-WL}
	$2$-WL and MPNNs cannot induced-subgraph-count any connected pattern with $3$ or more nodes \cite{chen2020can}.
\end{apptheorem}

\begin{applemma}
\label{thm:srg}
    No pair of strongly regular graphs in family SRG($v, r, \lambda, \mu$) can be distinguished by the 2-FWL test \cite{bodnar2021weisfeilers, bouritsas2022improving}.
\end{applemma}

Using Theorem \ref{thm:2-WL} and Lemma \ref{thm:srg}, we prove Proposition \ref{thm:express} as follows.

\def\theoremautorefname{Proposition}
\textbf{\autoref{thm:express}.}
\emph{PG-GNN is strictly more powerful than the 2-WL test and not less powerful than the 3-WL test.}

\begin{proof}
We first verify that the GIN (with the equivalent expressive power as the 2-WL test) \cite{xu2019powerful} can be instantiated by a GNN model with RNN aggregators (including our proposed PG-GNN).
Consider a single layer of GIN:
\begin{equation}
\label{eq:GIN}
	\bm h_v^{(k)} = \text{MLP}^{(k)} \left( \bm h_v^{(k-1)} + \sum\nolimits_{u \in \mathcal{N}(v)} \bm h_u^{(k-1)} \right)
\end{equation}
where $\text{MLP}^{(k)}$ has a linear mapping $\bm W_{\text{GIN}}^{(k)} \in \mathbb{R}^{d_k \times d_{k-1}}$ and a bias term $\bm b_{\text{GIN}}^{(k)} \in \mathbb{R}^{d_k}$.
Without loss of generality, we take the Simple Recurrent Network (SRN) \cite{elman1990finding} as the RNN aggregator in Eq.~\eqref{eq:agg}, formulated as follows:
\begin{align*}
    \bm z_t^{(k)} &= \bm U \bm y_{t-1}^{(k)} + \bm W \bm h_t^{(k-1)} + \bm b \\
    \bm y_t^{(k)} &= a(\bm z_t^{(k)})
\end{align*}
Let $\bm W = \bm W_{\text{GIN}}^{(k)}$, $\bm U = \bm I_{d_k}$, $\bm b = \bm b_{\text{GIN}}^{(k)}$, the initial state $\bm y_0^{(k)} = \mathbf{0}$, the activation function $a(\cdot)$ be an identity function.
And let the input sequence of the RNN aggregator be an arbitrarily ordered sequence of the set $\{ \bm h_u^{(k-1)} \}_{u \in \mathcal{N}(v) \cup v}$.
Then any GIN with Eq.~\eqref{eq:GIN} can be instantiated by a GNN model with RNN aggregators (in particular, a PG-GNN with Eq.~\eqref{eq:agg}), which implies that the permutation-sensitive GNNs can be at least as powerful as the 2-WL test.

Next, we prove that PG-GNN is strictly more powerful than MPNNs and 2-WL test from the perspective of substructure counting.
Without loss of generality, we take an arbitrary node $v$ into consideration.
According to the definition of incidence triangles and the fact that they always appear in the 1-hop neighborhood of the central node, the number of connections between neighboring nodes of the central node $v$ is equivalent to the number of incidence triangles over $v$.
Theorem \ref{thm:arrange} ensures that all the 2-ary dependencies can be modeled by Eq.~\eqref{eq:agg}.
Suppose we are aiming to capture the connections between two arbitrary neighbors of the central node, we can use an MSE loss to measure the mean squared error between the predicted and ground-truth counting values and guide our model to learn the correct 2-ary dependencies, thereby capturing the correct connections and counting the number of connections between neighboring nodes.
And if we mainly focus on specific downstream tasks (e.g., graph classification), these 2-ary dependencies will be learned adaptively with the guidance of a specific loss function (e.g., cross-entropy loss).
Thus PG-GNN is capable of counting incidence triangles\footnote{In fact, since PG-GNN can count incidence triangles, it is also capable of counting all incidence 3-node graphlets.
There are only two types of 3-node graphlets, i.e., wedges (\scalebox{0.32}{\GTHREE{1}}) and triangles (\scalebox{0.32}{\GTHREE{2}}), let $\tau_v$ be the number of incidence triangles over $v$ and $n$ be the number of 1-hop neighbors, then we have $\binom{n}{2} - \tau_v$ incidence wedges.}.
Moreover, since the incidence 4-cliques always appear in the 1-hop neighborhood of the central node and every 4-clique is entirely composed of triangles, PG-GNN can also leverage 2-ary dependencies to count incidence 4-cliques, similar to counting incidence triangles.
Thus PG-GNN can count all 3-node graphlets (\scalebox{0.4}{\GTHREE{1}}, \scalebox{0.4}{\GTHREE{2}}), even 4-cliques (\scalebox{0.4}{\GCLIQUE}) incident to node $v$.

In addition, \citet{chen2020can} proposed Theorem \ref{thm:2-WL}, which implies that 2-WL and MPNNs cannot count any connected induced subgraph with 3 or more nodes.
% (i.e., $k$-node graphlets, $k \ge 3$)
Since the incidence wedges, triangles, and 4-cliques are all connected induced subgraphs with $\ge 3$ nodes, the above arguments demonstrate that the expressivity of PG-GNN goes beyond the 2-WL test and MPNNs.

\begin{figure}
    \centering
    \includegraphics[width=0.6\linewidth]{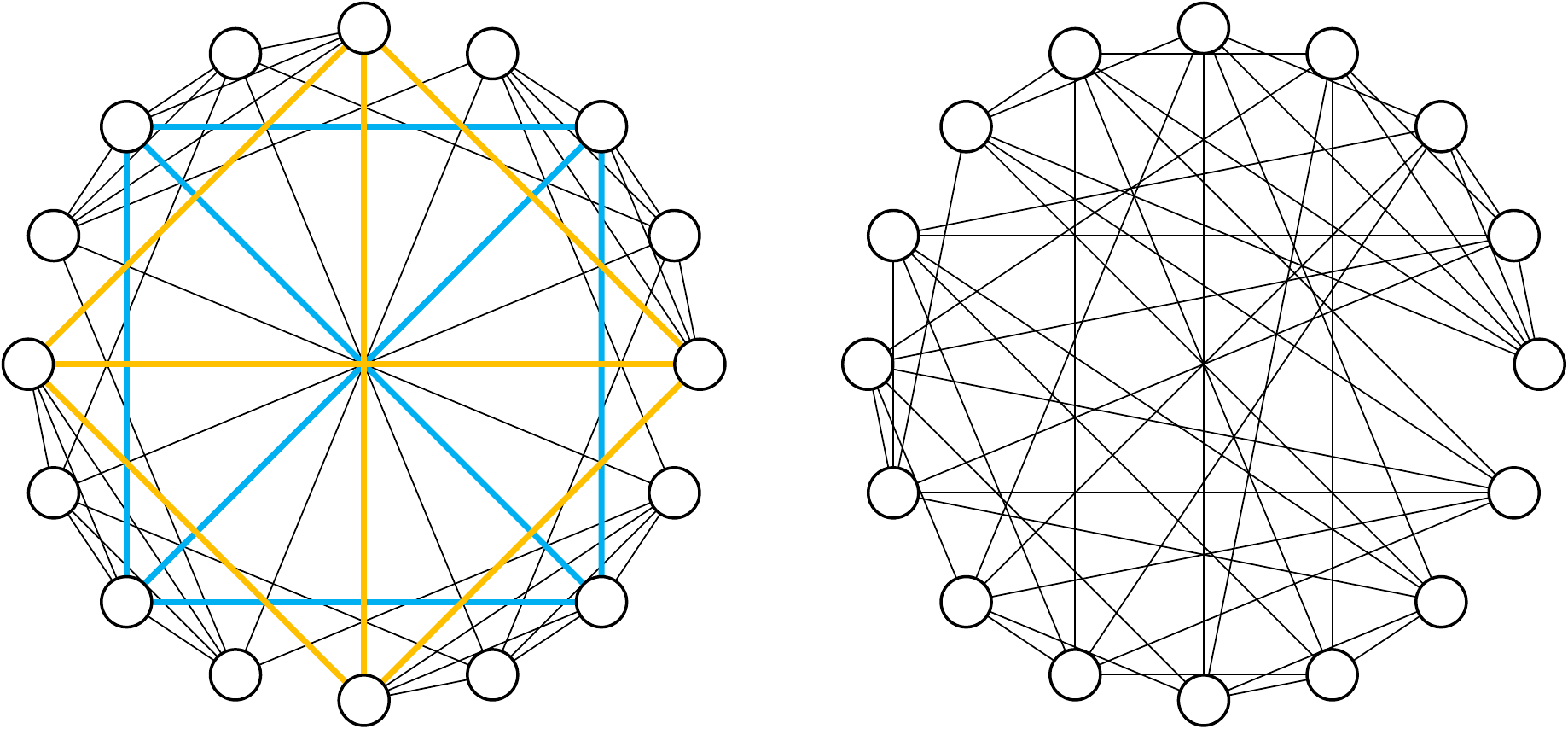}
    \caption{A pair of non-isomorphic strongly regular graphs in the family SRG(16,6,2,2): 4$\times$4 Rook’s graph and the Shrikhande graph.}
    \label{fig:SRG}
\end{figure}

To round off the proof, we finally prove that PG-GNN is not less powerful than the 3-WL test.
Consider a pair of strongly regular graphs in the family SRG(16,6,2,2): 4$\times$4 Rook’s graph and the Shrikhande graph.
As illustrated in Figure \ref{fig:SRG}, only Rook’s graph (left) possesses 4-cliques (some are emphasized by colors), but the Shrikhande graph (right) possesses no 4-cliques.
Since PG-GNN is capable of counting incidence 4-cliques, our approach can distinguish this pair of strongly regular graphs.
However, in virtue of Lemma \ref{thm:srg} and the fact that 2-FWL is equivalent to 3-WL \cite{maron2019provably}, the 3-WL test fails to distinguish them.
Thus PG-GNN is not less powerful than the 3-WL test\footnote{More accurately, PG-GNN is outside the WL hierarchy, and thus it is not easy to fairly compare it with 3-WL.
On the one hand, PG-GNN can distinguish some strongly regular graphs but 3-WL fails.
On the other hand, 3-WL considers all the 3-tuples $(i_1, i_2, i_3) \in \mathcal V^3$, which form a superset of (induced) subgraphs, but PG-GNN only considers the induced subgraphs and thus cannot completely achieve 3-WL.
In summary, 3-WL and PG-GNN have their own unique merits.
However, since 3-WL needs to consider all $\binom{N}{3} = \Theta(N^3)$ 3-tuples, the problem of complexity is inevitable.
In contrast, PG-GNN breaks from the WL hierarchy to make a trade-off between expressive power and computational efficiency.}.

In conclusion, our proposed PG-GNN is strictly more powerful than the 2-WL test and not less powerful than the 3-WL test.
\end{proof}

\section{Details of the Proposed Model}

In this section, we discuss the proposed model in detail.
The notations follow the definitions in Section \ref{sec:pre}, i.e., let $n$ denote the number of 1-hop neighbors of the central node $v$.
Suppose these $n$ neighbors are randomly numbered as $u_1, \ldots, u_n$ (also abbreviated as $1, \ldots, n$ for simplicity), the set of neighboring nodes is represented as $\mathcal{N}(v)$ (or $S = [n] = \{1, \ldots, n\}$).

\subsection{Illustration of the Proposed Model}
\label{sec:illustration}

\begin{figure}
    \centering
    \subfigure[$n = 5$]{
    \includegraphics[width=1.0\linewidth]{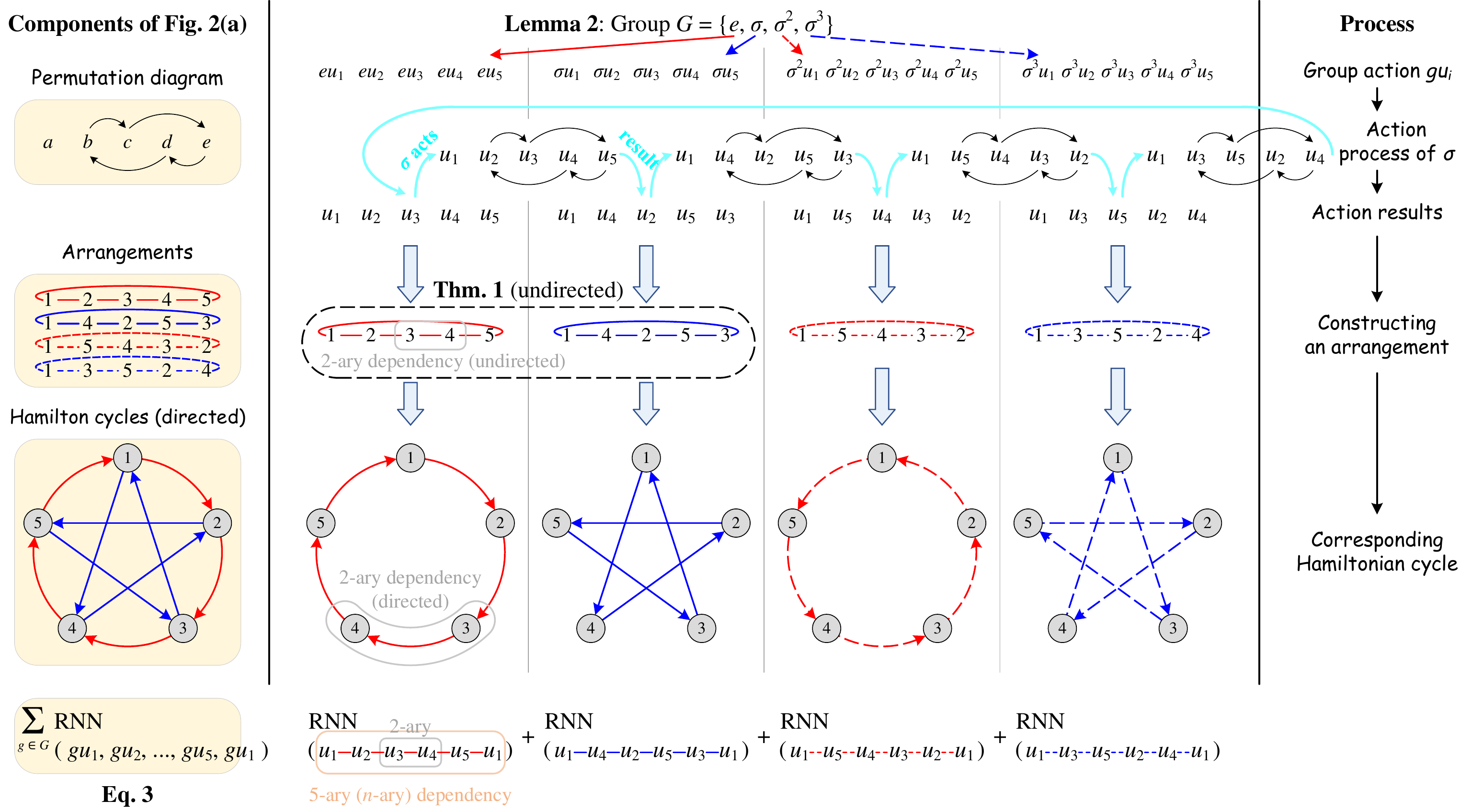}
    \label{fig:illus:K5}
    }
    \subfigure[$n = 6$]{
    \includegraphics[width=1.0\linewidth]{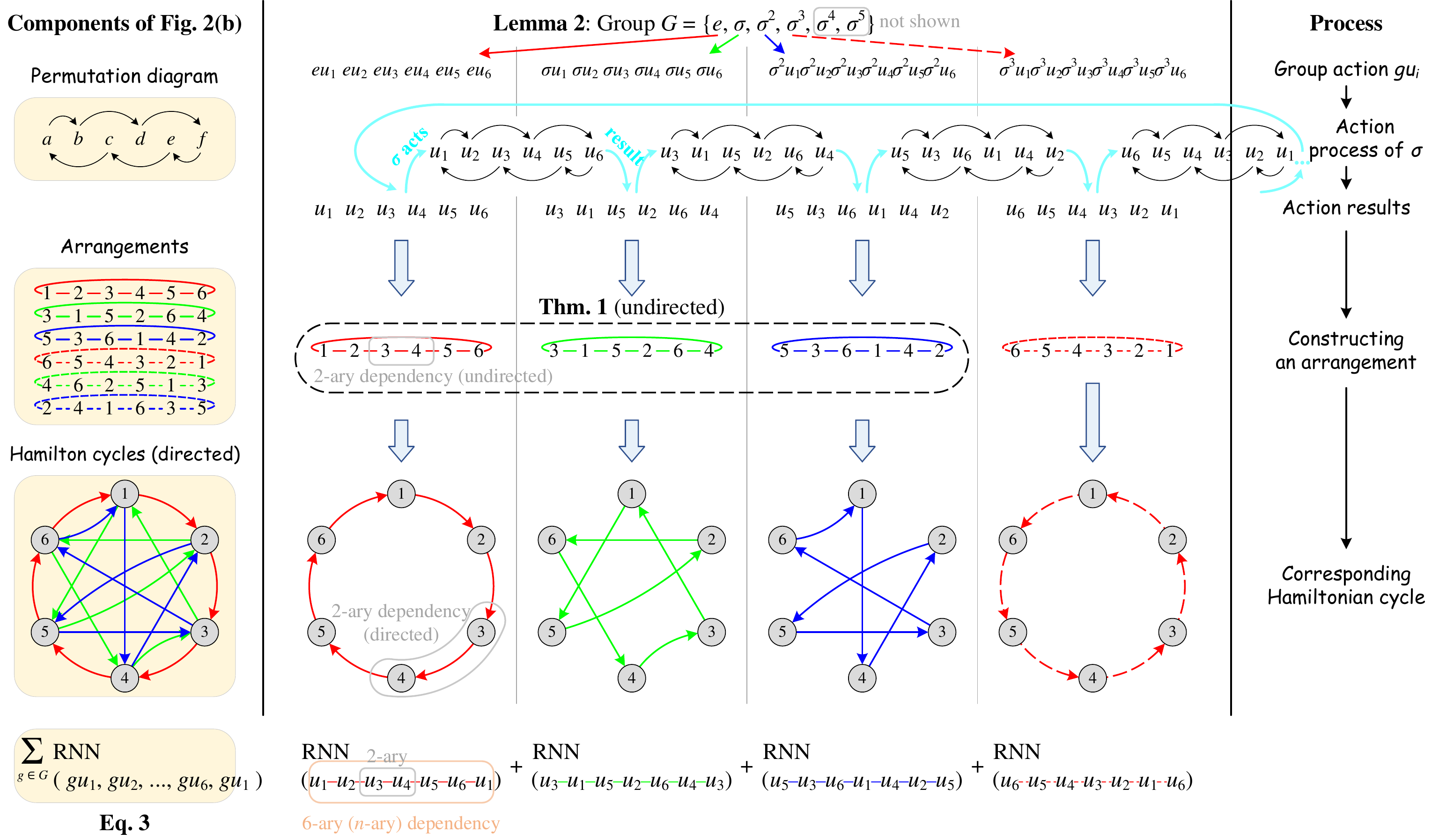}
    \label{fig:illus:K6}
    }
    \caption{Illustration of the proposed PG-GNN model.}
    \label{fig:illustration}
\end{figure}

Figure \ref{fig:illustration} presents a further explanation of Figure \ref{fig:complete} and the relationships among Theorem \ref{thm:arrange}, Lemma \ref{thm:cyclic}, Corollary \ref{thm:action}, Figure \ref{fig:complete}, and Eq.~\eqref{eq:agg}.
In this figure, we ignore the central node $v$ for clarity and illustrate for $n = 5$ and $n = 6$.
Here we take $n = 5$ as an example to explain Figure \ref{fig:illus:K5}.

The very left column shows the components of Figure \ref{fig:complete} and Eq.~\eqref{eq:agg}, and the right four columns provide the decoupled illustrations of Figure \ref{fig:complete} and Eq.~\eqref{eq:agg}.
The first row of the right four columns lists the group action $gu_i$ ($g$ acts on $u_i$) defined by Corollary \ref{thm:action}, where $u_i$ ranges from $u_1$ to $u_5$, $g \in \mathfrak{G} = \{ e, \sigma, \sigma^2, \sigma^3 \}$ and $\mathfrak{G}$ is defined by Lemma \ref{thm:cyclic}.
For readers unfamiliar with group theory, the third row of the right four columns explicitly provides the corresponding action results of $gu_i$, such as $\sigma^2 u_1 = u_1, \sigma^2 u_2 = u_5, \sigma^2 u_3 = u_4, \sigma^2 u_4 = u_3, \sigma^2 u_5 = u_2$ in the third column.
In addition, these four columns are associated with each other by the generator $\sigma$.
For example, in the third row, after $\sigma$ acts on the action results in the first column, they are transformed into the action results in the second column according to the permutation diagram, i.e., $\sigma u_1 = u_1, \sigma u_2 = u_4, \sigma u_3 = u_2, \sigma u_4 = u_5, \sigma u_5 = u_3$.
% Equivalently, in the first row, we have $\sigma (eu_1) = \sigma u_1, \sigma (eu_2) = \sigma u_2, \sigma (eu_3) = \sigma u_3, \sigma (eu_4) = \sigma u_4, \sigma (eu_5) = \sigma u_5$.
Action results in other columns are transformed in a similar manner and form a cyclic structure.
The second row of the right four columns illustrates this process.
% Similarly, after $\sigma$ acts consecutively, the 2nd column move to the 3rd one, the 3rd move to the 4th, the 4th move to the 1st, forming a cyclic structure.
% And the second row of the right four columns reveals such a dynamic action process of $\sigma$.

In each column, after obtaining the action results of $gu_1, \ldots, gu_n$, we arrange these $n = 5$ neighbors (action results) as an \emph{undirected} ring.
The first $\lfloor n/2 \rfloor = 2$ arrangements (marked by solid lines) are constructed according to Theorem \ref{thm:arrange}, and the last $\lfloor n/2 \rfloor = 2$ arrangements (marked by dashed lines) reverse the former.
Either the first or the last $\lfloor n/2 \rfloor = 2$ arrangements cover all \emph{undirected} 2-ary dependencies.
Then, we use permutation-sensitive RNNs to model the 2-ary dependencies in a \emph{directed} manner (since permutation-sensitive RNNs serve $a \to b$ and $b \to a$ as two different pairs) and construct the corresponding Hamiltonian cycles.
As a result, the Hamiltonian cycles are modeled bi-directionally, and edges in Hamiltonian cycles are transformed into an undirected manner.
The arrangement generation and Hamiltonian cycle construction are detailed in Section \ref{sec:strategy}.

Figure \ref{fig:illus:K6} presents in a similar way as Figure \ref{fig:illus:K5} does.
However, we do not show all six columns due to the limited space.
Here we omit the 5th and the 6th columns, which illustrate the modeling processes based on group elements $\sigma^4$ and $\sigma^5$.
% Please refer to \url{https://github.com/zhongyu1998/PG-GNN/tree/main/figures} for a complete illustration of Figure \ref{fig:illus:K6}.

\subsection{Discussion on Groups}
\label{sec:2-ary}

Since the (permutation) group in Eq.~\eqref{eq:agg} plays a pivotal role in our model, it is necessary to discuss the motivation for using groups and why we select the specific group.
In fact, the group is used to effectively model all 2-ary dependencies (pairwise correlations).
We first summarize why the modeling of all 2-ary dependencies is indispensable:

\begin{itemize}
\item \emph{Expressive power}.
Modeling all 2-ary dependencies can capture whether any two neighboring nodes are connected, helping our model count incidence triangles and 4-cliques hence improving its expressive power.
\item \emph{Generalization capability and computational complexity}.
Modeling all 2-ary dependencies can make these dependencies invariant to arbitrary permutations of node orderings.
Such an invariance to 2-ary dependencies is an approximation of the permutation-invariance and helps to guarantee the generalization capability.
Moreover, it also avoids considering all $n!$ permutations to strictly ensure the permutation-invariance, thereby significantly reducing the computational complexity.
\item \emph{Robustness}.
Modeling all 2-ary dependencies makes our model insensitive to a specific 2-ary dependency and robust to potential data noise and adversarial perturbations.
\end{itemize}

In order to effectively cover all 2-ary dependencies \emph{with the lowest complexity}, we try to design a special group to accomplish this goal.
According to Cayley’s Theorem (Theorem \ref{thm:Cayley}) that ``Every finite group is isomorphic to a permutation group'', we focus on finding permutation groups instead of all finite groups.
Hence the problem is converted to finding the basic element of the permutation group, i.e., the permutation.
Lemma \ref{thm:cyclic} defines a permutation group $\mathfrak{G}$ and constructs its permutation $\sigma$ (Eq.~\eqref{eq:sigma}) based on Theorem \ref{thm:arrange}, which has been proven to reach the theoretical lower bound of the sampling complexity when sampling permutations to cover all 2-ary dependencies.
This permutation group is isomorphic to the cyclic group, the simplest group to achieve linear sampling complexity.
On the contrary, other groups, such as dihedral group $\mathcal{D}_n$, alternating group $\mathcal{A}_n$, symmetric group $\mathcal{S}_n$, etc., will lead to higher nonlinear complexity hence sacrificing efficiency.
Thus, in terms of computational efficiency, group $\mathfrak{G}$ defined in Lemma \ref{thm:cyclic} is the best choice, which drives us to apply it to our model design (Eq.~\eqref{eq:agg}).

\subsection{Discussion on Model Variants}
\label{sec:variant}

Since our proposed model mainly focuses on modeling all 2-ary dependencies, the most intuitive way is to enumerate all $n(n-1)$ \emph{bi-directional} 2-ary dependencies between the $n$ neighbors of the central node $v$ and then sum them up, which can be formulated as follows:
\begin{equation}
\label{eq:twin}
    \bm h_{v}^{(k)} = \sum_{\substack{u_i, u_j \in \mathcal{N}(v) \\ u_i \neq u_j}} {\text{RNN} \left( \bm h_{u_i}^{(k-1)}, \bm h_{u_j}^{(k-1)} \right)} + \bm W_{\text{self}}^{(k-1)} \bm h_{v}^{(k-1)}
\end{equation}
Besides, we can also merge the central node $v$ into RNN to form $n(n-1)$ triplets:
\begin{equation}
\label{eq:triplet}
    \bm h_{v}^{(k)} = \sum_{\substack{u_i, u_j \in \mathcal{N}(v) \\ u_i \neq u_j}} {\text{RNN} \left( \bm h_{u_i}^{(k-1)}, \bm h_{u_j}^{(k-1)}, \bm h_{v}^{(k-1)} \right)}
\end{equation}
In fact, both these two naive variants and our proposed Eq.~\eqref{eq:agg} can model all 2-ary dependencies.
However, each term $\left( \bm h_{u_i}^{(k-1)}, \bm h_{u_j}^{(k-1)} \right)$ in Eq.~\eqref{eq:twin} can only capture a 2-ary dependency, and each term $\left( \bm h_{u_i}^{(k-1)}, \bm h_{u_j}^{(k-1)}, \bm h_{v}^{(k-1)} \right)$ in Eq.~\eqref{eq:triplet} can only capture a triplet (3-ary dependency).
Contrary to these two naive variants, each term $\left( \bm h_{g u_1}^{(k-1)}, \cdots , \bm h_{g u_n}^{(k-1)}, \bm h_{g u_1}^{(k-1)} \right)$ in Eq.~\eqref{eq:agg} encodes all neighbors as a higher-order $n$-ary dependency, which contains more information and is more powerful than 2-ary or 3-ary dependency.

On the other hand, we can also integrate all terms of Eq.~\eqref{eq:agg} into only one term, and use a single RNN to model it as follows:
\begin{equation}
\label{eq:seq}
    \bm h_{v}^{(k)} = \text{RNN} \left( \concat_{g \in \mathfrak{G}} {\left( \bm h_{g u_1}^{(k-1)}, \bm h_{g u_2}^{(k-1)}, \cdots , \bm h_{g u_n}^{(k-1)}, \bm h_{g u_1}^{(k-1)} \right)} \right) + \bm W_{\text{self}}^{(k-1)} \bm h_{v}^{(k-1)}, u_{1:n} \in \mathcal{N}(v)
\end{equation}
where $\parallel$ is the concatenation operation.
For example, if $n$ is even, it concatenates $g \in \mathfrak{G}$ as:
\begin{align*}
    \concat_{g \in \mathfrak{G}} {\left( \bm h_{g u_1}^{(k-1)}, \bm h_{g u_2}^{(k-1)}, \cdots , \bm h_{g u_n}^{(k-1)}, \bm h_{g u_1}^{(k-1)} \right)}
    =&~\bm h_{e u_1}^{(k-1)}, \bm h_{e u_2}^{(k-1)}, \cdots , \bm h_{e u_n}^{(k-1)}, \bm h_{e u_1}^{(k-1)}, \\
    &~\bm h_{\sigma u_1}^{(k-1)}, \bm h_{\sigma u_2}^{(k-1)}, \cdots , \bm h_{\sigma u_n}^{(k-1)}, \bm h_{\sigma u_1}^{(k-1)}, \\
    &~\cdots, \\
    &~\bm h_{\sigma^{n-1} u_1}^{(k-1)}, \bm h_{\sigma^{n-1} u_2}^{(k-1)}, \cdots , \bm h_{\sigma^{n-1} u_n}^{(k-1)}, \bm h_{\sigma^{n-1} u_1}^{(k-1)}
\end{align*}
Although this variant can model all $n(n-1)$ 2-ary dependencies in a single term, the time complexity is problematic.
Since the concatenation operation orders these representations $\bm h_{*}^{(k-1)}$, Eq.~\eqref{eq:seq} can only be processed serially with the time complexity of $\Theta(N\Delta^2 c^2)$.
This drawback hinders us from effectively balancing the expressive power and computational cost.
In contrast, as explained in Section \ref{sec:arch}, our proposed Eq.~\eqref{eq:agg} can be computed in parallel with lower time complexity of $\Theta(N\Delta c^2)$, making it more efficient in practice.

\section{Analysis of Sampling Complexity}
\label{sec:obs}

In this section, we first consider a variant of the coupon collector's problem (Problem \ref{thm:kcoupon}) and find the analytical solution to it.
Then, we use the solution of Problem \ref{thm:kcoupon} to estimate the sampling complexity of $\pi$-SGD optimization (proposed by Janossy Pooling \cite{murphy2019janossy} and Relational Pooling \cite{murphy2019relational}).
Finally, we conduct numerical experiments to verify the rationality of our estimation.

\subsection{A Variant of Coupon Collector's Problem}
\label{sec:coupon}

The coupon collector's problem is a famous probabilistic paradigm arising from the following scenario.

\begin{appproblem}[Coupon Collector's Problem]
\label{thm:coupon}
    Suppose there are $m$ different types of coupons, and each time one chooses a coupon independently and uniformly at random from the $m$ types.
    One needs to collect $m H(m) = m \ln m + \mathcal{O}(m)$ coupons on average before obtaining at least one of every type of coupon, here $H(m) = \sum_{i=1}^{m}{\frac{1}{i}}$ is the $m$-th harmonic number \cite{mitzenmacher2017probability}.
\end{appproblem}

In order to estimate the sampling complexity of $\pi$-SGD optimization, we need a more sophisticated analysis of the coupon collector’s problem.
The following problem is the generalization of Problem \ref{thm:coupon} from one coupon to $k (k \ge 1)$ coupons at each time, providing a theoretical foundation for our discussion in Section \ref{sec:obs_expt}.

\begin{appproblem}[$k$-Coupon Collector's Problem]
\label{thm:kcoupon}
	Suppose there are $m$ different types of coupons, and each time one chooses $k$ coupons ($k \ge 1$, without repetition) independently and uniformly at random from the $m$ types.
	One needs to collect $\sum\limits_{i=1}^{m} {(-1)^{i+1} \frac{\tbinom{m}{i}}{1 - \left. {\tbinom{m-i}{k}} \middle/ {\tbinom{m}{k}} \right.}}$ times on average before obtaining at least one of every type of coupon.
\end{appproblem}

\begin{proof}
Let $X$ be the collecting times until at least one of every type of coupon is obtained.
We start by considering the probability that $X$ is greater than $s$ when $s$ is fixed.
For $j = 1, \ldots, m$, let $A_j$ denote the event that no type $j$ coupon is collected in the first $s$ times.
By the inclusion-exclusion principle,
\begin{align*}
    \Pr(X>s)
    =& \Pr\left( \bigcup_{j=1}^{m}{A_j} \right) \\
    =& \sum_{1 \le j_1 \le m}{\Pr(A_{j_1})} - \sum_{1 \le j_1 < j_2 \le m}{\Pr(A_{j_1} \cap A_{j_2})} + \sum_{1 \le j_1 < j_2 < j_3 \le m}{\Pr(A_{j_1} \cap A_{j_2} \cap A_{j_3})} \\
    & - \cdots + {(-1)^{m+1}}\Pr(A_1 \cap \cdots \cap A_m) \\
    =& \sum_{i=1}^{m}{(-1)^{i+1} \sum_{1 \le j_1 < \cdots < j_i \le m}{\Pr(A_{j_1} \cap \cdots \cap A_{j_i})}}
\end{align*}
where $\Pr(A_{j_1} \cap \cdots \cap A_{j_i}) = {\left[ \frac{\dbinom{m-i}{k}}{\dbinom{m}{k}} \right]}^s$, and for $1 \le j_1 < \cdots < j_i \le m$ there are $\dbinom{m}{i}$ choices.
Thus, we have
\begin{equation}
\label{eq:inclusion}
    \Pr(X>s) = \sum_{i=1}^{m}{(-1)^{i+1} \binom{m}{i} {\left[ \frac{\dbinom{m-i}{k}}{\dbinom{m}{k}} \right]}^s}
\end{equation}
Since $X$ takes only positive integer values, we can compute its expectation by
\begin{equation}
\label{eq:exp}
    \mathbb{E}[X] = \sum_{s=1}^{\infty} {s \cdot \Pr(X=s)} = \sum_{s=0}^{\infty} {\Pr(X>s)}
\end{equation}
Using Eq.~\eqref{eq:inclusion} in Eq.~\eqref{eq:exp}, we obtain
\begin{align*}
    \mathbb{E}[X]
    &= \sum_{s=0}^{\infty}{\sum_{i=1}^{m}{(-1)^{i+1} \binom{m}{i} {\left[ \frac{\dbinom{m-i}{k}}{\dbinom{m}{k}} \right]}^s}} \\
    &= \sum_{i=1}^{m}{(-1)^{i+1} \binom{m}{i} \sum_{s=0}^{\infty} {\left[ \frac{\dbinom{m-i}{k}}{\dbinom{m}{k}} \right]}^s} \\
    &= \sum_{i=1}^{m} {(-1)^{i+1} \frac{\dbinom{m}{i}}{1 - \left. {\dbinom{m-i}{k}} \middle/ {\dbinom{m}{k}} \right.}}
\end{align*}
\end{proof}

\subsection{Sampling Complexity Analysis of \texorpdfstring{$\pi$}{\textpi}-SGD Optimization}
\label{sec:obs_expt}

Suppose there are $n$ neighboring nodes around the central node $v$.
$\pi$-SGD optimization samples a permutation of these $n$ nodes randomly at each time and models their dependencies based on the sampled permutation.
As mentioned in the main body, we are interested in the average times of modeling all the pairwise correlations between these $n$ nodes.
This problem can be equivalently formulated in graph-theoretic language as follows:

\begin{appproblem}[Complete Graph Covering Problem]
\label{thm:complete}
    Let $G'$ be an empty graph with $n$ nodes.
    Each time we generate a Hamiltonian path at random and add the corresponding $n-1$ edges to $G'$ (edges can be generated repeatedly at different times).
    How many times does it take on average before graph $G'$ covers a complete graph $K_n$?
\end{appproblem}

It is difficult to give an analytical solution to this problem, so we try to find an approximate solution.
In fact, the complete graph covering problem (Problem \ref{thm:complete}) has an interesting connection with the $k$-coupon collector's problem (Problem \ref{thm:kcoupon}) discussed above.
The generation of a Hamiltonian path among $n$ nodes at each time is equivalent to the drawing of $n-1$ \emph{interrelated} edges\footnote{They \emph{have to} be in an end-to-end manner, e.g., 1-2, 2-3, 3-4.} from all possible $\frac{n(n-1)}{2}$ edges.
Suppose we ignore the interrelations between these $n-1$ edges and each time choose $n-1$ edges \emph{independently}\footnote{They \emph{do not} have to be in an end-to-end manner, e.g., 1-2, 1-3, 1-4.} and randomly without repetition.
In that case, Problem \ref{thm:complete} will degenerate into a special case of Problem \ref{thm:kcoupon}.
Thus, we have the following conjecture:

\begin{appconjecture}
\label{thm:conjecture}
	Suppose there are $n$ neighboring nodes around the central node $v$, and each time we sample a permutation of these $n$ nodes at random.
	How many times does it take on average before any two nodes have become neighbors at least once?
	% sampling times
	This problem is equivalent to the complete graph covering problem, which shares a similar result to the $k$-coupon collector's problem:
	Suppose there are $m = \frac{n(n-1)}{2}$ different types of coupons, and each time one chooses $k = n-1$ coupons (without repetition) independently and uniformly at random from the $m$ types.
	How many times does it take on average before obtaining at least one of every type of coupon?
	% collecting times
\end{appconjecture}

\begin{figure}
    \centering
    \includegraphics[width=1.0\linewidth]{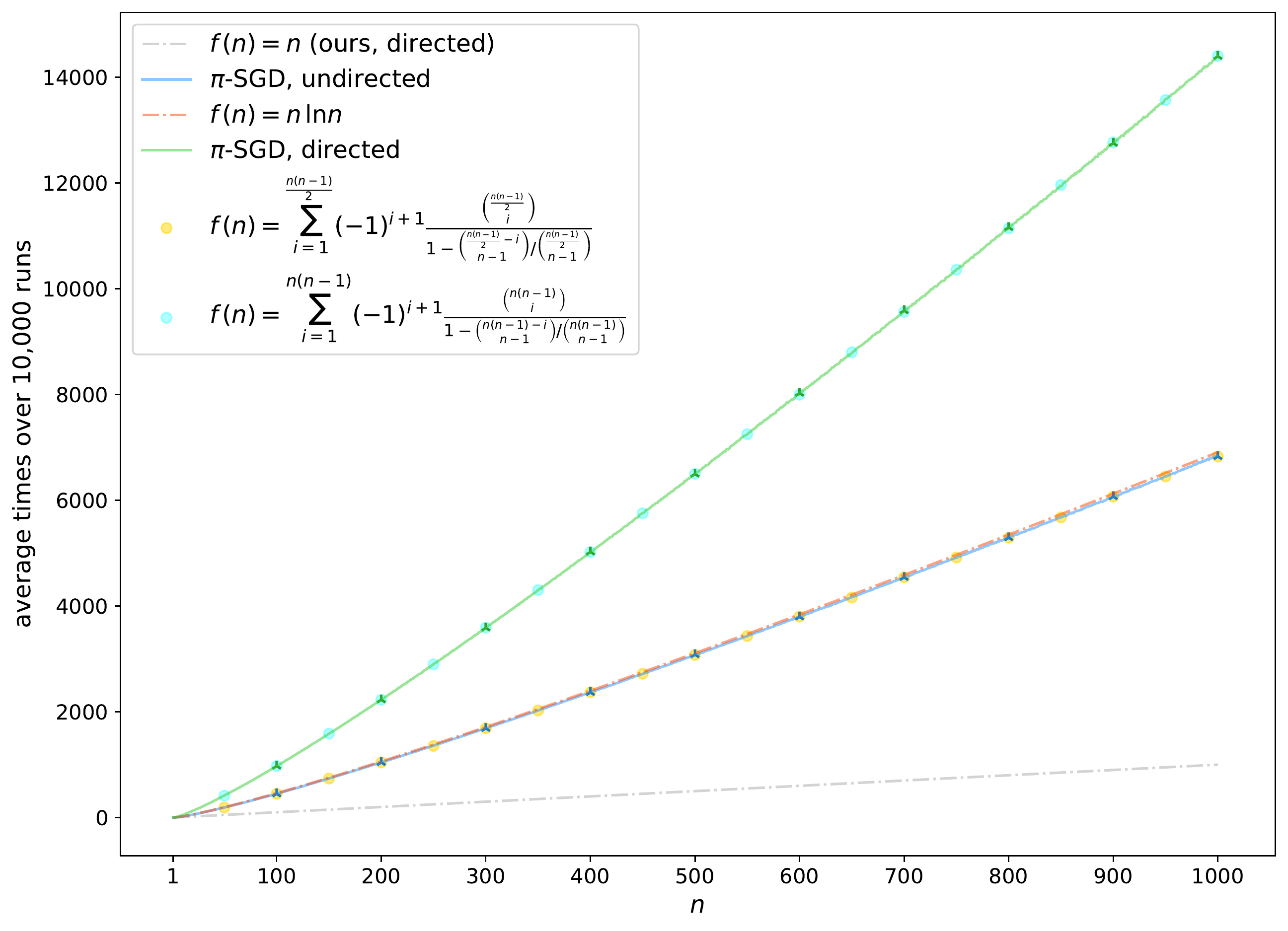}
    \caption{The experimental results of numerical simulation.
    Since the results of the complete graph covering problem are equal to the sampling times of $\pi$-SGD optimization, we label them as ``$\pi$-SGD''.
    The blue \textcolor{blue}{---} and green \textcolor{green}{---} lines represent undirected (with $\frac{n(n-1)}{2}$ undirected edges) and directed (with $n(n-1)$ bi-directional edges) cases, respectively.
    In addition, since the $k$-coupon collector's problem gives almost the same results as the complete graph covering problem, we only show 20 points ($\color{yellow}\bullet$,$\color{cyan}\bullet$) uniformly for the numerical results of the closed-form expression of the $k$-coupon collector's problem for clarity.
    The light yellow $\color{yellow}\bullet$ and light blue $\color{cyan}\bullet$ points represent undirected ($m = \frac{n(n-1)}{2}$, $k = n-1$) and directed ($m = n(n-1)$, $k = n-1$) cases, respectively.
    We highlight the results of $\pi$-SGD at the points that $n$ are multiples of 100 (marked by blue $\color{blue}\upY$ and green $\color{green}\upY$ triangular stars) for comparison with those of the $k$-coupon collector's problem (marked by light yellow $\color{yellow}\bullet$ and light blue $\color{cyan}\bullet$ points).}
    \label{fig:simulation}
\end{figure}

Since the analytical solution to the $k$-coupon collector's problem has been given by Problem \ref{thm:kcoupon} in Section \ref{sec:coupon}, we can use it to approximate the result of Problem \ref{thm:complete} and estimate the sampling complexity of $\pi$-SGD optimization.
We also conduct extensive numerical experiments to compare the results of Problem \ref{thm:complete} with those of Problem \ref{thm:kcoupon} when $n$ ranges from 1 to 1,000.
We consider both undirected and directed cases for Problem \ref{thm:complete}, there are $\frac{n(n-1)}{2}$ undirected and $n(n-1)$ bi-directional edges, respectively.
Correspondingly, Problem \ref{thm:kcoupon} takes $m = \frac{n(n-1)}{2}$ and $m = n(n-1)$ coupons.
For each $n$, we conduct experiments for 10,000 runs and report the average times of covering these edges/coupons.
As shown in Figure \ref{fig:simulation}, Problem \ref{thm:complete} ($\pi$-SGD) gives almost the same numerical results as Problem \ref{thm:kcoupon} (the closed-form expression), verifying the rationality of Conjecture \ref{thm:conjecture}.
Hence, we conclude the following observation:

\begin{appobservation}
\label{thm:obs}
    Suppose there are $n$ neighboring nodes around the central node $v$, and each time we sample a permutation of these $n$ nodes at random.
    Any two nodes have been neighbors at least once after $\sum\limits_{i=1}^{\tfrac{n(n-1)}{2}} {(-1)^{i+1} \tfrac{\tbinom{\tfrac{n(n-1)}{2}}{i}} {1 -  {\tbinom{\tfrac{n(n-1)}{2}-i}{n-1}} \Bigl/ {\tbinom{\tfrac{n(n-1)}{2}}{n-1}} }}$ approximately $\mathcal{O}(n \ln n)$ times on average.
\end{appobservation}

It is worth noting that our approach only needs $\frac{n}{2}$ times in the undirected cases and $n$ times in the directed cases.
According to the conclusion of Observation \ref{thm:obs}, if the degree of the central node $v$ is $n = 100$, our approach saves $\frac{100 \ln 100}{100/2} \approx 9 $ times compared to $\pi$-SGD optimization.
If $n = 1000$, e.g., a hub node in the large-scale network dataset, our approach saves $\frac{1000 \ln 1000}{1000/2} \approx 14 $ times.

\section{Analysis of Computational Complexity}

In this section, we first provide the time and space complexity analysis of the recent related works MPSN \cite{bodnar2021weisfeilers} and CWN \cite{bodnar2021weisfeilerc}, then conduct comprehensive experiments about memory consumption to validate the efficiency of our proposed PG-GNN.

\subsection{Computational Complexity Analysis of MPSN and CWN}
\label{sec:SCIN}

Recently, a batch of works \cite{bodnar2021weisfeilers, bodnar2021weisfeilerc} exploited local high-order interactions to effectively improve the expressive power of GNNs.
MPSN \cite{bodnar2021weisfeilers} focuses on simplicial complexes, which are composed of simplices that generalize the 2-dimensional triangle to arbitrary $k$ dimensions.
CWN \cite{bodnar2021weisfeilerc} further generalizes simplicial complexes of MPSN to cell complexes.
Hence these works are particularly relevant to our approach.
The core idea of these works is to use the \emph{lifting transformation}, which maps graphs to more distinguishable simplicial complexes \cite{bodnar2021weisfeilers} or cell complexes \cite{bodnar2021weisfeilerc} by adding additional structures, e.g., attaching $k$-simplices ($k \ge 2$) to $(k+1)$-cliques \cite{bodnar2021weisfeilers} and 2-cells to induced cycles \cite{bodnar2021weisfeilerc}.
% Higher-order structures are first-class objects thereon and message-passing is directly performed on them \cite{bevilacqua2022equivariant}.
Here we mainly discuss the CWN since it is more powerful yet efficient than MPSN.

Let $N$ and $M$ denote the number of nodes and edges, respectively.
Let $X$ be a $d$-dimensional regular cell complex, $B_p$ be the maximum boundary size of a $p$-cell in $X$, and $S_p$ be the number of $p$-cells.
For CWN, the time complexity is $\mathcal{O} \left( \sum_{p=1}^{d}{\left( B_p S_p + 2 \cdot \binom{B_p}{2} S_p \right)} \right)$, and the space complexity is $\mathcal{O} \left( N + \sum_{p=1}^{d}{\left( S_p + B_p S_p + 2 \cdot \binom{B_p}{2} S_p \right)} \right)$.
Next, we analyze the time complexity for a generic lifting transformation that maps a graph to a 2-dimensional cell complex and attaches 2-cells to all the induced cycles in the graph.
Since 0-cells, 1-cells, and 2-cells represent vertices, edges, and induced cycles, respectively, we have $d = 2$, $S_1 = M$, $B_1 = 2$, and $B_2$ equals the size of the maximum induced cycle considered.
In the case of molecular graphs, the number of induced cycles (chemical rings), $S_2$, is usually upper-bounded by a small constant.
Accordingly, CWN achieves outstanding empirical performance and efficiency on molecular tasks.
However, in the case of social networks, there are usually $\Omega(N^2)$ triangles (see IMDB-B, IMDB-M, and COLLAB in Table \ref{tab:realstats}), even without mentioning other types of induced cycles.
Thus we have $S_2 = \Omega(N^2)$, and the time complexity is $\mathcal{O} \left( 4M + B_2 S_2 + 2 \cdot \binom{B_2}{2} S_2 \right) \ge \mathcal{O}(N^2)$.
To make matters worse, for general graph distributions, $S_2$ may grow exponentially with the number of nodes \cite{bodnar2021weisfeilerc}, and the computation of the pre-processing step (lifting transformation) may also be intractable.
In a nutshell, the computational complexity may hinder the application of CWN outside of the molecular domain, where the importance of specific substructures is not well understood and their number may grow rapidly.
% Still, the worst-case exponential complexity may hinder their application in the presence of graphs of non-characterized distribution or when substructures of more efficient matching are not known to play a relevant role \cite{bevilacqua2022equivariant}.
% This makes these models of less obvious application outside of the molecular and social domain, where the importance of specific substructures is not well understood and their number may rapidly grow \cite{bevilacqua2022equivariant}.

\subsection{Memory Cost Analysis}
\label{sec:memory}

\begin{table}\small
    \centering
    \caption{CPU RAM consumption (MiB) on real-world datasets.
    The \textcolor{gray}{gray font} denotes the consumption of the pre-processing stage (i.e., lifting transformation) of CIN.}
    \label{tab:RAM}
    \begin{tabular}{lccccccc}
    \toprule
    Model   & PROTEINS & NCI1 & IMDB-B & IMDB-M & COLLAB & MNIST & ZINC\\
    \midrule
    GIN \cite{xu2019powerful}
            & 2,338 & 2,460 & 2,337 & 2,343 & 11,351 & 24,946 & 3,017 \\
    \textcolor{gray}{CIN \cite{bodnar2021weisfeilerc}, pre-proc.}
            & \textcolor{gray}{561}     & \textcolor{gray}{627}
            & \textcolor{gray}{769}     & \textcolor{gray}{745}
            & \textcolor{gray}{N/A}     & \textcolor{gray}{N/A}
            & \textcolor{gray}{1,558}   \\
    CIN \cite{bodnar2021weisfeilerc}, training
            & 2,689 & 2,749 & 2,953 & 3,001 & N/A    & N/A    & 2,993 \\
    PG-GNN (Ours)
            & 2,343 & 2,466 & 2,351 & 2,349 & 11,298 & 24,955 & 3,020 \\
    \bottomrule
    \end{tabular}
\end{table}

\begin{table}\small
    \centering
    \caption{GPU memory consumption (MiB) on real-world datasets. ``OOM'' means out of memory ($>$~24,220MiB).}
    \label{tab:memory}
    \begin{tabular}{lccccccc}
    \toprule
    Model   & PROTEINS & NCI1 & IMDB-B & IMDB-M & COLLAB & MNIST & ZINC\\
    \midrule
    GIN \cite{xu2019powerful}
            & 887   & 889   & 881   & 877    & 1,125  & 981   & 901    \\
    CIN \cite{bodnar2021weisfeilerc}
            & 2,039 & 1,033 & 3,891 & 13,361 & OOM    & N/A   & 1,371  \\
    PG-GNN (Ours)
            & 980   & 1,142 & 1,202 & 1,036  & 21,485 & 4,127 & 1,367  \\
    \bottomrule
    \end{tabular}
\end{table}

According to \citet{bodnar2021weisfeilerc}, in all experiments, they employ a model which stacks CWN layers with local aggregators as in GIN, thus naming their architecture ``Cell Isomorphism Network'' (CIN).
Here we use GIN and CIN as our baselines to compare the memory consumption of different models.
We use the codes released by the authors of GIN\footnote{\url{https://github.com/weihua916/powerful-gnns}} and CIN\footnote{\url{https://github.com/twitter-research/cwn}}, and run experiments with the (optimal) hyper-parameter configurations reported in their original papers to keep the comparison as fair as possible.
Tables \ref{tab:RAM} and \ref{tab:memory} summarize the CPU RAM and GPU memory consumption for various models, respectively.
Note that the total CPU RAM consumption of CIN should be computed as the consumption of ``\textcolor{gray}{pre-processing}'' $+$ ``training'', while other models do not require the extra pre-processing steps.
As shown in the tables, the memory cost of CIN grows rapidly outside of the molecular domain, such as on social networks and MNIST, consistent with our analysis above.
In contrast, our PG-GNN is memory-efficient and outperforms CIN in terms of memory cost on almost all datasets, even performing on par with GIN on most datasets.

\section{Details of the Experiments}

\subsection{Details of Datasets}
\label{sec:dataset}

In this subsection, we provide detailed descriptions of datasets used in our experiments.
The statistics of real-world datasets are summarized in Table \ref{tab:realstats}.

\begin{table}\small
    \centering
    \caption{Statistics of real-world datasets. The degree denotes in-degree / out-degree for MNIST containing directed graphs.}
    \label{tab:realstats}
    \begin{tabular}{lccccccc}
    \toprule
    Property    & PROTEINS & NCI1 & IMDB-B & IMDB-M & COLLAB & MNIST & ZINC \\
    \midrule
    Graphs      & 1,113 & 4,110 & 1,000 & 1,500 & 5,000 & 70,000 & 12,000   \\
    Classes     & 2     & 2     & 2     & 3     & 3     & 10    & N/A   \\
    Nodes (avg) & 39.06 & 29.87 & 19.77 & 13.00 & 74.49 & 70.57 & 23.16 \\
    Nodes (max) & 620   & 111   & 136   & 89    & 492   & 75    & 37    \\
    Degree (avg)& 3.73  & 2.16  & 9.76  & 10.14 & 65.97 & 8.00 / 8.00 & 2.15\\
    Degree (max)& 25    & 4     & 135   & 88    & 491   & 18 / 8& 4 \\
    Triangles (avg)
                & 27.40 & 0.05  & 391.99 & 305.90 & 124551.40 & 626.07 & 0.06\\
    Triangles (max)
                & 534   & 3     & 6,985 & 14,089 & 2,574,680 & 702 & 2 \\
    \bottomrule
    \end{tabular}
\end{table}

\subsubsection{Synthetic Datasets}

We conduct synthetic experiments of counting incidence substructures on two types of random graphs: Erd\H{o}s-R{\'e}nyi random graphs and random regular graphs, created by \citet{chen2020can}.
The first one consists of 5,000 Erd\H{o}s-R{\'e}nyi random graphs with 10 nodes in each graph, and each edge exists with a probability of 0.3.
The second one consists of 5,000 random regular graphs with $n$ nodes in each graph and the degree of $d$, where $(n, d)$ is uniformly sampled from \{(10, 6), (15, 6), (20, 5), (30, 5)\}.
Both datasets are randomly split into 30\%, 20\%, and 50\% for training, validation, and testing.
% The input feature is set as the adjacency matrix for each graph.

For the incidence triangle counting task, all nodes are labeled with Eq.~\eqref{eq:triangle}.
For the incidence 4-clique counting task, it is hard to derive such a closed-form expression as Eq.~\eqref{eq:triangle}, so we manually label each central node by counting how many groups of three neighboring nodes are fully connected.
% so we manually label each central node with the number of groups of three neighboring nodes that are fully connected.
The evaluation metric of the incidence substructure counting task is the mean absolute error (MAE) between the predicted and true number of incidence substructures for each node.

\subsubsection{TUDataset}

\paragraph{Bioinformatics.}
PROTEINS is a dataset in which each graph represents a protein, and nodes represent secondary structure elements (SSEs) within the protein structure, i.e., \emph{helices}, \emph{sheets}, and \emph{turns}.
An edge connects two nodes if they are neighbors in the amino-acid sequence or 3D space.
The task is to classify the proteins into enzymes and non-enzymes.
NCI1 is a publicly available dataset collected by the National Cancer Institute (NCI).
Each graph represents a chemical compound, in which nodes and edges represent atoms and chemical bonds.
This dataset is related to anti-cancer screening, and the task is to predict whether the chemical compounds are positive or negative for cell lung cancer.

\paragraph{Social Networks.}
IMDB-BINARY is a movie-collaboration dataset containing the actor/actress and genre information of different movies on IMDB.
Each graph corresponds to an actor/actress's ego network, in which nodes correspond to actors/actresses, and an edge indicates two actors/actresses appear in the same movie.
These graphs are derived from \emph{Action} and \emph{Romance} genres.
And the task is to classify the graphs into their genres.
IMDB-MULTI is the multi-class version of IMDB-BINARY and contains a balanced set of ego networks derived from \emph{Comedy}, \emph{Romance}, and \emph{Sci-Fi} genres.
COLLAB is a scientific collaboration dataset.
Each graph corresponds to a researcher's ego network, in which nodes correspond to the researcher and its collaborators, and an edge indicates the collaboration between two researchers.
These researchers come from different fields, i.e., \emph{High Energy Physics}, \emph{Condensed Matter Physics}, and \emph{Astro Physics}.
The task is to classify the graphs into the fields of corresponding researchers.

\subsubsection{MNIST and ZINC}

\paragraph{MNIST.}
MNIST \cite{lecun1998gradient} is a classical image classification dataset.
The original MNIST images are converted into graphs using super-pixels \cite{achanta2012slic}.
% Super-pixels represent small regions of homogeneous intensity in images and can be extracted with the SLIC technique \cite{achanta2012slic}.
Each graph represents an image, and its adjacency matrix is built with 8-nearest neighbors for each node (super-pixel).
Note that since the relationship between each super-pixel (node) and its nearest neighbors is asymmetric, the resultant adjacency matrices are also asymmetric.
For more details about the generation, please refer to Appendix A.2 in \citet{dwivedi2020benchmarking}.
The resultant graphs are of sizes 40-75 super-pixels, and each node’s features are assigned with super-pixel coordinates and intensity.
MNIST has 55,000 training, 5,000 validation, and 10,000 testing graphs, where the 5,000 graphs for the validation set are randomly sampled from the training set.
The evaluation metric for MNIST is the classification accuracy between the predicted class and ground-truth label for each graph.

\paragraph{ZINC.}
ZINC \cite{irwin2012zinc} is one of the most popular real-world molecular datasets with 250K graphs, out of which \citet{dwivedi2020benchmarking} randomly select 12K for efficiency.
Each graph represents a molecule, where nodes and edges represent atoms and chemical bonds, respectively.
The node features are the types of heavy atoms encoded in a one-hot manner.
The task is to predict the constrained solubility, an important chemical property for molecules.
ZINC has 10,000 training, 1,000 validation, and 1,000 testing graphs.
The evaluation metric for ZINC is the mean absolute error (MAE) between the predicted and true constrained solubility for each molecular graph.

\begin{table}\small
    \centering
    \caption{Hyper-parameter configurations on synthetic datasets.}
    \label{tab:synparam}
    \begin{tabular}{lccccccc}
    \toprule
    Hyper-parameter
                & GCN & GraphSAGE & GIN & rGIN & RP & LRP & PG-GNN \\
    \midrule
    batch size  & 32    & 32    & 32    & 32    & 16    & 16    & 16    \\
    hidden units& 64    & 64    & 64    & 64    & 64    & 64    & 64    \\
    layers      & 3     & 3     & 5     & 5     & 5     & 5     & 5     \\
    dropout     & 0.5   & 0.0   & 0.5   & 0.5   & 0.0   & 0.0   & 0.0   \\
    initial lr  & 0.01  & 0.01  & 0.01  & 0.01  & 0.01  & 0.001 & 0.001 \\
    \bottomrule
    \end{tabular}
\end{table}

\subsection{Details of Hyper-Parameters}
\label{sec:hyper}

\subsubsection{Synthetic Datasets}

We select the architectural hyper-parameters based on the performance in the validation set.
The hyper-parameter search space is listed as follows:
the batch size in \{16, 32, 64\}, the number of hidden units in \{16, 32, 64\}, the number of layers in \{3, 4, 5\}, the dropout ratio in \{0.0, 0.5\} after the final prediction layer, the initial learning rate in \{0.01, 0.005, 0.001\}, the decay rate in \{0.5, 0.9\}, the decay rate patience in \{5, 10, 15, 20, 25\}, and the aggregator in \{SRN, GRU, LSTM\}.

\paragraph{Configurations of Baselines.}
We use the default hyper-parameter configurations reported in their original papers.
Specifically, we follow \citet{hamilton2017inductive} to sample 2-hop neighborhoods for each node, set the neighborhood sample sizes $S_1$ and $S_2$ of 1-hop and 2-hop to both 5, and use LSTM \cite{hochreiter1997long} as the aggregator in GraphSAGE.
We use the uniform distribution over $D = \{0, 0.01, 0.02, \ldots, 0.99\}$ as the random distribution $\mu$ in rGIN like \citet{sato2021random}.
We set the dimension $m$ of one-hot node IDs to 10 and use GIN \cite{xu2019powerful} as the backbone in RP following \citet{murphy2019relational}.
According to \citet{chen2020can}, we set the depth $l$ and width $k$ to 1 and 3 in LRP.
Other hyper-parameters on different models are shown in Table \ref{tab:synparam}.

\paragraph{Configurations of PG-GNN.}
We report the hyper-parameters chosen by our model selection procedure as follows.
For all tasks and datasets, 5 GNN layers (including the input layer) are applied, and the LSTMs with 2 layers are used as the aggregation functions.
Batch normalization \cite{ioffe2015batch} is applied to every hidden layer.
All models are initialized using Glorot initialization \cite{glorot2010understanding} and trained using the Adam SGD optimizer \cite{kingma2015adam} with an initial learning rate of 0.001.
If the performance on the validation set does not improve after 20 epochs, the learning rate is then decayed by a factor of 0.5, except for the 4-clique counting task on ER graphs, whose patience is set to 25 epochs.
The training is stopped when the learning rate reaches the minimum value of 5E-6.

\subsubsection{Real-World Datasets}

\paragraph{TUDataset.}
We select the architectural hyper-parameters based on the accuracy in one random training fold.
The hyper-parameter search space is listed as follows:
the batch size in \{16, 32, 64\}, the number of hidden units in \{8, 16, 32, 64\}, the number of layers in \{3, 4, 5\}, the dropout ratio in \{0.0, 0.5\} after the final prediction layer, the initial learning rate in \{0.01, 0.005, 0.001\}, the decay rate in \{0.5, 0.9\}, the readout function in \{SUM, MEAN\}, and the aggregator in \{SRN, GRU, LSTM\}.

\paragraph{MNIST and ZINC.}
We select the architectural hyper-parameters based on the performance in the validation set.
The hyper-parameter search space is listed as follows:
the batch size in \{32, 64, 128\}, the number of hidden units in \{32, 64, 128\}, the number of layers in \{3, 4, 5\}, the dropout ratio in \{0.0, 0.5\} after the final prediction layer, the initial learning rate in \{0.01, 0.005, 0.001\}, the decay rate in \{0.5, 0.9\}, the decay rate patience in \{5, 10, 15, 20, 25\}, the readout function in \{SUM, MEAN\}, and the aggregator in \{SRN, GRU, LSTM\}.

\paragraph{Configurations.}
We report the hyper-parameters chosen by our model selection procedure as follows.
For all datasets, 3 or 5 GNN layers (including the input layer) are applied, and the LSTMs with 2 layers are used as the aggregation functions.
Batch normalization \cite{ioffe2015batch} is applied to every hidden layer.
All models are initialized using Glorot initialization \cite{glorot2010understanding} and trained using the Adam SGD optimizer \cite{kingma2015adam} with an initial learning rate of 0.001.
For TUDataset, the learning rate is decayed by a factor of 0.5 every 50 epochs.
The training is stopped when the number of epochs reaches the maximum value of 400.
For MNIST and ZINC, if the performance on the validation set does not improve after 20 and 25 epochs, the learning rate is then decayed by a factor of 0.5.
The training is stopped when the learning rate reaches the minimum value of 5E-6.
Other hyper-parameters on different datasets are shown in Table \ref{tab:realparam}.

\begin{table}\small
    \centering
    \caption{Hyper-parameter configurations on real-world datasets.}
    \label{tab:realparam}
    \begin{tabular}{lccccccc}
    \toprule
    Hyper-parameter
                & PROTEINS & NCI1 & IMDB-B & IMDB-M & COLLAB & MNIST & ZINC \\
    \midrule
    batch size  & 16    & 32    & 16    & 32    & 32    & 64    & 64    \\
    hidden units& 8     & 32    & 16    & 16    & 64    & 128   & 128   \\
    layers      & 5     & 5     & 5     & 5     & 3     & 5     & 5     \\
    dropout     & 0.5   & 0.0   & 0.0   & 0.5   & 0.5   & 0.0   & 0.0   \\
    degree      & False & False & True  & True  & True  & N/A   & N/A   \\
    readout     & SUM   & SUM   & \makecell[c]{SUM or\\MEAN}
                & SUM   & \makecell[c]{SUM or\\MEAN}    & MEAN  & SUM   \\
    \bottomrule
    \end{tabular}
\end{table}

\subsection{Computing Infrastructures}
\label{sec:infra}

\paragraph{Hardware Infrastructures.}
The experiments are conducted on Linux servers equipped with an Intel(R) Xeon(R) CPU E5-2650 v4 @ 2.20GHz, 256GB RAM and 8 NVIDIA TITAN RTX GPUs.

\paragraph{Software Infrastructures.}
All models are implemented using Python version 3.6, NetworkX version 2.4 \cite{hagberg2008exploring}, PyTorch version 1.4.0 \cite{paszke2019pytorch} with CUDA version 10.0.130, and cuDNN version 7.6.5.
In addition, the benchmark datasets are loaded by Deep Graph Library (DGL) version 0.4.2 \cite{wang2019deep}.

\end{document}